\documentclass[11pt]{article}

\usepackage[preprint]{acl}

\usepackage{times}
\usepackage{latexsym}

\usepackage[T1]{fontenc}

\usepackage[utf8]{inputenc}

\usepackage{microtype}

\usepackage{inconsolata}

\usepackage{graphicx}

\usepackage{booktabs}
\usepackage{multirow}
\usepackage{rotating}
\usepackage{amsmath}
\usepackage[table]{xcolor}
\usepackage{hyperref}
\usepackage{amssymb}
\usepackage{enumitem}
\usepackage{subcaption}
\usepackage{todonotes}
\usepackage{placeins}

\definecolor{takeawaybg}{HTML}{F5E6D3}
\definecolor{takeawayborder}{HTML}{D4A574}

\definecolor{promptsysborder}{HTML}{4A6FA5}
\definecolor{promptuserborder}{HTML}{6B8E5A}
\definecolor{promptasstborder}{HTML}{9B6B9E}
\definecolor{promptbg}{HTML}{F7F7F7}
\definecolor{promptvarcolor}{HTML}{B25400}
\definecolor{promptcfgcolor}{HTML}{006D5B}
\newcommand{\promptvar}[1]{\textcolor{promptvarcolor}{#1}}
\newcommand{\promptcfg}[1]{\textcolor{promptcfgcolor}{#1}}
\newcommand{\promptmsg}[3]{%
  \par\noindent
  \fcolorbox{#1}{promptbg}{\parbox{\dimexpr\linewidth-2\fboxsep\relax}{%
    {\footnotesize\textbf{\textcolor{#1}{#2}}}\\[2pt]%
    \footnotesize\ttfamily\raggedright #3}}%
  \par\smallskip
}
\newcommand{\promptsys}[1]{\promptmsg{promptsysborder}{system}{#1}}
\newcommand{\promptuser}[1]{\promptmsg{promptuserborder}{user}{#1}}
\newcommand{\promptasst}[1]{\promptmsg{promptasstborder}{assistant}{#1}}

\title{Imputation of Public Opinion Data with Large Language Models}

\author{
    \textbf{Tobias Holtdirk\textsuperscript{1,2}},
    \textbf{Georg Ahnert\textsuperscript{3}},
    \textbf{Joseph W Sakshaug\textsuperscript{1,3,4}},
    \textbf{Anna-Carolina Haensch\textsuperscript{1,2,5}} \\
    \textsuperscript{1}LMU Munich;
    \textsuperscript{2}Munich Center for Machine Learning (MCML);
    \textsuperscript{3}University of Mannheim; \\
    \textsuperscript{4}Institute for Employment Research (IAB);
    \textsuperscript{5}University of Maryland, College Park
}

\begin{document}
\maketitle
\begin{abstract}
  Large language models (LLMs) are widely evaluated as simulators of survey responses, yet applied survey research mostly faces item nonresponse.
  \textit{Imputation} fills in those missing values to restore the overall structure of a partially observed dataset and has its own well-defined evaluation criteria, which differ fundamentally from \textit{prediction} accuracy.
  We present the first evaluation of LLM imputation against proper multiple-imputation metrics and statistical baselines on more than 5 million imputed responses from four public opinion surveys under various missingness mechanisms.
  We evaluate in-context learning (ICL) for this task by investigating 84 ICL configurations and comparing the best-performing configurations on six open-weight LLMs.
  The strongest ICL approach improves on zero-shot persona prompting but remains, at best, competitive with statistical baselines (MICE PMM, TabPFN).
  ICL approaches are sensitive to survey topics and consume 2–3 orders of magnitude more compute.
  Moreover, prediction metrics commonly used in LLM survey simulation (macro-F1, TVD) do not reliably track imputation quality. Our findings highlight the importance of fit-for-purpose evaluation for LLM imputation, where established multiple imputation remains the better default.
\end{abstract}

\section{Introduction}

A large body of research uses large language models (LLMs) to simulate human survey responses~\citep[][inter alia]{argyle2023out, hu_simbench_2025} by providing an LLM with attributes of individuals or groups and prompting it to predict their answers to a specific survey question. These so-called \textit{silicon samples} use LLMs to generate entire synthetic datasets with mixed success. Applied survey research, however, rarely needs to simulate entire samples: unit nonresponse is typically addressed by weighting, whereas respondents who participate often leave individual items unanswered. Such item nonresponse is a pervasive problem in survey data analysis~\citep{leeuwPreventionTreatmentItem2003}.

\begin{figure}[t!]
  \centering
  \includegraphics[width=\linewidth]{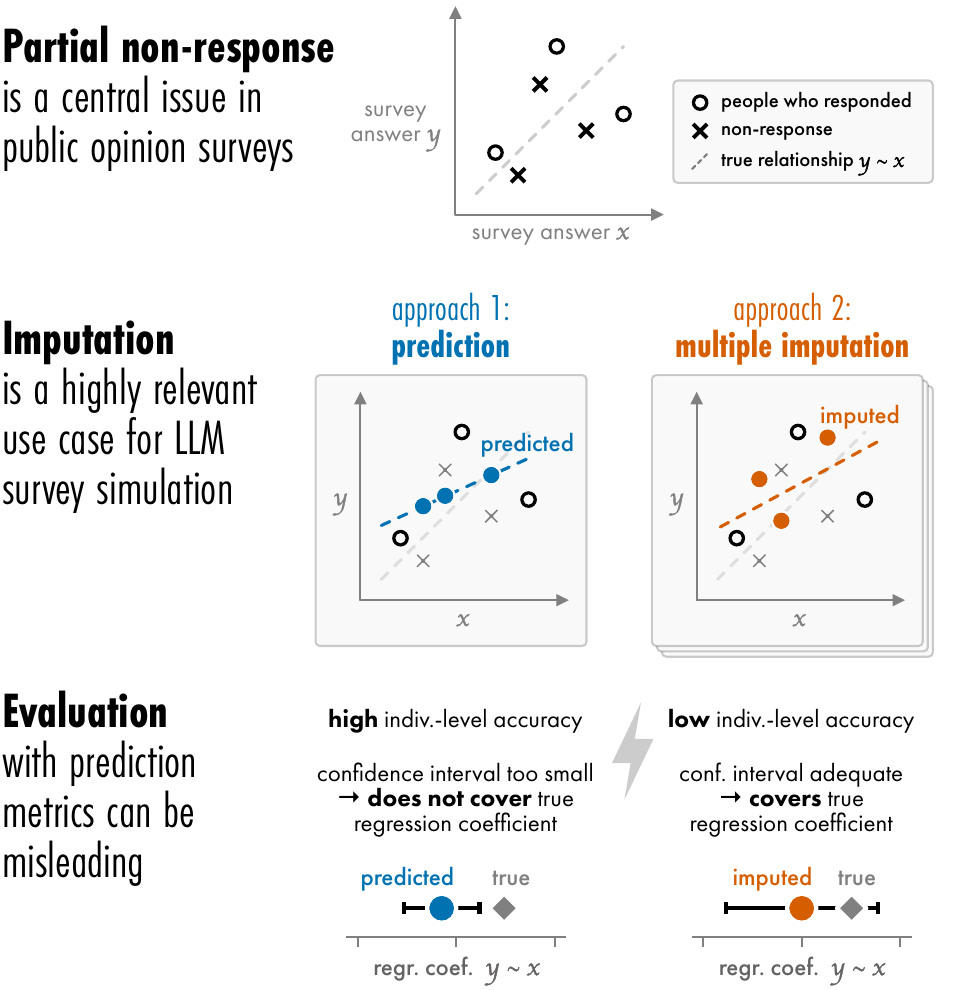}
  \caption{\textbf{Imputation differs from prediction.} The evaluation of methods that fill-in (\textit{impute}) missing survey responses is well-established and focuses on valid uncertainty estimation under different missingness settings. We evaluate LLM survey simulation for imputation and show that prediction metrics are misleading.}
  \label{fig:figure_1}
\end{figure}

Imputation fills in (\textit{imputes}) the missing values so that analyses of the completed data yield valid inferences, that is, approximately unbiased estimates with confidence intervals that reflect the uncertainty due to missingness~\citep{little2019}. If left untreated or treated incorrectly, missing data can lead to biased estimates and reduced statistical efficiency \citep{hansen1946problem,little1982models}. Therefore, \textbf{imputation is a highly relevant evaluation setting for LLM-based survey simulations.} However, prior work has mainly evaluated how well LLMs predict individual answers or group-level answer distributions, with metrics such as F1 or total variation distance (TVD). Imputation differs from prediction: it must preserve the joint distribution of the variables so that downstream analyses remain valid, and evaluating it by accuracy alone favors overly deterministic imputations that understate uncertainty~\citep{rubin1987, van2018flexible}, as Figure~\ref{fig:figure_1} illustrates.

In this paper, \textbf{we assess LLM-based imputation of missing survey responses}. We impute over 5 million survey responses across 4 public opinion surveys from the U.S. and Germany, and compare 6 open-weight LLMs against statistical imputation methods (MICE PMM, MICE Forest) and a tabular foundation model (TabPFN-3). We make five main findings:

\begin{enumerate}[topsep=5pt, itemsep=3pt, parsep=4pt]
\setlength{\abovedisplayskip}{2pt}
\setlength{\belowdisplayskip}{3pt}
\setlength{\abovedisplayshortskip}{2pt}
\setlength{\belowdisplayshortskip}{3pt}
    \item Previous work ignores central aspects in evaluating imputation: non-random missingness, joint distributions, or statistical baselines (§\ref{sec:related_work}).
    \item Imputation with in-context learning (ICL) performs best with similar cases as examples and with verbalized distribution output (§\ref{sec:config_selection}).
    \item While ICL-based imputation significantly outperforms zero-shot LLMs, it remains, at best, competitive with statistical baselines (§\ref{sec:comparison}).
    \item Unlike statistical methods, ICL-based imputation is sensitive to survey topics (§\ref{par:survey_topics}).
    \item Common prediction metrics such as TVD are poor guides to imputation quality in LLM survey simulation (§\ref{par:metrics_config}, §\ref{par:metrics_methods}).
\end{enumerate}

Our findings show a clear mismatch between prediction and imputation performance in LLM survey simulations. We therefore recommend that LLM survey simulation be \textbf{evaluated for specific applications using fit-for-purpose evaluation methods} that build on established practices and include statistical baselines. To facilitate further research, we publish a Python package with a sklearn-like API\footnote{\url{https://anonymous.4open.science/r/icl-survey-imputation/}}.

\section{Imputation Background}
The implications of missing data depend critically on the underlying missingness mechanism~\citep{Rubin1976}, which influences the complexity and feasibility of different imputation approaches (Section~\ref{sec:simulation_study_design}). Multiple imputation (MI) has become the standard method to deal with missing data \citep{rubin1987, little2019}; the standard package \texttt{mice} \cite{van2011mice} has collected over 25000 citations. It generates several completed datasets that reflect uncertainty about the missing values and combines their results to produce standard errors and confidence intervals that propagate this uncertainty. Predictive Mean Matching \citep[PMM;][]{little1988, allison2015imputation} combines regression-based predictions with donor-based substitution to ensure that imputed values remain realistic~\citep{little1988}.

\subsection{Missingness mechanisms} \label{subsec:missingness_mech}

The imputation literature discerns three general missingness mechanisms. Under \textbf{missing completely at random (MCAR)}, values are removed uniformly at random. Under \textbf{missing at random (MAR)}, missingness depends on an observed variable (e.g. political ideology). Under \textbf{missing not at random (MNAR)}, missingness depends on the unobserved value of $v_o$ itself. MCAR and MAR are \emph{ignorable}~\citep{Rubin1976}: the observed data suffice for valid inference, without modeling why values are missing. MNAR is non-ignorable, so methods that rely solely on observed data, including standard multiple imputation, can be biased. Note that ignorability concerns the imputation model, not necessarily the analysis model: because MNAR missingness depends only on the regressor of the analysis model, complete-case estimates of $\beta$ remain unbiased if the analysis model is linear \cite{whiteMultipleImputationUsing2011}. MNAR is nevertheless challenging for imputation, since methods that impute $v_o$ from the observed data assume missingness at random given the auxiliary variables.

\section{Related Work}
\label{sec:related_work}

Within the broader LLM survey simulation literature, a subset of studies focuses specifically on the imputation of missing data in public opinion surveys~\citep{kim_ai-augmented_2026, ji2024predicting, holtdirk2025learningconveniencesamplescase, zhaoLargeLanguageModels2025, suh-etal-2026-graph}.
However, these studies lack established statistical imputation methods as baselines~\citep{van2011mice, wilson2022miceforest} or depart from standard evaluation practices for data imputation~\citep{schoutenGeneratingMissingValues2018, van2018flexible}, which require assessing whether a method preserves marginal and joint distributions in the underlying data and adequately reflects the uncertainty associated with missing values~\citep{Rubin1976}.

Instead, all of these studies primarily evaluate individual-level prediction accuracy, in addition to the omissions detailed below. \citet{kim_ai-augmented_2026} showed that fine-tuned LLMs outperform MICE PMM across various missingness mechanisms, including MNAR, when measured with prediction accuracy (AUC), but do not evaluate distributions or model uncertainty. \citet{holtdirk2025learningconveniencesamplescase} found that fine-tuned models surpass both traditional classifiers and zero-shot approaches on biased samples, but omit comparisons to statistical imputation methods. \citet{ji2024predicting} applied a retrieval-augmented generation (RAG) framework with competitive or superior results over conventional imputation, but evaluate only on a single missingness mechanism, leaving it unclear how the approach generalizes to MNAR. \citet{zhaoLargeLanguageModels2025} compared zero-shot prompts to a few-shot setting with a fixed set of examples, but did not differentiate missingness mechanisms and did not include statistical baselines. \citet{suh-etal-2026-graph} find that GNNs can outperform LLMs in imputing missing survey responses, but only consider MCAR and individual-level accuracy.

A parallel line of work fine-tunes LLMs on existing survey data, leveraging responses from related surveys or held-out participants, before predicting responses of unseen individuals~\citep{suh2025language, cao2025specializing, ramezani2023knowledge, krsteski_valid_2025}.
We focus specifically on missing-data imputation and evaluate in-context learning as a complement to fine-tuning, following standard practice for imputation evaluation and including relevant statistical baselines.

\section{Experimental Setup}

We use three disjoint sets of survey data, each for one purpose. The first two come from the Pew Research Center's American Trends Panel (ATP, Appendix~\ref{app:dataset}), as provided in the OpinionQA dataset \citep{santurkarWhoseOpinionsLanguage2023}. On one ATP wave (W92 with 10 opinion variables), we select the ICL configuration and tune the hyperparameters of the statistical baselines (Section~\ref{sec:config_selection}). On 140 opinion variables from 14 further ATP waves, we compare the selected configuration with the baselines (Section~\ref{sec:comparison}). Finally, we repeat this comparison on two German election surveys (GLES 2021 and 2025) and a U.S.\ health survey (BRFSS 2024), with the latter two collected after gpt-oss-120b's training cutoff (Appendix~\ref{app:gles_brfss}).

\subsection{Simulation Study Design}\label{sec:simulation_study_design}

\begin{figure}[t!]
  \centering
  \includegraphics[width=\linewidth]{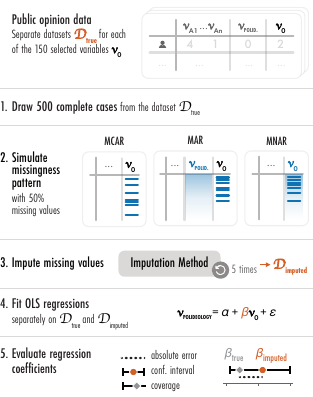}
  \caption{\textbf{Simulation study design} for one opinion variable. We repeat these steps under five missingness settings: MCAR, and left- and right-tailed MAR and MNAR.}
  \label{fig:evaluation_overview}
\end{figure}

Each ATP wave includes 12 demographic variables, among them political ideology $v_{\text{POLIDEOLOGY}}$, and a wave-specific set of opinion variables $V_O$. For each opinion variable $v_o \in V_O$, we construct a dataset of $v_o$, $v_{\text{POLIDEOLOGY}}$, and the other demographic variables, which serve as auxiliary variables $V_A$ for the imputation. Listwise deletion yields the complete-case pool $D_{\text{true}}$ (Figure~\ref{fig:evaluation_overview}). From $D_{\text{true}}$, we sample 500 respondents without replacement to obtain $D_{\text{full}}$, introduce missingness into $v_o$ to obtain $D_{\text{missing}}$, and impute it to obtain $D_{\text{imputed}}$, or $m$ completed datasets for multiple imputation. All missingness settings of a variable share the same sample $D_{\text{full}}$, and we do not use survey weights.

We simulate the three missingness mechanisms described in §\ref{subsec:missingness_mech}: MCAR, MAR, and MNAR. We remove 50\% of the values of $v_o$, a common choice in the imputation literature~\citep{van2018flexible, morrisUsingSimulationStudies2019}, with the amputation procedure of \citet{schoutenGeneratingMissingValues2018}. For MAR and MNAR, we vary the direction of dependence: in the \textit{right-tailed} variant, respondents with higher values of the variable on which missingness depends are more likely to be missing, and in the \textit{left-tailed} variant, respondents with lower values are. We call each combination of mechanism and tail a \emph{missingness setting} (five in total) and one opinion variable under one missingness setting a \emph{condition}.

\paragraph{Opinion variables.} From each wave, we select 10 opinion variables in two stages. We retain variables with at least 500 complete cases and, among them, select those for which the complete-case regression coefficient is most affected by MNAR amputation (highest absolute bias across 10 repetitions). The latter criterion ensures that ignoring the missingness has consequences: for variables whose complete-case coefficients are barely affected by MNAR, complete-case analysis already performs well, and the comparison would reveal little about how imputation methods handle non-ignorable missingness.

\subsection{Evaluation}\label{sec:evaluation}
For each condition, we fit an ordinary least-squares (OLS) regression of political ideology on the opinion variable, our \emph{analysis model}:
\setlength{\abovedisplayskip}{4pt}
\setlength{\belowdisplayskip}{5pt}
\setlength{\abovedisplayshortskip}{4pt}
\setlength{\belowdisplayshortskip}{5pt}
\[
  v_{\text{POLIDEOLOGY}} = \alpha + \beta \, v_o + \varepsilon,
\]
where both variables are standardized with the mean and standard deviation of $D_{\text{true}}$. It yields the coefficient $\hat\beta$ and its 95\,\% confidence interval. For $m$ completed datasets, we pool the estimates with Rubin's rules \citep{rubin1987,little2019} (Appendix~\ref{app:rubins_rules}). Among the evaluated methods, only MICE PMM draws the parameters of its imputation model, as proper multiple imputation in Rubin's sense requires. The other methods, including the LLMs, therefore do not propagate the uncertainty about the imputation model, which can understate the variance of the pooled estimate~\citep{buuren_llms_2026}.

\paragraph{Evaluation metrics.} We evaluate imputation with the performance measures of simulation studies~\citep{van2018flexible, morrisUsingSimulationStudies2019}, which we call \emph{inference metrics} and introduce briefly because they are less common in NLP. They compare the (pooled) coefficient $\hat\beta_{\text{imp}}$ from the imputed data with the reference coefficient $\hat\beta_{\text{true}}$ estimated on $D_{\text{true}}$, which serves as the population:

\begin{enumerate}[itemsep=5pt, topsep=4pt, parsep=0pt]
  \item \textbf{Absolute error.} The deviation of the imputed coefficient from the reference coefficient:
  \[
    \bigl|\hat\beta_{\text{imp}} - \hat\beta_{\text{true}}\bigr|.
  \]
  \item \textbf{Coverage.} Whether the 95\,\% confidence interval contains the reference coefficient:
  \[
    \mathbf{1}\!\bigl[\,\mathrm{CI}_{\text{lower}} \le \hat\beta_{\text{true}} \le \mathrm{CI}_{\text{upper}}\,\bigr].
  \]
  Its average over conditions is the coverage rate. Because each sample makes up a sizable share of its pool (20\,\% at the median), we apply a finite population correction to all intervals (Appendix~\ref{app:rubins_rules}).
  \item \textbf{CI width.} The precision of the estimate:
  \[
    \mathrm{CI}_{\text{upper}} - \mathrm{CI}_{\text{lower}}.
  \]
  Narrower intervals are only preferable when coverage is adequate, so width does not rank methods on its own.
\end{enumerate}
Our comparisons with prediction metrics therefore use absolute error and coverage (Figures~\ref{fig:study_1_metric_agreement} and~\ref{fig:study_2_prediction_vs_inference}). These \emph{prediction metrics} are computed on the imputed cells and averaged over the $m$ imputations: accuracy and macro-F1 of the imputed answers, and the total variation distance (TVD) and scale-normalized Wasserstein-1 distance (W1) between the imputed and true answer distributions of $v_o$.

\paragraph{Simulation repetitions.} Simulation studies of multiple imputation usually repeat sampling and amputation many times per condition and report bias, which requires such repetitions~\citep{van2018flexible}. Because LLM-based imputation draws on knowledge about specific survey questions, its performance may vary more across variables and survey contexts than that of statistical methods, which are agnostic to question content. We therefore allocate the computational budget to breadth: we draw one sample per variable ($n_{\text{sim}}=1$), report absolute error instead of bias, and aggregate over conditions rather than repetitions. Unless stated otherwise, we summarize each metric by its median over conditions and coverage by the coverage rate. Appendix~\ref{app:variable_level} checks this design with $n_{\text{sim}}=100$ repetitions for two variables.

\section{Configuration Selection}\label{sec:config_selection}

We tailor LLM-based survey simulation to imputation through in-context learning (ICL) and use survey data from ATP wave W92 to identify the best ICL configuration out of 84 plausible ones. We test whether prediction metrics rank the configurations as inference metrics do.

\subsection{Setup}

We impute each missing value of $v_o$ by prompting an LLM with $k$ complete cases from the same dataset as in-context examples (Appendix Figure~\ref{fig:simulation_setup}). Respondents are serialized as text and embedded with EmbeddingGemma-300M \citep{veraEmbeddingGemmaPowerfulLightweight2025}, and examples are retrieved by cosine similarity. Because LLM predictions are often overconfident \citep{xiong2024can,bisbee2024synthetic,boelaert2025machine}, we compare configurations that vary which examples enter the context and how the model reports its answer, aiming for imputations that better reflect uncertainty. With each of Qwen3-8B and Qwen3-30B-A3B-Instruct-2507 in its FP8-quantized release (hereafter Qwen3-30B-A3B, \citealp{qwen3technicalreport}), we compare 84 configurations, crossing four retrieval strategies (\textit{Most Similar}, \textit{Least Similar}, \textit{Random}, \textit{Diverse}), two serializations (\textit{String} or \textit{List}, i.e., one embedding per respondent or per variable), retrieval with or without stratification by answer option, three prompt formats (\textit{Multi-turn}, \textit{Single-turn}, \textit{Single-turn verbalized}), and $k=10$ or $k=100$ in-context examples (Appendix~\ref{app:icl}). The \textit{Single-turn verbalized} prompt asks the model for a verbalized probability distribution over the answer options~\citep{meister-etal-2025-benchmarking, ahnert-etal-2026-survey}, from which we draw the imputations.

\begin{table*}[t!]
  \centering
  \small
  {
    \setlength{\extrarowheight}{3.5pt}
    \renewcommand{\arraystretch}{0.8}
  \begin{tabular}{lrrr@{\hspace{1.8em}}rrr@{\hspace{1.8em}}rrr}
    \toprule
     & \multicolumn{3}{c}{\textbf{MCAR}} & \multicolumn{3}{c}{\textbf{MAR}} & \multicolumn{3}{c}{\textbf{MNAR}} \\
    \cmidrule(r{1.8em}){2-4}\cmidrule(r{1.8em}){5-7}\cmidrule{8-10}
    \textbf{Method} & \shortstack{Abs.\\Error $\downarrow$} & \shortstack{CI\\Width $\downarrow$} & \shortstack{Cover-\\age $\uparrow$} & \shortstack{Abs.\\Error $\downarrow$} & \shortstack{CI\\Width $\downarrow$} & \shortstack{Cover-\\age $\uparrow$} & \shortstack{Abs.\\Error $\downarrow$} & \shortstack{CI\\Width $\downarrow$} & \shortstack{Cover-\\age $\uparrow$} \\
\midrule
\textit{Full Data} & 0.026 & 0.156 & 0.964 & 0.026 & 0.156 & 0.964 & 0.026 & 0.156 & 0.964 \\
\textit{Complete Case} & 0.040 & 0.231 & 0.957 & 0.046 & 0.213 & 0.914 & 0.062 & 0.254 & 0.886 \\
\midrule
\multicolumn{10}{l}{\textit{Zero-Shot LLM}} \\
Qwen3-30B-A3B & {\cellcolor[HTML]{FFF3AC}} \color[HTML]{000000} 0.064 & {\cellcolor[HTML]{05713C}} \color[HTML]{F1F1F1} 0.183 & {\cellcolor[HTML]{FB9D59}} \color[HTML]{000000} 0.636 & {\cellcolor[HTML]{E44C34}} \color[HTML]{F1F1F1} 0.081 & {\cellcolor[HTML]{16914D}} \color[HTML]{F1F1F1} 0.189 & {\cellcolor[HTML]{EF633F}} \color[HTML]{F1F1F1} 0.596 & {\cellcolor[HTML]{FEEDA1}} \color[HTML]{000000} 0.065 & {\cellcolor[HTML]{4BB05C}} \color[HTML]{F1F1F1} 0.196 & {\cellcolor[HTML]{FED683}} \color[HTML]{000000} 0.682 \\
gpt-oss-120b & {\cellcolor[HTML]{5AB760}} \color[HTML]{F1F1F1} 0.045 & {\cellcolor[HTML]{006837}} \color[HTML]{F1F1F1} 0.181 & {\cellcolor[HTML]{E8F59F}} \color[HTML]{000000} 0.763 & {\cellcolor[HTML]{FFF3AC}} \color[HTML]{000000} 0.064 & {\cellcolor[HTML]{05713C}} \color[HTML]{F1F1F1} 0.183 & {\cellcolor[HTML]{FDB365}} \color[HTML]{000000} 0.651 & {\cellcolor[HTML]{8CCD67}} \color[HTML]{000000} 0.049 & {\cellcolor[HTML]{0E8245}} \color[HTML]{F1F1F1} 0.186 & {\cellcolor[HTML]{DAF08D}} \color[HTML]{000000} 0.778 \\
\midrule
\multicolumn{10}{l}{\textit{ICL (k=10)}} \\
Qwen3-30B-A3B & {\cellcolor[HTML]{30A356}} \color[HTML]{F1F1F1} 0.042 & {\cellcolor[HTML]{54B45F}} \color[HTML]{F1F1F1} 0.197 & {\cellcolor[HTML]{7FC866}} \color[HTML]{000000} 0.850 & {\cellcolor[HTML]{E3F399}} \color[HTML]{000000} 0.058 & {\cellcolor[HTML]{219C52}} \color[HTML]{F1F1F1} 0.191 & {\cellcolor[HTML]{FEEDA1}} \color[HTML]{000000} 0.711 & {\cellcolor[HTML]{ADDC6F}} \color[HTML]{000000} 0.052 & {\cellcolor[HTML]{AFDD70}} \color[HTML]{000000} 0.210 & {\cellcolor[HTML]{ADDC6F}} \color[HTML]{000000} 0.818 \\
gpt-oss-120b & {\cellcolor[HTML]{04703B}} \color[HTML]{F1F1F1} 0.036 & {\cellcolor[HTML]{54B45F}} \color[HTML]{F1F1F1} 0.197 & {\cellcolor[HTML]{219C52}} \color[HTML]{F1F1F1} 0.907 & {\cellcolor[HTML]{7FC866}} \color[HTML]{000000} 0.048 & {\cellcolor[HTML]{73C264}} \color[HTML]{000000} 0.201 & {\cellcolor[HTML]{91D068}} \color[HTML]{000000} 0.839 & {\cellcolor[HTML]{8CCD67}} \color[HTML]{000000} 0.049 & {\cellcolor[HTML]{F4FAB0}} \color[HTML]{000000} 0.224 & {\cellcolor[HTML]{42AC5A}} \color[HTML]{F1F1F1} 0.889 \\
\midrule
\multicolumn{10}{l}{\textit{ICL (k=100)}} \\
Qwen3-30B-A3B & {\cellcolor[HTML]{04703B}} \color[HTML]{F1F1F1} 0.036 & {\cellcolor[HTML]{8ECF67}} \color[HTML]{000000} 0.205 & {\cellcolor[HTML]{89CC67}} \color[HTML]{000000} 0.843 & {\cellcolor[HTML]{F8FCB6}} \color[HTML]{000000} 0.061 & {\cellcolor[HTML]{7AC665}} \color[HTML]{000000} 0.202 & {\cellcolor[HTML]{D7EE8A}} \color[HTML]{000000} 0.782 & {\cellcolor[HTML]{98D368}} \color[HTML]{000000} 0.050 & {\cellcolor[HTML]{DCF08F}} \color[HTML]{000000} 0.218 & {\cellcolor[HTML]{87CB67}} \color[HTML]{000000} 0.846 \\
gpt-oss-120b & {\cellcolor[HTML]{006837}} \color[HTML]{F1F1F1} 0.035 & {\cellcolor[HTML]{39A758}} \color[HTML]{F1F1F1} 0.194 & {\cellcolor[HTML]{249D53}} \color[HTML]{F1F1F1} 0.906 & {\cellcolor[HTML]{4BB05C}} \color[HTML]{F1F1F1} 0.044 & {\cellcolor[HTML]{63BC62}} \color[HTML]{F1F1F1} 0.199 & {\cellcolor[HTML]{4EB15D}} \color[HTML]{F1F1F1} 0.882 & {\cellcolor[HTML]{7FC866}} \color[HTML]{000000} 0.048 & {\cellcolor[HTML]{F1F9AC}} \color[HTML]{000000} 0.223 & {\cellcolor[HTML]{5AB760}} \color[HTML]{F1F1F1} 0.875 \\
\midrule
\multicolumn{10}{l}{\textit{Baselines}} \\
Most Similar Embed. & {\cellcolor[HTML]{3CA959}} \color[HTML]{F1F1F1} 0.043 & {\cellcolor[HTML]{006837}} \color[HTML]{F1F1F1} 0.181 & {\cellcolor[HTML]{75C465}} \color[HTML]{000000} 0.857 & {\cellcolor[HTML]{A50026}} \color[HTML]{F1F1F1} 0.089 & {\cellcolor[HTML]{026C39}} \color[HTML]{F1F1F1} 0.182 & {\cellcolor[HTML]{A50026}} \color[HTML]{F1F1F1} 0.514 & {\cellcolor[HTML]{C9E881}} \color[HTML]{000000} 0.055 & {\cellcolor[HTML]{54B45F}} \color[HTML]{F1F1F1} 0.197 & {\cellcolor[HTML]{C7E77F}} \color[HTML]{000000} 0.796 \\
MICE PMM & {\cellcolor[HTML]{0E8245}} \color[HTML]{F1F1F1} 0.038 & {\cellcolor[HTML]{FEEC9F}} \color[HTML]{000000} 0.232 & {\cellcolor[HTML]{006837}} \color[HTML]{F1F1F1} 0.957 & {\cellcolor[HTML]{73C264}} \color[HTML]{000000} 0.047 & {\cellcolor[HTML]{A50026}} \color[HTML]{F1F1F1} 0.272 & {\cellcolor[HTML]{016A38}} \color[HTML]{F1F1F1} 0.954 & {\cellcolor[HTML]{ADDC6F}} \color[HTML]{000000} 0.052 & {\cellcolor[HTML]{C41E27}} \color[HTML]{F1F1F1} 0.266 & {\cellcolor[HTML]{4BB05C}} \color[HTML]{F1F1F1} 0.884 \\
MICE Forest & {\cellcolor[HTML]{C9E881}} \color[HTML]{000000} 0.055 & {\cellcolor[HTML]{279F53}} \color[HTML]{F1F1F1} 0.192 & {\cellcolor[HTML]{C1E57B}} \color[HTML]{000000} 0.800 & {\cellcolor[HTML]{8CCD67}} \color[HTML]{000000} 0.049 & {\cellcolor[HTML]{73C264}} \color[HTML]{000000} 0.201 & {\cellcolor[HTML]{7FC866}} \color[HTML]{000000} 0.850 & {\cellcolor[HTML]{FDBF6F}} \color[HTML]{000000} 0.071 & {\cellcolor[HTML]{D7EE8A}} \color[HTML]{000000} 0.217 & {\cellcolor[HTML]{FFFCBA}} \color[HTML]{000000} 0.732 \\
TabPFN & {\cellcolor[HTML]{128A49}} \color[HTML]{F1F1F1} 0.039 & {\cellcolor[HTML]{AFDD70}} \color[HTML]{000000} 0.210 & {\cellcolor[HTML]{08773F}} \color[HTML]{F1F1F1} 0.943 & {\cellcolor[HTML]{219C52}} \color[HTML]{F1F1F1} 0.041 & {\cellcolor[HTML]{DCF08F}} \color[HTML]{000000} 0.218 & {\cellcolor[HTML]{48AE5C}} \color[HTML]{F1F1F1} 0.886 & {\cellcolor[HTML]{73C264}} \color[HTML]{000000} 0.047 & {\cellcolor[HTML]{FEEC9F}} \color[HTML]{000000} 0.232 & {\cellcolor[HTML]{70C164}} \color[HTML]{000000} 0.861 \\
\bottomrule
\end{tabular}

  }
  \caption{\textbf{In-context imputation versus baseline methods} on 140 opinion variables from 14 ATP waves. ICL uses the selected configuration with $k=10$ or $k=100$ in-context examples. Values summarize up to 140 conditions under MCAR and 280 under MAR and MNAR. Cell colors indicate relative performance within each metric (\textcolor[HTML]{006837}{green} = better, \textcolor[HTML]{A50026}{red} = worse).}
  \label{tab:study_2_table}
\end{table*}

\subsection{Results}

Qwen3-30B-A3B is the stronger of the two models. Its best configurations reach coverage rates of up to 0.72, whereas no Qwen3-8B configuration exceeds 0.63, and the best median absolute errors are similar (0.067 for Qwen3-30B-A3B and 0.060 for Qwen3-8B). We therefore select the configuration with Qwen3-30B-A3B and focus the following analyses on it. For Qwen3-30B-A3B, the prompt format mattered most. The \textit{Single-turn verbalized} prompt gave among the lowest absolute errors with both 10 and 100 in-context examples, especially with \textit{Most Similar} retrieval, whereas the other retrieval choices mattered mainly in combination with the prompt format (Appendix Figures~\ref{fig:study_1_abs_error_appendix}--\ref{fig:study_1_width_appendix}). Based on absolute error, coverage, and width, we select the \textit{Single-turn verbalized} prompt with \textit{Most Similar} retrieval, \textit{String} serialization, and no stratification. With 100 in-context examples, it has the lowest median absolute error of all 84 configurations (0.067) and the second-highest coverage rate. For Qwen3-8B, the \textit{Single-turn verbalized} prompt is less effective. We keep both 10 and 100 in-context examples for the main comparison.

\begin{figure}[t!]
  \centering
  \includegraphics[width=\linewidth]{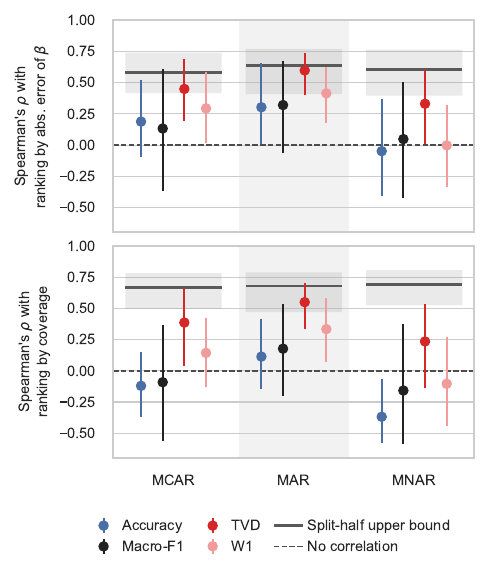}
  \caption{\textbf{Across ICL configurations, prediction metrics agree only weakly with inference metrics.} Spearman's rank correlation between the rankings of the 84 Qwen3-30B-A3B configurations by a prediction metric (colors) and by an inference metric (rows). The prediction metric is computed on a random half of the conditions and the inference metric on the other half (bars span 95\% of 200 random splits). The split-half upper bound is the correlation of the inference metric with itself across the two halves. Prediction metrics rest on one imputation per run (Appendix~\ref{app:pred_inference}).}
  \label{fig:study_1_metric_agreement}
\end{figure}

\paragraph{Do prediction metrics rank configurations like inference metrics?}\label{par:metrics_config} We test whether ranking the 84 configurations by the prediction quality of their imputations reproduces their ranking by inference quality (Figure~\ref{fig:study_1_metric_agreement}). With Qwen3-30B-A3B, the rankings by accuracy and macro-F1 correlate with those by absolute error and coverage at 0.32 at most (Spearman's $\rho_s$), well below the split-half upper bounds of 0.58--0.69, and under MNAR the correlations are near zero or negative. TVD comes closest and approaches the upper bound under MAR but not under MNAR. The low correlation is not due to differences between variables, since it persists within single conditions (Appendix~\ref{app:pred_inference}). It partly reflects the difference between prediction and imputation: accuracy and macro-F1 reward predicting the most likely answer, whereas the verbalized prompt draws each imputation from a distribution. For Qwen3-8B, for which the verbalized prompt is less effective, accuracy and macro-F1 do track the inference metrics (Appendix~\ref{app:pred_inference}). Whether prediction metrics rank configurations like inference metrics thus depends on the generator.

\section{Comparison with Statistical Imputation}\label{sec:comparison}

We now compare the selected configuration with tuned statistical imputation methods on 140 opinion variables from 14 further ATP waves, as well as two German election surveys (GLES 2021/2025) and a U.S.\ health survey (BRFSS 2024).

\subsection{Setup}

\paragraph{Baselines.}\label{sec:baseline_methods} \textbf{MICE PMM} \citep{van2011mice}, our primary statistical baseline, is multiple imputation by Predictive Mean Matching in the R \texttt{mice} package. \textbf{MICE Forest} \citep{wilson2022miceforest} is a LightGBM-based variant of MICE (\texttt{miceforest} Python package). \textbf{TabPFN} (TabPFN-3, \citealp{priorlabs2026tabpfn3}) is a pretrained tabular foundation model. We fit its classifier on the observed rows of $v_o$ with all other variables as features and sample each missing value from its predicted class probabilities (temperature 1.0). We tune all three on W92 (Appendix~\ref{app:baseline_tuning}). All imputation models, including the LLM prompts, condition on political ideology, the outcome of the analysis model. \textbf{Full Data} analyzes the sample before amputation and serves as a reference, and \textbf{Complete Case} is complete-case analysis (listwise deletion). To disentangle retrieval from generation, \textbf{Zero-Shot LLM} uses the same generator and prompt format without in-context examples ($k=0$, Appendix Figure~\ref{fig:prompt_zero_shot}), and \textbf{Most Similar Embeddings} uses the same retrieval ($k=10$) but imputes the modal answer among the retrieved cases instead of querying an LLM (single imputation). All multiple-imputation methods use $m=5$. Two simple baselines, Mode Imputation and Random Sample, are reported in Appendix Table~\ref{tab:study_2_table_appendix}.

\paragraph{Generators.} We compare Qwen3-30B-A3B and the reasoning-tuned gpt-oss-120b \citep{openai2025gptoss120bgptoss20bmodel}, both with the selected configuration and $k=10$ or $k=100$ retrieved in-context examples (10-shot and 100-shot ICL). For robustness across models, we additionally evaluate Qwen3-8B, Qwen3-30B-A3B-Thinking-2507 (FP8-quantized release) \citep{qwen3technicalreport}, Olmo-3-7B-Instruct \citep{olmo2025olmo3}, and GLM-4.7-Flash \citep{5team2025glm45agenticreasoningcoding} (Appendix Table~\ref{tab:study_2_table_appendix}). For the two reasoning models, gpt-oss-120b and Qwen3-30B-A3B-Thinking, we generate one completion per missing value and draw all $m$ imputations from its stated distribution to limit cost. For the other models, each imputation comes from a separate completion.

\paragraph{Statistical tests.} For absolute error, we test method differences separately for each mechanism with Friedman tests blocked by opinion variable and tail, followed by Holm-corrected pairwise Wilcoxon signed-rank tests. For coverage, we use Cochran's $Q$ test and Holm-corrected pairwise McNemar tests (Appendix Tables~\ref{tab:sig-abs-error-mcar}--\ref{tab:sig-coverage-mnar}).

\subsection{Results}

\paragraph{Absolute error.}\label{par:abs_error} The strongest ICL variant, gpt-oss-120b with 100-shot ICL, has a median absolute error within 0.004 of MICE PMM and TabPFN under every mechanism (Table~\ref{tab:study_2_table}), and none of its differences to MICE PMM survives the Holm correction (paired Wilcoxon, smallest Holm $p=.07$, under MNAR). The Qwen3-30B-A3B variants fall behind MICE PMM and TabPFN under MAR. MICE Forest is significantly outperformed by all four ICL variants under MNAR, but by none under MCAR or MAR. Retrieval significantly improves over zero-shot prompting for Qwen3-30B-A3B under all three mechanisms, but for gpt-oss-120b only under MAR.

\paragraph{Coverage and width.}\label{par:coverage_width} MICE PMM keeps coverage at the nominal level under MCAR and MAR and drops to 0.88 under MNAR (Table~\ref{tab:study_2_table}). The intervals of gpt-oss-120b with 100-shot ICL are 16--27\,\% narrower but have lower coverage under every mechanism (0.88--0.91), significantly so under MAR (McNemar, Holm $p=.03$). TabPFN lies between the two under MCAR and below both under MNAR. The Qwen3-30B-A3B variants have lower coverage still (0.71--0.85). Narrower intervals thus come at the cost of coverage.

\paragraph{Overall.}\label{par:overall} Even with gpt-oss-120b, in-context imputation does not outperform the tuned statistical baselines: MICE PMM and TabPFN attain similar errors, and MICE PMM has higher coverage under MCAR and MAR. The four additional generators do not change this picture, and in-context examples improve over zero-shot prompting for all of them (Appendix Table~\ref{tab:study_2_table_appendix}).

\begin{figure}[t!]
  \centering
  \includegraphics[width=\linewidth]{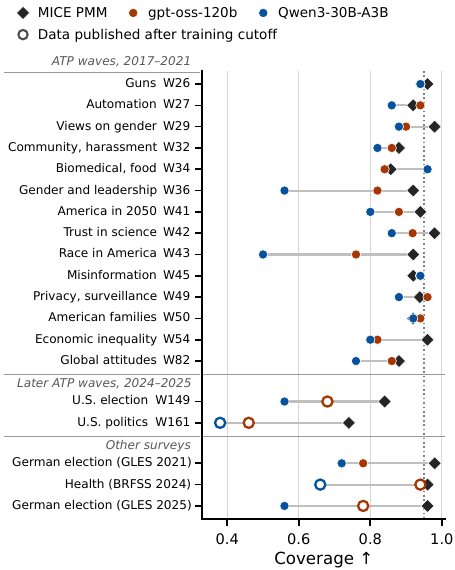}
  \caption{\textbf{The coverage of in-context imputation varies with the survey topic, whereas that of MICE PMM varies less.} Coverage rate per survey over 50 conditions (10 variables under five missingness settings). Within each group, the surveys are in the order in which they were fielded. The LLMs use 100-shot ICL. The 14 ATP waves are those of the main comparison. Hollow markers indicate that the survey's data were published after the model's training cutoff (gpt-oss-120b) or, since no cutoff is documented for Qwen3-30B-A3B, its release. The dotted line marks the nominal coverage of 0.95.}
  \label{fig:coverage_by_survey}
\end{figure}

\paragraph{Survey topics.}\label{par:survey_topics} The coverage of in-context imputation varies with the survey topic (Figure~\ref{fig:coverage_by_survey}). Across the 14 ATP waves, the coverage of MICE PMM never falls below 0.86, whereas that of 100-shot ICL falls to 0.76 with gpt-oss-120b and to 0.50 with Qwen3-30B-A3B, with the lowest values on W43 (race in America) and W36 (gender and leadership, see also Appendix~\ref{app:variable_level}). The same holds on two German election surveys (GLES) and a U.S.\ health survey (BRFSS, Appendix~\ref{app:gles_brfss}): MICE PMM reaches at least 0.96, whereas gpt-oss-120b drops to 0.78 on both GLES surveys and Qwen3-30B-A3B to 0.72 or lower on all three. On two later ATP waves on U.S.\ politics (W149 and W161), coverage is lower for all three methods and in-context imputation falls further behind MICE PMM. The lower coverage of MICE PMM stems from a more severe MNAR setting on these waves, as the results by missingness mechanism show (Appendix~\ref{app:gles_brfss}).

\begin{figure}[t!]
  \centering
  \includegraphics[width=\linewidth]{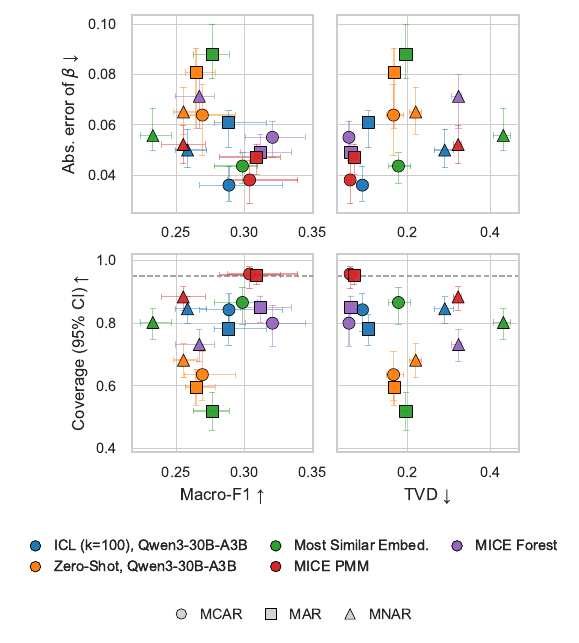}
  \caption{\textbf{Across imputation methods, prediction quality and inference quality are only loosely related.} Each marker is one method under one mechanism, with 95\% bootstrap CIs (Wilson CIs for coverage). Arrows mark the better direction of each metric, and the dashed line marks the nominal coverage of 0.95. If prediction metrics tracked inference quality, better prediction would go with better inference in every panel. Appendix~\ref{app:pred_inference} reports rank correlations for ten methods and further metrics.}
  \label{fig:study_2_prediction_vs_inference}
\end{figure}

\paragraph{Would prediction metrics have selected the same methods?}\label{par:metrics_methods} Figure~\ref{fig:study_2_prediction_vs_inference} relates absolute error and coverage to macro-F1 and TVD on the same runs. MICE Forest has the highest macro-F1 under MCAR and MAR and nearly the highest under MNAR, yet among the largest absolute errors under MCAR and MNAR and a coverage of only 0.73--0.85. Across the ten imputation methods of Appendix~\ref{app:pred_inference}, TVD and macro-F1 rank the methods similarly to absolute error under MAR (method-level $\rho_s = 0.90$ and 0.83) but not under MNAR (0.08 and 0.01). When missingness depends on the answer itself, every method that learns from the observed answers reproduces their shifted distribution, so zero-shot prompting attains the lowest TVD while having among the lowest coverage. Ranking imputation methods by prediction accuracy would therefore not have selected the methods that best support inference, and ranking them by distributional fidelity fails under non-ignorable missingness.

\section{Discussion}

Our results have two main implications. First, \textbf{prediction metrics are a poor guide to the quality of imputations.} Accuracy and macro-F1 reward predicting the most likely answer, whereas valid inference requires imputations that reflect uncertainty. Selecting by prediction metrics can therefore favor configurations and methods that yield biased estimates or intervals that undercover. Distributional metrics such as TVD fare better but fail under MNAR, where the observed answers are themselves a shifted sample. Evaluations of LLM survey simulations that target downstream analyses should therefore report inference metrics directly.

Second, \textbf{in-context imputation with a strong LLM is competitive with, but not better than, established imputation methods.} With gpt-oss-120b and 100-shot ICL, the median absolute error is within 0.004 of MICE PMM and TabPFN.
Where ICL falls short, the cause is bias rather than variance: In our repetition check for two variables with Qwen3-30B-A3B (Appendix~\ref{app:variable_level}), the intervals are not too narrow for the sampling variability of the estimates, and the undercoverage stems from bias. Finally, ICL is far more expensive: a run takes 1--7\,s with the statistical baselines on local machines but about 15 minutes with gpt-oss-120b and 100-shot ICL on H100 GPUs (Appendix~\ref{app:walltime}).
Under MNAR, MAR-based imputation rests on an assumption the data cannot verify; LLM imputation rests on an external prior that is equally unverifiable. LLMs may thus be better suited to specifying departures from MAR in sensitivity analyses than to primary analyses. Given the topic-dependent bias we observe (Appendix~\ref{app:variable_level}), such scenarios should complement MAR-based estimates.

For practitioners, \textbf{established multiple imputation methods such as MICE PMM therefore remain the better default.} In our evaluation, they match the best LLM configuration at a small fraction of the cost and keep coverage closer to the nominal level. Their coverage is also stable across survey topics, whereas that of ICL is not (Figure~\ref{fig:coverage_by_survey}). For research on LLM survey simulation, our results show that LLMs become competitive with statistical methods on an established survey task only with a setup tailored to that task. Without in-context examples, as in the persona prompting used for silicon samples, the same LLMs have mostly larger errors and a coverage rate of only 0.60--0.78 (Table~\ref{tab:study_2_table}).

In conclusion, our findings show a clear mismatch between prediction and imputation performance in LLM survey simulations. Given the high relevance of imputation in applied survey research, we recommend that LLM survey simulations be \textbf{evaluated with fit-for-purpose evaluation methods} on this, and similar, specific applications in the survey lifecycle.

\section*{Limitations}

Our setup probes only a narrow slice of what imputation should preserve. The estimand is a single bivariate regression coefficient ($v_{\text{POLIDEOLOGY}} \sim v_o$), and missingness affects one variable at a time rather than following the multivariate patterns of real surveys. This design isolates imputation performance and is standard in simulation studies~\citep{van2018flexible, morrisUsingSimulationStudies2019}, but future work should cover other estimands and multivariate missingness.

The main comparison uses a single U.S.\ survey panel (the ATP), and the additional surveys add only one other country and language (Germany) and one other survey domain (health, Appendix~\ref{app:gles_brfss}). We draw one sample per variable ($n_{\text{sim}}=1$) and check this design with repetitions for only two variables (Appendix~\ref{app:variable_level}).

Larger LLMs might improve ICL further, but we focus on open-weight LLMs with up to 120B parameters, which researchers can run locally to keep sensitive survey data private.

We do not evaluate hybrid approaches that combine LLM imputations with debiasing estimators such as prediction-powered inference \citep[PPI;][]{angelopoulos2023ppi} and related frameworks \citep{egami2024dsl, broska2025mixed, byun2025synthetic, krsteski_valid_2025}, which could correct the bias behind ICL's undercoverage. Their guarantees, however, typically require a representative labeled subset, and PPI remains biased under MNAR \citep{song2026demystifying}.

\section*{Acknowledgements}

This work is supported by the DAAD programme Konrad Zuse Schools of Excellence in Artificial Intelligence, sponsored by the Federal Ministry of Research, Technology and Space.

\bibliography{main}

\clearpage

\appendix

\raggedbottom
\makeatletter
\setlength{\@fptop}{0pt}
\setlength{\@dblfptop}{0pt}
\makeatother
\renewcommand{\topfraction}{0.9}
\renewcommand{\dbltopfraction}{0.9}
\renewcommand{\bottomfraction}{0.8}
\renewcommand{\textfraction}{0.07}
\renewcommand{\floatpagefraction}{0.8}
\renewcommand{\dblfloatpagefraction}{0.8}
\setcounter{topnumber}{3}
\setcounter{dbltopnumber}{3}
\setcounter{totalnumber}{6}

\FloatBarrier
\section{Dataset}\label{app:dataset}

We use the OpinionQA dataset \citep{santurkarWhoseOpinionsLanguage2023}, which provides human survey responses from the Pew Research Center's American Trends Panel (ATP). The ATP is a nationally representative panel of U.S.\ adults recruited through random sampling of residential addresses. Each survey wave covers a specific topic and was administered online between April 2017 and July 2021.

\paragraph{Waves and opinion variables.}
We draw on 15 ATP waves (W26--W92), spanning topics from guns and gender to economic inequality and political typology. From each wave, we select 10 opinion variables based on the two-stage screening described in Section~\ref{sec:simulation_study_design}: variables must have at least 500 complete cases after listwise deletion and must be the most sensitive to MNAR amputation. This yields 150 opinion variables in total. The configuration selection uses the 10 variables from W92, and the main comparison uses the remaining 140 variables from the other 14 waves. Table~\ref{tab:dataset_waves} provides per-wave sample sizes and topics. Raw wave sizes range from 2,524 (W41) to 10,221 (W92) respondents; after listwise deletion on the 12 demographic variables and a given opinion variable, median complete-case counts range from 1,090 to 9,370 across waves. The selected opinion variables have between 2 and 6 answer options (median 4). Following \citet{santurkarWhoseOpinionsLanguage2023}, we encode each response option using the integer ordering supplied with the OpinionQA dataset.

\begin{table}[h]
\centering
\small
\resizebox{\columnwidth}{!}{\begin{tabular}{@{}llrrrr@{}}
\toprule
Wave & Topic & $N_{\text{raw}}$ & $\widetilde{N}_{\text{cc}}$ & $N_{\text{cc}}^{\min}$ & $|V_O|$ \\
\midrule
W26 & Guns & 4,168 & 1,973 & 1,278 & 10 \\
W27 & Automation and driverless vehicles & 4,135 & 3,957 & 1,962 & 10 \\
W29 & Views on gender & 4,867 & 2,949 & 2,229 & 10 \\
W32 & Community types, Sexual harassment & 6,251 & 4,387 & 696 & 10 \\
W34 & Biomedical and food issues & 2,537 & 2,354 & 1,185 & 10 \\
W36 & Gender and leadership & 4,587 & 1,090 & 1,030 & 10 \\
W41 & Views of America in 2050 & 2,524 & 2,293 & 1,132 & 10 \\
W42 & Trust in science & 4,464 & 2,034 & 1,987 & 10 \\
W43 & Race in America & 6,637 & 3,847 & 2,562 & 10 \\
W45 & Misinformation & 6,127 & 5,613 & 2,815 & 10 \\
W49 & Privacy and surveillance & 4,272 & 1,941 & 1,923 & 10 \\
W50 & American families & 9,834 & 7,453 & 1,862 & 10 \\
W54 & Economic inequality & 6,878 & 3,517 & 3,037 & 10 \\
W82 & 2021 Global Attitudes Project U.S. survey & 2,596 & 2,380 & 1,187 & 10 \\
W92 & Political typology & 10,221 & 9,370 & 9,302 & 10 \\
\midrule
\textbf{Total} & 15 waves & 80,098 & --- & --- & 150 \\
\bottomrule
\end{tabular}}
\caption{\textbf{Survey waves used in the experiments.} $N_{\text{raw}}$: total respondents in the wave (before any filtering). $\widetilde{N}_{\text{cc}}$ and $N_{\text{cc}}^{\min}$: median and minimum number of complete cases across selected opinion variables after listwise deletion on all 12 demographic variables and the opinion variable. $|V_O|$: number of opinion variables selected for the experiments. All waves are from the Pew Research Center's American Trends Panel.}
\label{tab:dataset_waves}
\end{table}

\paragraph{Demographic variables.}
Each wave shares a common set of 12 demographic variables that serve as auxiliary variables $V_A$ (and, in the case of political ideology, as the outcome of the analysis model). Table~\ref{tab:dataset_demographics} lists these variables with their types and response categories. Demographic values are consistent across waves.

\begin{table*}[t]
\centering
\small
\begin{tabular}{@{}lllrl@{}}
\toprule
Variable & Code & Type & $K$ & Values \\
\midrule
Region & \texttt{CREGION} & categorical & 4 & Northeast, Midwest, South, West \\
Gender & \texttt{SEX} & categorical & 2 & Male, Female \\
Age & \texttt{AGE} & ordinal & 4 & 18-29, 30-49, 50-64, 65+ \\
Education & \texttt{EDUCATION} & ordinal & 6 & Less than high school, \ldots, Postgraduate \\
Citizenship & \texttt{CITIZEN} & categorical & 2 & Yes, No \\
Marital Status & \texttt{MARITAL} & categorical & 5 & Married, Divorced, Separated, Widowed, Never been married \\
Religion & \texttt{RELIG} & categorical & 12 & Protestant, \ldots, Nothing in particular \\
Religious Attendance & \texttt{RELIGATTEND} & ordinal & 6 & More than once a week, \ldots, Never \\
Political Party & \texttt{POLPARTY} & categorical & 4 & Republican, Democrat, Independent, Other \\
Income & \texttt{INCOME} & ordinal & 5 & Less than \$30k, \$30k-\$50k, \$50k-\$75k, \$75k-\$100k, {>}\$100k \\
Political Ideology & \texttt{POLIDEOLOGY} & ordinal & 5 & Very conservative, Conservative, Moderate, Liberal, Very liberal \\
Race & \texttt{RACE} & categorical & 5 & White, Black, Asian, Hispanic, Other \\
\bottomrule
\end{tabular}

\caption{\textbf{Demographic variables} shared across all waves. Code: original column name in the dataset. Type: ordinal variables are mapped to numeric values $\{1, \ldots, K\}$ for the tabular methods, and categorical variables are one-hot encoded. $K$: number of response categories.}
\label{tab:dataset_demographics}
\end{table*}

\paragraph{Preprocessing.}
We preprocess the data in two ways:
\begin{itemize}
  \item \textbf{LLM-based methods.} Column headers are replaced with natural-language question text (e.g., \texttt{AGE} $\to$ ``Age:''), and all values remain as strings. Missing values are represented by the token ``Refused.'' Survey-wave metadata (region, field dates) is provided in the system prompt.
  \item \textbf{Tabular methods.} Ordinal variables (Age, Education, Religious Attendance, Income, Political Ideology) are mapped to numeric values $\{1, \ldots, K\}$, preserving order. Categorical variables (Region, Gender, Citizenship, Marital Status, Religion, Political Party, Race) are one-hot encoded. Opinion variables are mapped to ordinal scales using the response-to-number mappings provided by OpinionQA. Missing values are represented as \texttt{NaN}.
\end{itemize}

For each condition, we select the relevant demographic variables and a single opinion variable, apply listwise deletion to obtain a complete-case dataset, and then introduce missingness as described in Section~\ref{sec:simulation_study_design}, using the \texttt{mice::ampute} implementation. We do not use survey weights: all regressions are unweighted, for the imputed datasets and for the reference estimate on the complete cases alike.

\FloatBarrier
\section{Pooling via Rubin's Rules}\label{app:rubins_rules}
For multiple imputation ($m$ completed datasets), we pool estimates using Rubin's rules \citep{rubin1987,little2019}. Let $\hat\beta^{(j)}$ and $\mathrm{SE}^{(j)}$ denote the coefficient estimate and standard error from the $j$-th imputed dataset. The pooled coefficient is
\[
  \bar\beta = \frac{1}{m}\sum_{j=1}^{m}\hat\beta^{(j)}.
\]
Its total variance combines within- and between-imputation components:
\begin{align*}
  \bar{U} &= \frac{1}{m}\sum_{j=1}^{m}\bigl(\mathrm{SE}^{(j)}\bigr)^{2}, \\
  B &= \frac{1}{m-1}\sum_{j=1}^{m}\bigl(\hat\beta^{(j)}-\bar\beta\bigr)^{2}, \\
  T &= \bar{U} + \Bigl(1+\frac{1}{m}\Bigr)B,
\end{align*}
where $\bar{U}$ is the mean within-imputation variance, $B$ the between-imputation variance, and $T$ the total variance. The pooled 95\,\% confidence interval is
\begin{align*}
  \mathrm{CI} &= \bar\beta \;\pm\; t_{\nu,\,1-\alpha/2}\,\sqrt{T}, \\
  \nu &= (m-1)\Bigl(1+\frac{\bar{U}}{(1+1/m)\,B}\Bigr)^{\!2},
\end{align*}
where $\nu$ are the approximate degrees of freedom \citep{rubin1987}.

\paragraph{Finite population correction.} Each $D_{\text{full}}$ is drawn without replacement from the $N$ respondents of $D_{\text{true}}$, and coverage is evaluated against the coefficient on $D_{\text{true}}$. The sampling variance around this coefficient is therefore $(1-n/N)$ times its usual value. We apply this factor to $\bar{U}$ in $T$ and $\nu$ and leave $B$ unchanged \citep{vinkPoolingMultipleImputations2014}. For fits on a single dataset, the standard error is multiplied by $\sqrt{1-n/N}$. With $n=500$, $n/N$ ranges from 0.05 to 0.72 across the 150 ATP variables (median 0.19). Without the correction, the coverage with complete data is 0.979 instead of 0.964.

For single-imputation methods (e.g., Mode Imputation), confidence intervals reflect only sampling variability and do not account for imputation uncertainty. For LLM-based imputation, drawing the $m$ imputations from the LLM's output does not constitute \textit{proper} multiple imputation in Rubin's sense, because the variability across draws does not necessarily reflect the full uncertainty about the missing values. Improper imputations are expected to underestimate total variance, producing narrower confidence intervals and potential undercoverage. Whether and to what extent this occurs is an empirical question we investigate through the coverage and interval-width metrics defined in Section~\ref{sec:evaluation} and by comparing the variability of different ICL configurations.

\FloatBarrier
\section{In-Context Imputation}\label{app:icl}

\begin{figure}[t!]
  \centering
  \includegraphics[width=\linewidth]{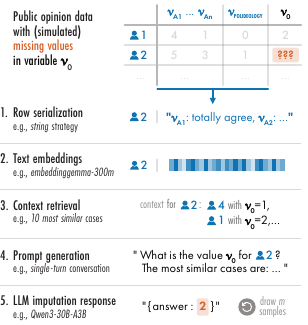}
  \caption{\textbf{Imputation through in-context learning.}
  Missing survey responses are imputed by an LLM conditioned on complete cases retrieved from the same dataset.
  }
  \label{fig:simulation_setup}
\end{figure}

We frame univariate imputation as a constrained generation task over the observed answer options of the missing variable. For each missing value, we retrieve $k \in \{10, 100\}$ in-context examples from complete cases and prompt an LLM to output one of the valid answer options or a probability distribution over them.
We compare several retrieval and prompting configurations because we hypothesize that they can improve uncertainty quantification in LLM-based imputation: by controlling which examples enter the context (and how they are presented), we aim to elicit more calibrated output variability across multiple imputations. This is particularly important because prior work has found that LLM predictions can be systematically overconfident \citep{xiong2024can,bisbee2024synthetic,boelaert2025machine}.

\paragraph{Embedding-based retrieval.}
We compute dense text embeddings for all cases using the EmbeddingGemma-300M \citep{veraEmbeddingGemmaPowerfulLightweight2025} embedding model and use cosine similarity for nearest-neighbor selection. Retrieved candidates exclude cases with missing values in the target variable.

We distinguish two \textbf{row-serialization strategies}. The \textit{String} strategy concatenates all variable--value pairs into a single string (one embedding per respondent). The \textit{List} strategy embeds each variable--value pair separately (one embedding per variable); we then aggregate per-variable rankings using Reciprocal Rank Fusion \citep{cormack2009reciprocal}. For the \textit{List} strategy, we set the weight of the to-be-imputed variable to zero to avoid trivial retrieval based on the target itself.

We compare the following \textbf{retrieval strategies}: \textit{Random}, \textit{Most Similar}, \textit{Least Similar}, and \textit{Diverse}, where the similarity of two embeddings is determined by their cosine similarity. The \textit{Diverse} strategy selects $k$ examples that are spread out in embedding space via $k$-medoids clustering \citep{schubert2021fast}.

We additionally consider \textbf{stratified retrieval} with respect to the to-be-imputed variable $v_o$. When enabled, we retrieve examples separately for each answer option and allocate the context budget uniformly across options, ensuring that the prompt exposes the model to all possible labels even when the marginal distribution is imbalanced.

For \textbf{prompting}, we compare three prompt formats. \textit{Multi-turn} presents each retrieved example as a separate user--assistant exchange. \textit{Single-turn} places all examples into a single user message and explicitly describes the selection strategy (e.g., ``most similar'' vs.\ ``diverse''). \textit{Single-turn verbalized} extends the single-turn format by instructing the model to output a verbalized probability distribution over answer options~\citep[following][]{meister-etal-2025-benchmarking, ahnert-etal-2026-survey}, from which we sample imputations. We provide survey-wave metadata (region and field dates) as additional context in the system message. The full prompt for each format is reproduced in Appendix~\ref{app:prompts} (Appendix Figures~\ref{fig:prompt_multi_turn}--\ref{fig:prompt_single_turn_verbalized}).

\paragraph{Answer extraction.}
We use structured outputs to constrain the model to produce valid labels (or, optionally, a probability distribution over labels in JSON). We sample at temperature 1.0 and treat the resulting completed datasets as multiple imputations for pooling. With the verbalized prompt, each imputation is a draw from a stated distribution, and for the reasoning models all $m$ draws for a missing value come from one completion (Section~\ref{sec:comparison}).

\paragraph{Configuration selection.} We run a factorial sweep over 10 or 100 in-context examples, retrieval strategies, row-serialization strategies, stratified vs.\ non-stratified retrieval, and prompt formats with Qwen3-8B and Qwen3-30B-A3B on W92, using the design and evaluation of Sections~\ref{sec:simulation_study_design} and~\ref{sec:evaluation}. Of the full grid, we omit \textit{List} serialization for \textit{Random} retrieval, where serialization does not affect which examples are drawn, and the verbalized distribution for the \textit{Multi-turn} format, which leaves 84 configurations per generator.

\begin{figure*}[t]
  \centering
  \includegraphics[width=1.0\linewidth]{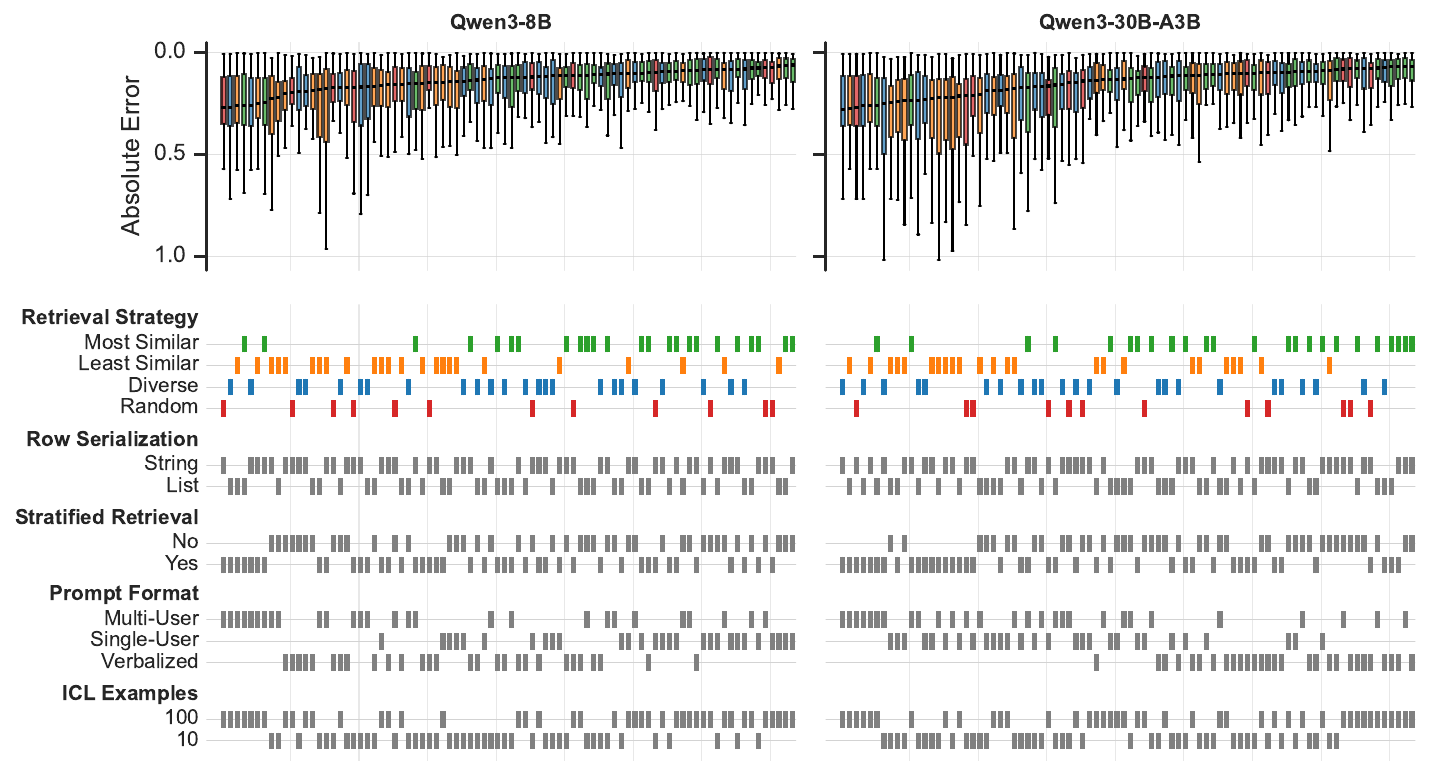}
  \caption{\textbf{Configuration selection: Absolute error by ICL configuration ($\downarrow$).} Configurations are sorted by median absolute error, shown separately for Qwen3-8B and Qwen3-30B-A3B. Each column shows the distribution over variables and missingness settings for one configuration. The lower panels indicate the active design choice in each configuration. Across metrics, the two models show broadly similar ordering of retrieval and serialization choices, but the \textit{Single-turn verbalized} prompt is less competitive for Qwen3-8B than for Qwen3-30B-A3B.}
  \label{fig:study_1_abs_error_appendix}
\end{figure*}

\begin{figure*}[t]
  \centering
  \includegraphics[width=1.0\linewidth]{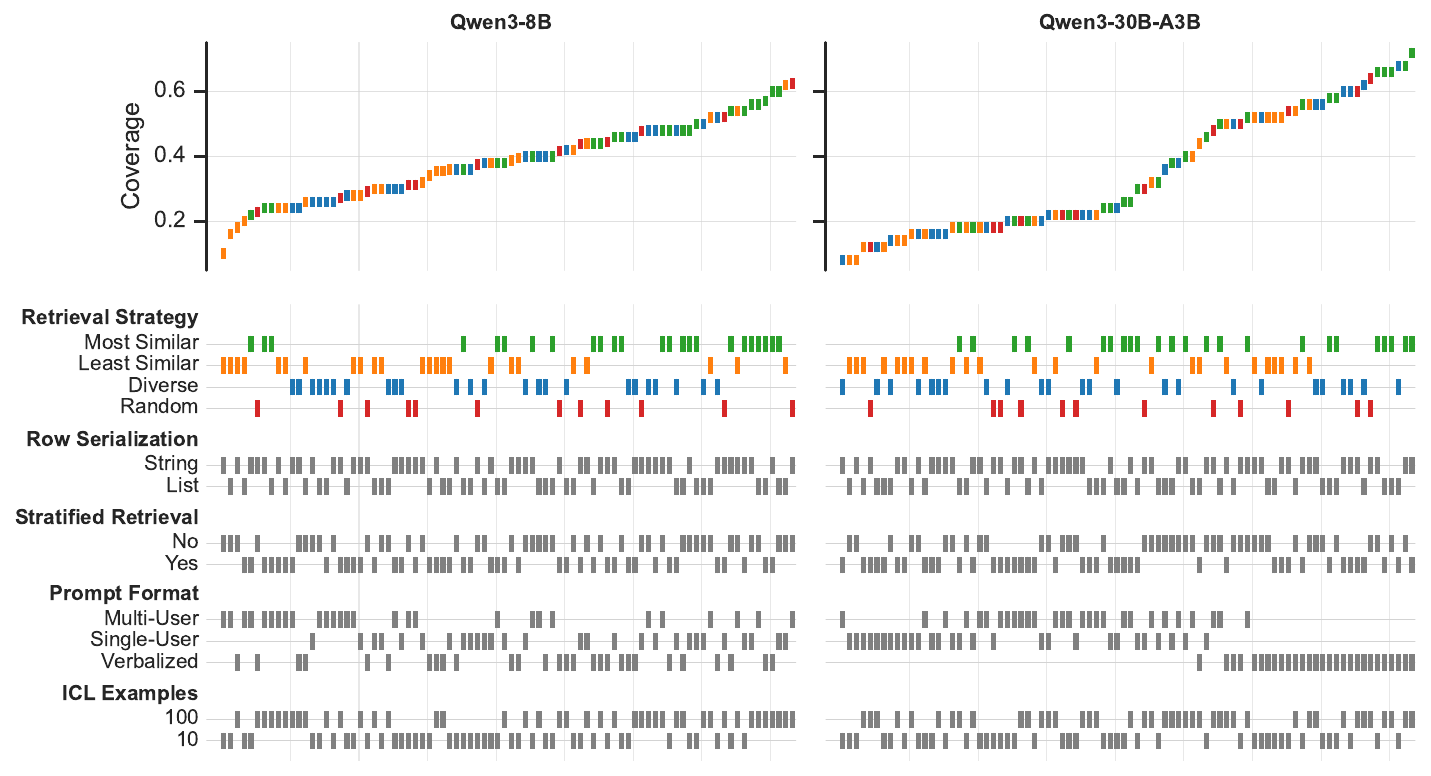}
  \caption{\textbf{Configuration selection: Coverage by ICL configuration ($\uparrow$).} Configurations are sorted by aggregate coverage rate, shown separately for Qwen3-8B and Qwen3-30B-A3B. Each marker is one configuration's aggregate coverage rate across all variables and missingness settings of W92. The lower panels indicate the active design choice in each configuration. Coverage rankings only partially track the absolute-error rankings (Figure~\ref{fig:study_1_abs_error_appendix}), reinforcing the need to evaluate calibration jointly with absolute error.}
  \label{fig:study_1_coverage_appendix}
\end{figure*}

\begin{figure*}[t]
  \centering
  \includegraphics[width=1.0\linewidth]{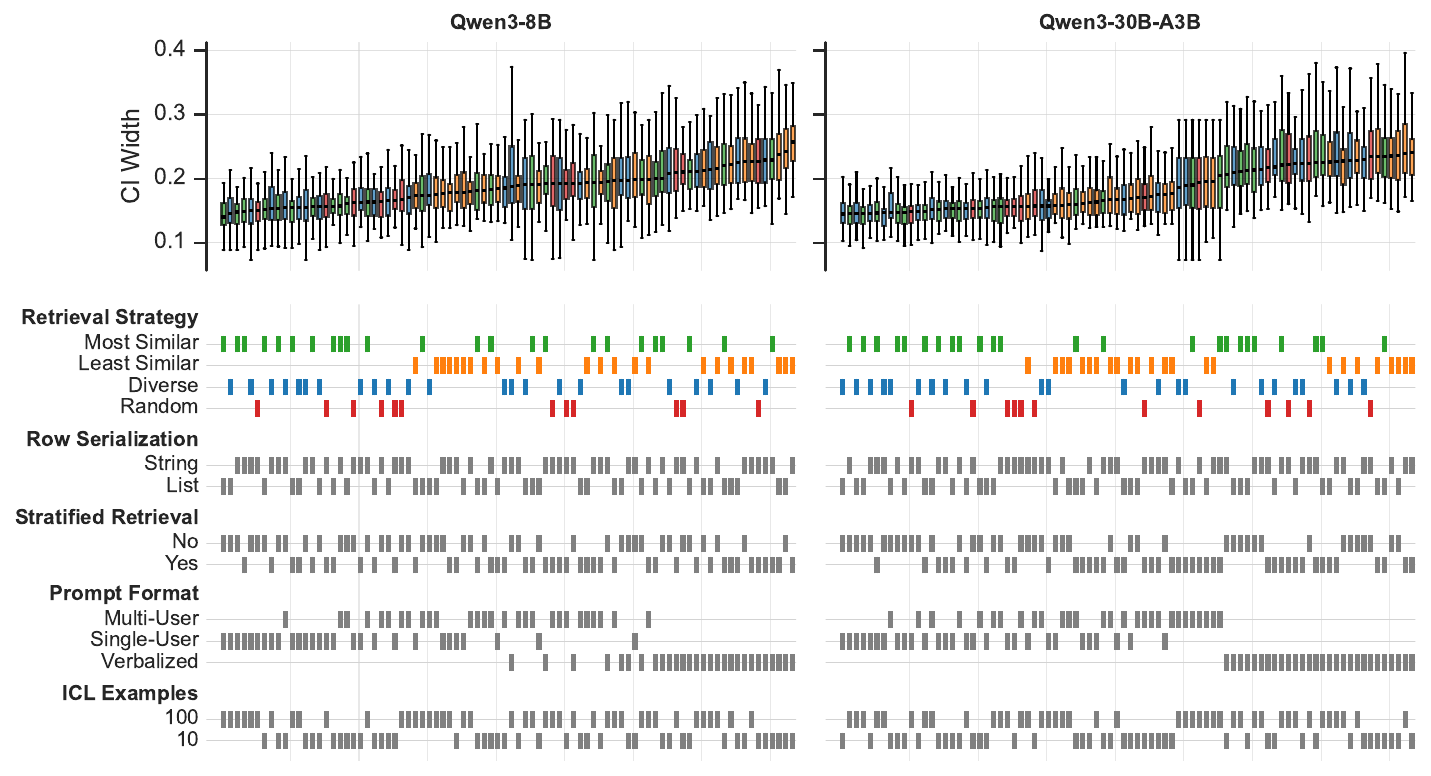}
  \caption{\textbf{Configuration selection: 95\% confidence interval width by ICL configuration ($\downarrow$).} Configurations are sorted by median interval width, shown separately for Qwen3-8B and Qwen3-30B-A3B. Each column shows the distribution over variables and missingness settings for one configuration. The lower panels indicate the active design choice in each configuration. Some high-coverage configurations achieve coverage by producing wider intervals, which is why width must be interpreted jointly with absolute error and coverage.}
  \label{fig:study_1_width_appendix}
\end{figure*}

Appendix Figure~\ref{fig:study_1_abs_error_appendix} shows that, for Qwen3-30B-A3B, the prompt format matters most, but its effect depends on the retrieval design. The most consistent pattern is that configurations using the \textit{Single-turn verbalized} prompt rank among the lowest-error configurations with both 10 and 100 in-context examples, particularly when combined with \textit{Most Similar} retrieval. Within retrieval design, \textit{Most Similar} retrieval with \textit{String} row serialization and non-stratified selection performs robustly, whereas alternative retrieval combinations vary more substantially. Appendix Figures~\ref{fig:study_1_coverage_appendix} and~\ref{fig:study_1_width_appendix} confirm that this configuration also achieves competitive coverage with narrow interval widths.

The same broad ordering of configurations holds for Qwen3-8B, though the \textit{Single-turn verbalized} prompt is less effective for the smaller model (see Appendix Figures~\ref{fig:study_1_abs_error_appendix}--\ref{fig:study_1_width_appendix} for the corresponding 8B/30B comparison across absolute error, coverage, and interval width). Based on this joint assessment, we select \textbf{the \textit{Single-turn verbalized} prompt, \textit{Most Similar} retrieval, \textit{String} row serialization, and no stratification}, and keep both 10 and 100 in-context examples for the main comparison.

\section{Baseline Hyperparameter Tuning}\label{app:baseline_tuning}

\begin{table*}[!tbp]
  \centering
  \small
  \begin{tabular}{lrrrrrr}
\toprule
min. rows per leaf & $k=0$ & $k=1$ & $k=3$ & $k=5$ & $k=10$ & $k=20$ \\
\midrule
1 & 0.124 / 0.42 & 0.056 / 0.70 & 0.061 / 0.70 & 0.059 / 0.68 & 0.064 / 0.70 & 0.056 / 0.70 \\
5 & 0.149 / 0.30 & 0.029 / 0.78 & 0.036 / 0.84 & 0.030 / 0.78 & 0.034 / 0.80 & 0.033 / 0.78 \\
10 & 0.159 / 0.30 & 0.033 / 0.90 & 0.026 / 0.88 & 0.026 / 0.92 & 0.033 / 0.88 & 0.029 / 0.90 \\
20 & 0.165 / 0.36 & 0.043 / 0.90 & 0.044 / 0.92 & 0.043 / 0.88 & 0.047 / 0.88 & 0.044 / 0.90 \\
40 & 0.110 / 0.52 & 0.051 / 0.88 & 0.050 / 0.82 & 0.049 / 0.78 & 0.055 / 0.80 & 0.058 / 0.80 \\
\midrule
\multicolumn{7}{l}{\textit{Full Data}: 0.027 / 1.00} \\
\multicolumn{7}{l}{\textit{Complete Case}: 0.066 / 0.78} \\
\bottomrule
\end{tabular}

  \caption{\textbf{MICE Forest by minimum rows per leaf and mean-match candidates $k$} on W92. Each cell shows the median absolute error / coverage rate over 50 runs. We use ten rows per leaf with $k=5$.}
  \label{tab:miceforest_mmc}
\end{table*}

We tune the hyperparameters of the three statistical imputation baselines on W92 only (10 opinion variables under MCAR and left- and right-tailed MAR and MNAR, i.e., 50 runs per hyperparameter value), so that the main comparison is not tuned on its own evaluation data. The number of MICE iterations plays no role in our setting because only one variable has missing values, so the chained equations converge in a single pass.

\paragraph{MICE PMM.}\label{par:tuning_pmm} We tune the donor pool size $d$ (Table~\ref{tab:pmm_donors}). Across $d \in \{1, 3, 5, 10, 20\}$, the package default $d=5$ gives the lowest median absolute error and the highest coverage (0.96), and we use it.

\paragraph{MICE Forest.}\label{par:tuning_forest} We tune the number of mean-match candidates $k$ jointly with the minimum number of rows per leaf of the LightGBM models (Table~\ref{tab:miceforest_mmc}). Ten rows per leaf give a coverage of at least 0.88 for every $k>0$, whereas the default of one row per leaf reaches at most 0.70. We use ten rows per leaf with $k=5$, which has the joint-lowest error and the highest coverage.

\paragraph{TabPFN.}\label{par:tuning_tabpfn} We tune the sampling temperature $T$ (Table~\ref{tab:tabpfn_temperature}). $T=0.75$ gives the lowest error and $T=1.0$ the highest coverage (0.90), whereas greedy decoding reaches a coverage of only 0.42. We use $T=1.0$ for its coverage.

\begin{table}[h]
  \centering
  \resizebox{\columnwidth}{!}{\begin{tabular}{lrrrrr}
\toprule
 & \multicolumn{2}{c}{Absolute Error} & \multicolumn{2}{c}{CI Width} & Coverage \\
\cmidrule(lr){2-3} \cmidrule(lr){4-5}
Method & Median & IQR & Median & IQR & \\
\midrule
\textit{Full Data} & 0.027 & 0.020 & 0.161 & 0.008 & 1.000 \\
Complete Case & 0.066 & 0.075 & 0.223 & 0.031 & 0.780 \\
MICE PMM (d=1) & 0.061 & 0.072 & 0.234 & 0.131 & 0.860 \\
MICE PMM (d=3) & 0.053 & 0.068 & 0.241 & 0.094 & 0.940 \\
MICE PMM (d=5) & 0.046 & 0.057 & 0.249 & 0.105 & 0.960 \\
MICE PMM (d=10) & 0.052 & 0.062 & 0.235 & 0.095 & 0.920 \\
MICE PMM (d=20) & 0.050 & 0.059 & 0.236 & 0.084 & 0.900 \\
\bottomrule
\end{tabular}
}
  \caption{\textbf{MICE PMM by donor pool size $d$} on W92 (median absolute error and CI width, and coverage rate, over 50 runs per setting). The default $d=5$ is used throughout.}
  \label{tab:pmm_donors}
\end{table}

\begin{table}[h]
  \centering
  \resizebox{\columnwidth}{!}{\begin{tabular}{lrrrrr}
\toprule
 & \multicolumn{2}{c}{Absolute Error} & \multicolumn{2}{c}{CI Width} & Coverage \\
\cmidrule(lr){2-3} \cmidrule(lr){4-5}
Method & Median & IQR & Median & IQR & \\
\midrule
\textit{Full Data} & 0.027 & 0.020 & 0.161 & 0.008 & 1.000 \\
Complete Case & 0.066 & 0.075 & 0.223 & 0.031 & 0.780 \\
TabPFN (greedy) & 0.101 & 0.100 & 0.184 & 0.073 & 0.420 \\
TabPFN (T=0.5) & 0.056 & 0.081 & 0.199 & 0.062 & 0.780 \\
TabPFN (T=0.75) & 0.043 & 0.060 & 0.211 & 0.059 & 0.880 \\
TabPFN (T=1) & 0.049 & 0.057 & 0.214 & 0.074 & 0.900 \\
TabPFN (T=1.5) & 0.071 & 0.088 & 0.209 & 0.056 & 0.660 \\
\bottomrule
\end{tabular}
}
  \caption{\textbf{TabPFN by sampling temperature $T$} on W92 (median absolute error and CI width, and coverage rate, over 50 runs per setting). Greedy assigns each cell its most likely class. We use $T=1.0$.}
  \label{tab:tabpfn_temperature}
\end{table}

\section{Further Surveys}\label{app:gles_brfss}

\begin{figure*}[t!]
  \centering
  \includegraphics[width=\linewidth]{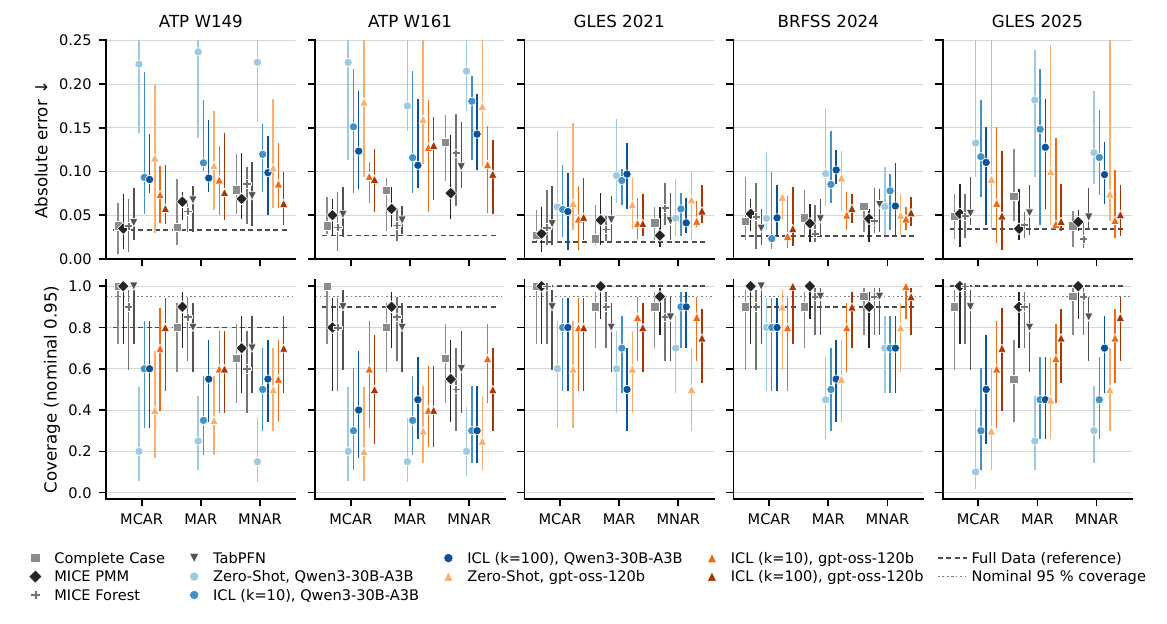}
  \caption{\textbf{Absolute error and coverage on the two later ATP waves and the GLES and BRFSS surveys.} Each marker summarizes 10 runs under MCAR and 20 runs (10 items, left- and right-tailed) under MAR and MNAR, with 95\% bootstrap CIs for the median error and Wilson CIs for coverage. The data of all surveys except GLES 2021 were published after gpt-oss-120b's training cutoff (June 2024), and those of ATP W161 and BRFSS 2024 after the release of Qwen3-30B-A3B.}
  \label{fig:gles_brfss}
\end{figure*}

\begin{figure}[t!]
  \centering
  \includegraphics[width=\linewidth]{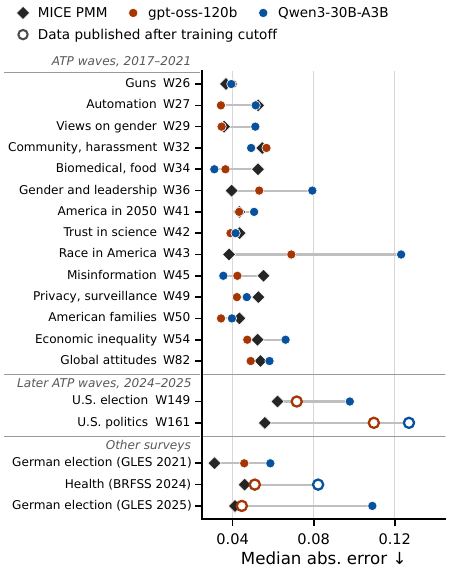}
  \caption{\textbf{Median absolute error per survey}, for the 14 ATP waves of the main comparison and the two later ATP waves and three other surveys of this section, over the same conditions as in Figure~\ref{fig:coverage_by_survey}. The LLMs use 100-shot ICL. Hollow markers as in Figure~\ref{fig:coverage_by_survey}.}
  \label{fig:abs_error_by_survey}
\end{figure}

Our main comparison draws on a single U.S.\ opinion panel. Here we repeat it on three datasets that differ from the ATP in country and language (GLES), in survey domain (BRFSS), and in their timing relative to the generators' training data, and on two later waves of the ATP itself.

\paragraph{Datasets.}
The \textbf{German Longitudinal Election Study} (GLES) post-election cross-sections of the 2021 and 2025 German federal elections \citep{gles2021,gles2025} provide political attitudes of German voters, surveyed in German (GESIS ZA6835 v2.0.0, 3{,}431 respondents in 2021; ZA10100 v4.0.0, 7{,}336 respondents in 2025). The \textbf{Behavioral Risk Factor Surveillance System} (BRFSS) 2024 \citep{brfss2024} is a U.S.\ telephone survey on health behaviors, health status, and social determinants of health; we restrict it to the 254{,}773 interviews conducted from July~1, 2024 onwards (BRFSS 2024-H2). GLES 2021 was fielded before gpt-oss-120b's training cutoff (June~2024); GLES 2025 (fielded February--April 2025) and BRFSS 2024-H2 were fielded after it. The BRFSS 2024 microdata were released in August 2025, after the release of both generators; the GLES 2025 microdata were first released in June 2025 and thus before Qwen3-30B-A3B-Instruct-2507 (July 2025), so for Qwen only BRFSS 2024-H2 offers a strict guarantee. No training cutoff is documented for Qwen3-30B-A3B-Instruct-2507, and reported cutoffs can differ from effective ones \citep{chengDatedDataTracing2024}. We therefore use its release date as an upper bound.

\paragraph{Variables.}
Each dataset replaces the ATP's demographic block and the outcome of the analysis model with its own analogues. For GLES, the outcome is the 11-point left--right self-placement, the counterpart of $v_{\text{POLIDEOLOGY}}$, and the auxiliary variables $V_A$ are age, gender, East/West Germany, education, employment, marital status, household income, religiosity, and party identification. For BRFSS, the outcome is 5-point self-rated general health, and $V_A$ comprises age, sex, race/ethnicity, census region, urban/rural county, education, employment, marital status, household income, and home ownership. We select 10 items per dataset with the two-stage screening of the ATP variables. For GLES, the same 10 items are imputed in 2021 and 2025: we screen the items asked in both years and rank them on the 2021 data only, so that no post-cutoff data enter the selection. The GLES items cover reducing income differences, state involvement in the economy, whether voting makes a difference, immigrants and crime, immigration (11-point scale), taxes versus welfare benefits (11-point scale), two populism items (a strong leader; the people rather than politicians should decide), the general economic situation, and the federal government's performance. The BRFSS items cover food insecurity, loneliness, the last routine checkup and dental visit, days of poor mental and physical health, e-cigarette use, and three adverse childhood experiences. GLES prompts use the original German question wording and answer labels; the system prompt states the survey region and field dates as for the ATP.

\paragraph{Design and methods.}
The simulation design and evaluation are identical to the main comparison: one sample of 500 respondents per condition, 50\% missingness in the item under MCAR, MAR (driven by the outcome), and MNAR (driven by the item), left- and right-tailed MAR and MNAR, and $m=5$ imputations. With 10 items, each method is evaluated on 10 runs under MCAR and 20 runs (both tails) under MAR and MNAR per dataset. We compare Qwen3-30B-A3B and gpt-oss-120b, each with zero-shot prompting and the selected ICL configuration with 10 or 100 in-context examples, against Complete Case, MICE PMM, MICE Forest, and TabPFN, with Full Data as a reference. As for the ATP, each condition samples from the complete cases on the auxiliary variables, the outcome, and the item (2{,}300--2{,}400 per item in GLES 2021, 5{,}100--5{,}400 in GLES 2025, and 14{,}000--193{,}000 in BRFSS, where the pool is capped at a random subsample of 20{,}000); answers such as ``don't know'' and refusals are treated as missing, and survey weights are not used.

\paragraph{Results.}
Figure~\ref{fig:gles_brfss} shows absolute error and coverage per dataset and mechanism. With gpt-oss-120b, 100-shot ICL has absolute errors similar to those of the statistical baselines on the GLES and BRFSS surveys (medians of 0.04--0.06 against 0.02--0.06), including the two surveys fielded after its training cutoff. Its coverage matches that of the baselines on BRFSS but is lower on both GLES surveys (0.70--0.85), whereas MICE PMM and MICE Forest reach at least 0.85 everywhere. Qwen3-30B-A3B does worse: its ICL errors reach 0.10 under MAR on BRFSS and 0.15 on GLES 2025, with coverage down to 0.30. Zero-shot prompting and most ICL variants have lower coverage on GLES 2025 than on GLES 2021. This single pair of surveys cannot be attributed to the training cutoff: BRFSS also postdates the cutoff of gpt-oss-120b, and there its 100-shot coverage is 0.90 or above. With 10 to 20 runs per cell, these differences are descriptive. They are consistent with the main comparison: in-context imputation with a strong generator is competitive with, but not better than, the statistical baselines. Figure~\ref{fig:abs_error_by_survey} places the three surveys next to the 14 ATP waves and the two later ATP waves described below. The median absolute error of MICE PMM stays at 0.06 or below on all 19 surveys. Among the 14 ATP waves and the three other surveys, that of gpt-oss-120b reaches 0.07 only on W43, and that of Qwen3-30B-A3B exceeds 0.10 on W43 and GLES 2025.

\paragraph{Later ATP waves.}
We also repeat the comparison on two ATP waves fielded after those of the main comparison, W149 (July 2024) and W161 (January 2025), both on U.S.\ politics. They are not part of OpinionQA, so we convert the Pew Research Center's microdata into its format and select 10 opinion variables per wave with the same two-stage screening. The microdata of W161 were released after both generators, and those of W149 after gpt-oss-120b's training cutoff. In-context imputation is not competitive with the baselines on these waves (Figure~\ref{fig:gles_brfss}): with 100 in-context examples, both generators have higher errors and lower coverage than MICE PMM under almost every mechanism. MICE PMM stays close to the nominal coverage under MAR and loses coverage mainly under MNAR. The selected variables of these waves are more strongly related to political ideology than those of the main comparison, which makes MNAR more severe. W161 is the only ATP wave published after the cutoff or release of both generators, but both waves also share a topic, so the cutoff and the topic cannot be separated with these data.

\section{Prediction Metrics versus Inference Metrics}\label{app:pred_inference}

Tables~\ref{tab:pred_inference_within} and~\ref{tab:pred_inference_method} extend Figure~\ref{fig:study_2_prediction_vs_inference} to six prediction metrics common in NLP evaluation and to all ten imputation methods of the main comparison. All metrics are computed on the imputed cells against the true answers and averaged over the $m=5$ imputations. Accuracy, macro-F1, and mean absolute error (MAE) score the individual answers, and TVD, Wasserstein-1, and Jensen--Shannon divergence compare the marginal answer distributions. Macro-F1 on the majority vote over the imputations gives the same picture.

The \emph{within-condition} correlation (Table~\ref{tab:pred_inference_within}) ranks the ten methods within each condition and averages the correlation over the 140 conditions under MCAR and the 280 conditions under MAR and MNAR. It asks whether, within a condition, the method that predicts better also estimates $\beta$ better. Because each condition has a single sampled dataset, the realized absolute error is noisy, so these correlations are lower bounds. The marginal-distribution metrics still reach 0.35--0.38 under MAR, so the near-zero values for accuracy and MAE are not explained by noise alone. The coverage columns rest on the 83, 239, and 183 conditions (MCAR, MAR, MNAR) on which the ten methods neither all cover nor all miss.

The \emph{method-level} correlation (Table~\ref{tab:pred_inference_method}) is computed over the ten method medians and asks whether ranking methods by a prediction metric reproduces their ranking by inference quality. The medians are stable across random halves of the opinion variables (Spearman--Brown corrected rank correlation of 0.73--0.99, except 0.59 for absolute error under MNAR), but each correlation rests on only ten points.

\begin{table*}[!tbp]
\centering
\small
\begin{tabular}{lrrrrrrrrr}
\toprule
 & \multicolumn{3}{c}{MCAR} & \multicolumn{3}{c}{MAR} & \multicolumn{3}{c}{MNAR} \\
\cmidrule(lr){2-4} \cmidrule(lr){5-7} \cmidrule(lr){8-10}
Prediction metric & Abs. Error & Cov. & Width & Abs. Error & Cov. & Width & Abs. Error & Cov. & Width \\
\midrule
\multicolumn{10}{l}{\textit{Individual-level prediction}} \\
\quad Accuracy & \textbf{+0.10} & \textbf{+0.23} & \textbf{$-$0.08} & \textbf{+0.12} & \textbf{+0.12} & \textbf{$-$0.11} & \textbf{+0.05} & \textbf{+0.10} & \textbf{$-$0.11} \\
\quad Macro-F1 & \textbf{+0.14} & \textbf{+0.21} & \textbf{$-$0.06} & \textbf{+0.29} & \textbf{+0.31} & \textbf{$-$0.06} & \textbf{+0.11} & +0.06 & \textbf{+0.15} \\
\quad MAE & \textbf{+0.11} & \textbf{+0.22} & \textbf{$-$0.11} & \textbf{+0.13} & \textbf{+0.13} & \textbf{$-$0.13} & \textbf{+0.06} & \textbf{+0.12} & \textbf{$-$0.11} \\
\addlinespace
\multicolumn{10}{l}{\textit{Marginal distribution}} \\
\quad TVD & \textbf{+0.11} & \textbf{+0.22} & \textbf{$-$0.18} & \textbf{+0.35} & \textbf{+0.40} & \textbf{$-$0.21} & \textbf{+0.05} & $-$0.02 & \textbf{+0.10} \\
\quad Wasserstein-1 & \textbf{+0.09} & \textbf{+0.23} & \textbf{$-$0.19} & \textbf{+0.38} & \textbf{+0.42} & \textbf{$-$0.21} & \textbf{+0.08} & +0.01 & \textbf{+0.17} \\
\quad JS divergence & \textbf{+0.11} & \textbf{+0.25} & \textbf{$-$0.19} & \textbf{+0.38} & \textbf{+0.42} & \textbf{$-$0.20} & \textbf{+0.08} & +0.02 & \textbf{+0.13} \\
\bottomrule
\end{tabular}

\caption{\textbf{Within-condition Spearman correlations between prediction and inference metrics} in the main comparison. The ten imputation methods are ranked within each condition, and the correlation is averaged over the 140 (MCAR) or 280 (MAR, MNAR) conditions. Cov.\ and Width refer to the 95\% CI. Positive values mean that better prediction goes with better inference, and bold values have a cluster-bootstrap 95\% CI that excludes 0.}
\label{tab:pred_inference_within}
\end{table*}

\begin{table*}[!tbp]
\centering
\small
\begin{tabular}{lrrrrrrrrr}
\toprule
 & \multicolumn{3}{c}{MCAR} & \multicolumn{3}{c}{MAR} & \multicolumn{3}{c}{MNAR} \\
\cmidrule(lr){2-4} \cmidrule(lr){5-7} \cmidrule(lr){8-10}
Prediction metric & Abs. Error & Cov. & Width & Abs. Error & Cov. & Width & Abs. Error & Cov. & Width \\
\midrule
\multicolumn{10}{l}{\textit{Individual-level prediction}} \\
\quad Accuracy & +0.09 & \textbf{+0.45} & $-$0.26 & \textbf{+0.26} & \textbf{+0.35} & \textbf{$-$0.43} & $-$0.19 & +0.05 & $-$0.37 \\
\quad Macro-F1 & +0.21 & \textbf{+0.53} & \textbf{$-$0.41} & \textbf{+0.83} & \textbf{+0.90} & \textbf{$-$0.76} & +0.01 & $-$0.31 & +0.39 \\
\quad MAE & +0.10 & \textbf{+0.54} & $-$0.20 & \textbf{+0.30} & \textbf{+0.36} & \textbf{$-$0.41} & $-$0.36 & $-$0.09 & $-$0.27 \\
\addlinespace
\multicolumn{10}{l}{\textit{Marginal distribution}} \\
\quad TVD & \textbf{+0.50} & \textbf{+0.56} & \textbf{$-$0.60} & \textbf{+0.90} & \textbf{+0.87} & \textbf{$-$0.72} & +0.08 & $-$0.26 & \textbf{+0.50} \\
\quad Wasserstein-1 & +0.39 & \textbf{+0.50} & \textbf{$-$0.52} & \textbf{+0.92} & \textbf{+0.92} & \textbf{$-$0.77} & $-$0.01 & \textbf{$-$0.39} & \textbf{+0.56} \\
\quad JS divergence & \textbf{+0.41} & \textbf{+0.52} & \textbf{$-$0.59} & \textbf{+0.89} & \textbf{+0.85} & \textbf{$-$0.75} & +0.04 & $-$0.35 & \textbf{+0.56} \\
\bottomrule
\end{tabular}

\caption{\textbf{Method-level Spearman correlations between prediction and inference metrics} in the main comparison, computed across the medians of the ten imputation methods per mechanism (means for coverage). Cov.\ and Width refer to the 95\% CI. Positive values mean that better prediction goes with better inference, and bold values have a cluster-bootstrap 95\% CI that excludes 0.}
\label{tab:pred_inference_method}
\end{table*}

\paragraph{Configuration selection.} Figure~\ref{fig:study_1_metric_agreement} ranks all 84 Qwen3-30B-A3B configurations by accuracy, macro-F1, TVD, and Wasserstein-1 of the imputed answers, and by absolute error and coverage, where coverage is ranked by its distance to 0.95. There are 20 conditions under MAR and MNAR and 10 under MCAR, which has no tail. Because the configuration-selection runs stored a single imputed dataset each, their prediction metrics rest on one imputation. For the \textit{Single-turn verbalized} prompt, that imputation is a draw from the stated distribution rather than its most likely answer, which lowers accuracy and macro-F1 by construction. All correlations are Spearman rank correlations. Ranking the configurations within each condition instead, where the variable and the sampled data are fixed, accuracy and macro-F1 correlate with absolute error at only 0.14--0.38 and with coverage at $-0.03$--0.18. Each configuration has a single run per condition, so no split-half upper bound can be computed there.

Repeating the across-condition analysis for the 84 Qwen3-8B configurations gives a different picture. The Qwen3-8B runs use the same sampled datasets and missingness masks as the Qwen3-30B-A3B runs. Here, accuracy and macro-F1 correlate with the ranking by absolute error at 0.30--0.67 and with the ranking by coverage at 0.13--0.53, close to or at the split-half upper bounds (0.37--0.60 and 0.23--0.39), and TVD does no better (0.12--0.50).

\FloatBarrier
\section{Computational Cost}\label{app:walltime}

\begin{figure}[!tbp]
  \centering
  \includegraphics[width=\linewidth]{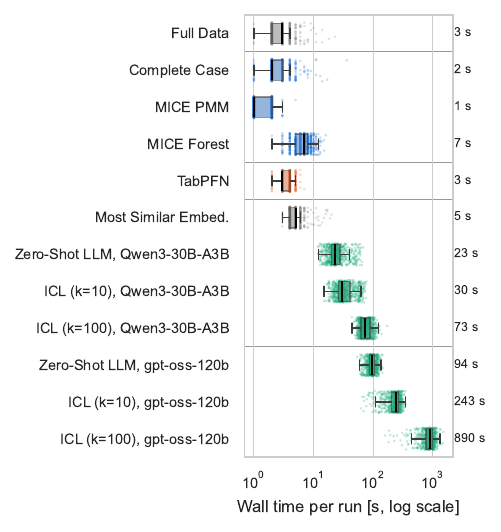}
  \caption{\textbf{Wall time per run in the main comparison} (log scale). One run imputes one opinion variable and includes evaluation and logging. Boxes show the interquartile range over the 700 conditions, and the numbers are medians. The methods ran on different hardware (Appendix~\ref{app:walltime}).}
  \label{fig:walltime}
\end{figure}

Figure~\ref{fig:walltime} shows the wall time of one run in the main comparison, i.e., one method imputing one opinion variable (500 respondents, 50\% missingness, $m=5$ imputations), including evaluation and logging. The statistical baselines take 1--7\,s per run (median), whereas in-context imputation takes 30--73\,s with Qwen3-30B-A3B and 243--890\,s with gpt-oss-120b, up to three orders of magnitude more. The methods ran on different hardware: MICE PMM, MICE Forest, and TabPFN on local machines, mostly a laptop, and the LLMs on NVIDIA H100 GPUs in a compute cluster. The times are therefore indicative rather than a controlled benchmark, but the gap is far larger than hardware differences can explain, especially since the LLMs ran on the faster hardware.

\FloatBarrier
\section{Variable-Level Analysis}\label{app:variable_level}

To investigate which characteristics of a survey question predict ICL imputation performance, we embed all 140 opinion questions of the main comparison using the EmbeddingGemma-300M model and project them into two dimensions via PCA on the standardized embedding matrix.

\begin{figure*}[!tbp]
  \centering
  \includegraphics[width=\linewidth]{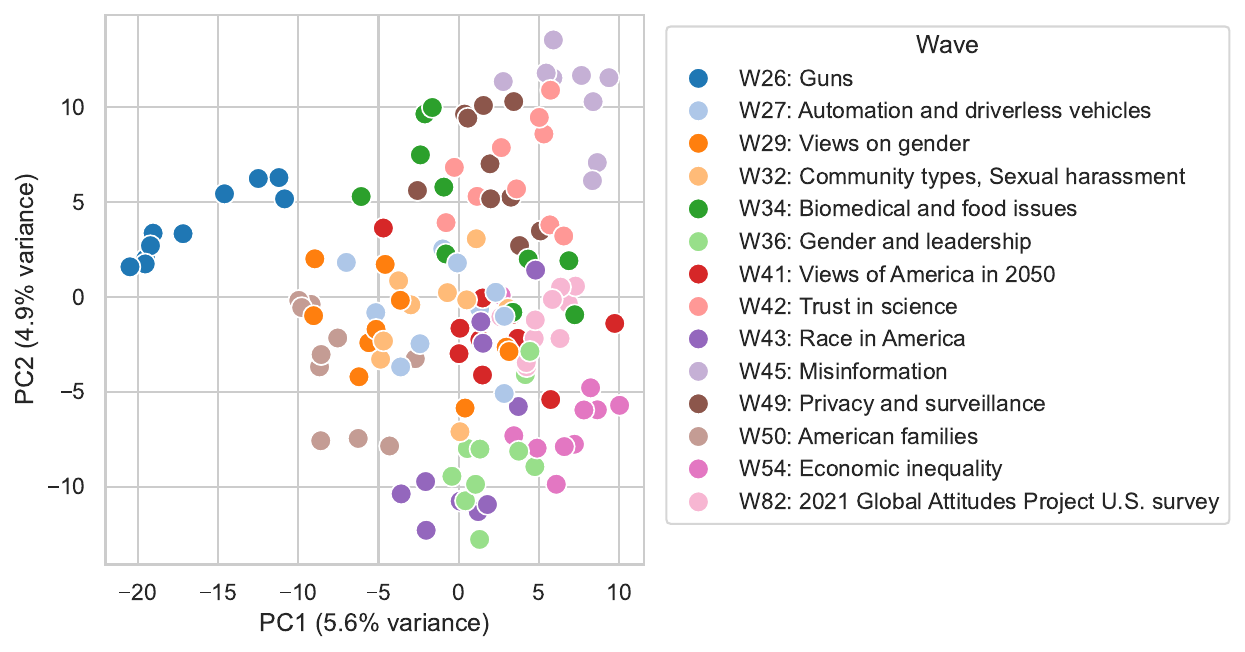}
  \caption{\textbf{Question embedding PCA colored by survey wave.} Each point represents one of the 140 opinion questions of the main comparison, projected onto the first two principal components of the standardized embedding space. Questions cluster by survey wave and topic.}
  \label{fig:pca_survey_wave}
\end{figure*}

\begin{figure*}[!tbp]
  \centering
  \includegraphics[width=\linewidth]{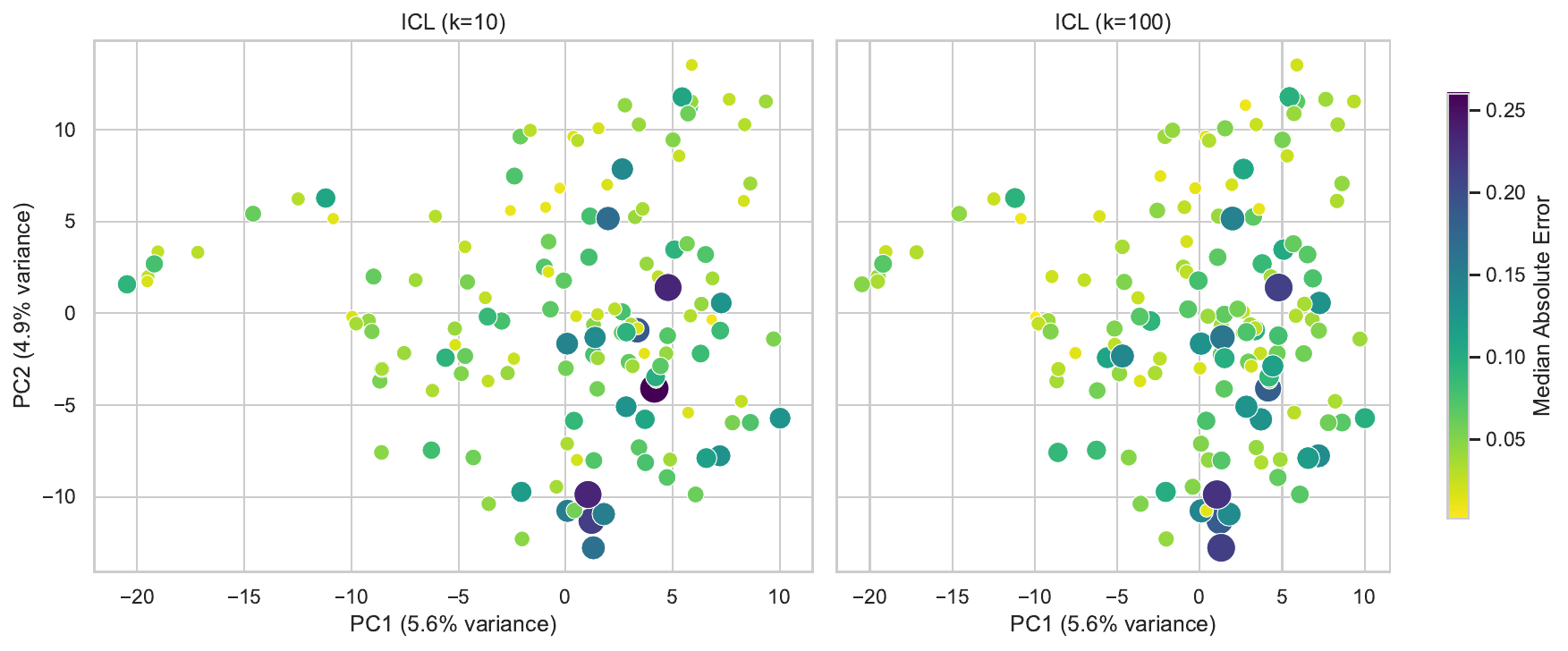}
  \caption{\textbf{Question embedding PCA colored by median absolute error.} Point color and size encode the median absolute error (aggregated across missingness settings) for 10-shot and 100-shot ICL. Performance varies spatially across the embedding space, indicating that question semantics are predictive of imputation quality.}
  \label{fig:pca_abs_error}

  \bigskip
  \includegraphics[width=\linewidth]{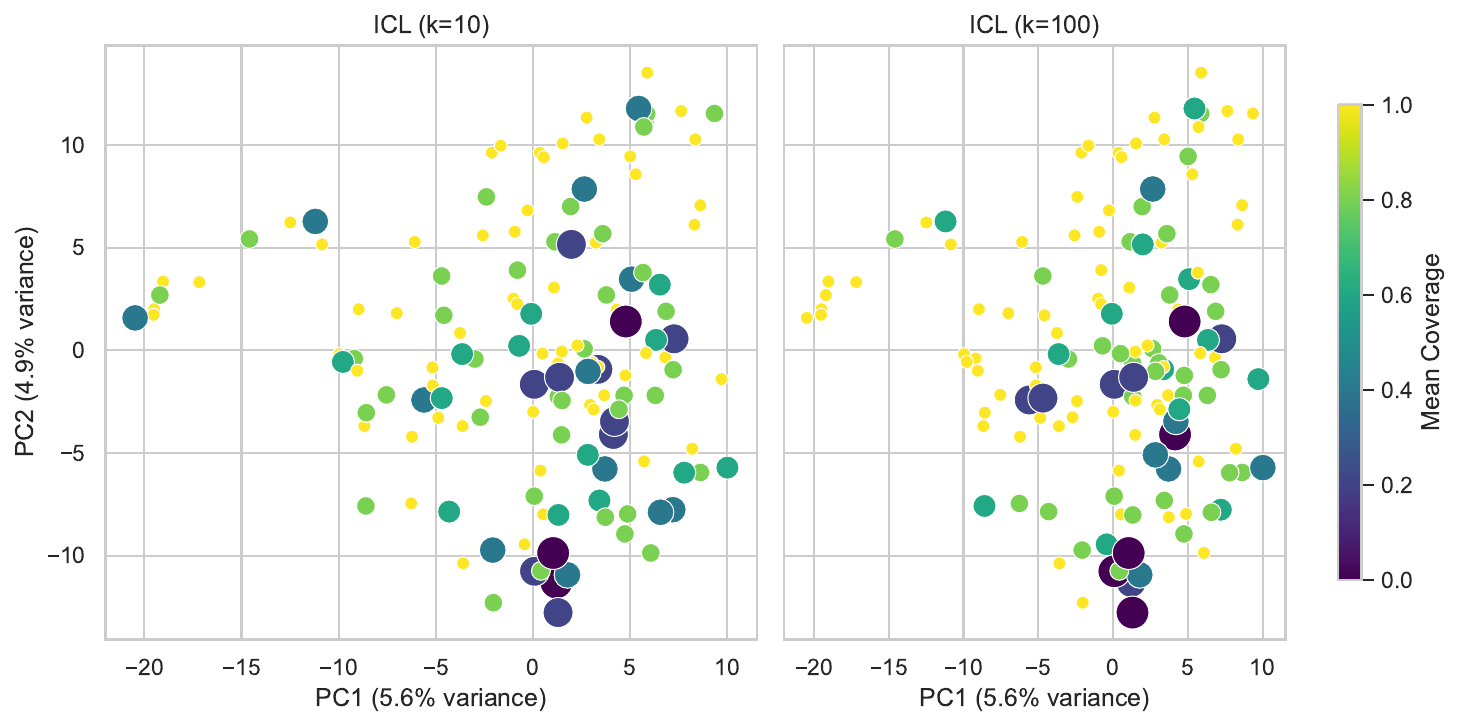}
  \caption{\textbf{Question embedding PCA colored by mean coverage.} Point color and size encode the mean coverage rate (aggregated across missingness settings) for 10-shot and 100-shot ICL. Regions of lower coverage partially overlap with regions of higher absolute error in Figure~\ref{fig:pca_abs_error}.}
  \label{fig:pca_coverage}
\end{figure*}

Figure~\ref{fig:pca_survey_wave} shows that questions cluster by survey wave and topic, indicating that the embedding space captures thematic similarity. Figures~\ref{fig:pca_abs_error} and~\ref{fig:pca_coverage} overlay imputation performance (using Qwen3-30B-A3B as the generator model) on this structure for 10-shot and 100-shot ICL. The results reveal that performance is not uniformly distributed across the embedding space: certain regions consistently exhibit higher absolute error or lower coverage, suggesting that question semantics, and by extension the topic domain, are a meaningful source of variation in ICL-based imputation quality. 10-shot and 100-shot ICL show similar spatial performance patterns (per-variable Spearman $\rho = 0.72$), and 100-shot ICL yields a slightly lower median absolute error (0.045 versus 0.052), mainly for the hardest questions.

\paragraph{Check with repetitions.}
To complement the embedding-space analysis with concrete case studies and to probe the single-repetition ($n_{\text{sim}}=1$) design used throughout (Section~\ref{sec:evaluation}), we select one variable from each extreme of the second principal component, the axis along which ICL performance varies most visibly in Figures~\ref{fig:pca_abs_error} and~\ref{fig:pca_coverage}: $v_{\text{TRAITBIZWF2C}}$ from W36, located in a region associated with poor ICL performance, and $v_{\text{VIDOFT}}$ from W45, located in a region associated with strong ICL performance. For each variable, we repeat the full simulate--ampute--impute--evaluate pipeline 100 times and report Monte Carlo standard errors \citep[MCSEs;][]{morrisUsingSimulationStudies2019, whiteMultipleImputationUsing2011} for bias, coverage, and CI width across five missingness settings (MCAR and left- and right-tailed MAR and MNAR) and eleven imputation methods (Tables~\ref{tab:study_var_level_bad_llm} and~\ref{tab:study_var_level_good_llm}). One zero-shot repetition for $v_{\text{VIDOFT}}$ under right-tailed MNAR failed to store its completed datasets, so that cell rests on 99 repetitions.

For these two variables, the standard deviation of a method's estimate across the 100 repetitions (ten times its bias MCSE) is 0.02--0.10, comparable to or larger than many differences between methods in a single condition. Single conditions therefore say little about which method is better, and our main comparison relies on aggregation over many conditions. Coverage MCSEs reach 0.05. This check covers two deliberately selected variables only and is descriptive rather than a formal hypothesis test. Repetitions across all 140 variables are left to future work.

Table~\ref{tab:study_var_level_bad_llm} shows results for $v_{\text{TRAITBIZWF2C}}$. The variable is in W36 with the topic \textit{``Gender and leadership''}. The question text of the variable reads: \textit{``In general, how do you think being ambitious impacts a woman's chances of getting a top executive business position?''}. For this variable, ICL produces substantially larger bias in absolute value than MICE PMM, MICE Forest, and TabPFN, and than simple baselines such as Mode Imputation in most missingness settings. Notably, the retrieval-only baseline (Most Similar Embeddings) often outperforms the full ICL pipeline, suggesting that for this variable, the generator model introduces systematic distortion rather than correcting retrieval errors. The coverage rates of both 10-shot and 100-shot ICL are far below the nominal 95\% level, confirming that the tighter confidence intervals produced by LLM-based imputation do not adequately reflect imputation uncertainty for this variable.

Table~\ref{tab:study_var_level_good_llm} presents a contrasting pattern for $v_{\text{VIDOFT}}$. The variable is in W45 with the topic \textit{Misinformation}. The question text of the variable reads: \textit{``How often do you come across videos or images that have been altered or made-up to mislead the public?
''}. Here, ICL performs considerably better: under MCAR, both 10-shot and 100-shot ICL achieve coverage rates comparable to MICE PMM (0.94--0.96 versus 0.93), with narrower confidence intervals but a larger bias ($-0.05$ versus 0.00). Under right-tailed MAR, 100-shot ICL achieves near-zero bias with near-perfect coverage, outperforming all baselines, including MICE PMM. Under MNAR, ICL maintains good coverage (at least 0.84) with competitive bias. Across both variables, MICE Forest undercovers (0.44--0.83) and TabPFN has a consistently negative bias. The comparison between these two variables illustrates that ICL's variable-level performance is heterogeneous: while some variables exhibit the systematic distortion seen for $v_{\text{TRAITBIZWF2C}}$, others benefit substantially from in-context learning, particularly under MAR and MNAR.

\begin{table*}[!tbp]
  \centering
  \footnotesize\renewcommand{\arraystretch}{0.92}
  \begin{tabular}{lllrrrrr}
\toprule
 &  &  & Bias & Bias MCSE & CR & CR MCSE & CI Width \\
Mechanism & Tail & Model &  &  &  &  &  \\
\midrule
\multirow[c]{11}{*}{MCAR} & \multirow[c]{11}{*}{RIGHT} & Full Data & -0.004 & 0.003 & 0.960 & 0.020 & 0.126 \\
 &  & Complete Case & -0.004 & 0.005 & 0.950 & 0.022 & 0.216 \\
 &  & Mode Imputation & {\cellcolor[HTML]{3CA959}} \color[HTML]{F1F1F1} -0.041 & 0.005 & {\cellcolor[HTML]{51B35E}} \color[HTML]{F1F1F1} 0.770 & 0.042 & {\cellcolor[HTML]{ABDB6D}} \color[HTML]{000000} 0.158 \\
 &  & Random Sample & {\cellcolor[HTML]{BBE278}} \color[HTML]{000000} -0.085 & 0.004 & {\cellcolor[HTML]{FDB365}} \color[HTML]{000000} 0.290 & 0.045 & {\cellcolor[HTML]{006837}} \color[HTML]{F1F1F1} 0.123 \\
 &  & MICE PMM & {\cellcolor[HTML]{006837}} \color[HTML]{F1F1F1} -0.006 & 0.005 & {\cellcolor[HTML]{006837}} \color[HTML]{F1F1F1} 0.930 & 0.026 & {\cellcolor[HTML]{A50026}} \color[HTML]{F1F1F1} 0.234 \\
 &  & MICE Forest & {\cellcolor[HTML]{73C264}} \color[HTML]{000000} 0.058 & 0.006 & {\cellcolor[HTML]{ABDB6D}} \color[HTML]{000000} 0.640 & 0.048 & {\cellcolor[HTML]{FFF8B4}} \color[HTML]{000000} 0.181 \\
 &  & TabPFN & {\cellcolor[HTML]{6BBF64}} \color[HTML]{000000} -0.055 & 0.005 & {\cellcolor[HTML]{48AE5C}} \color[HTML]{F1F1F1} 0.780 & 0.041 & {\cellcolor[HTML]{FDBD6D}} \color[HTML]{000000} 0.198 \\
 &  & Most Similar Embed. & {\cellcolor[HTML]{0D8044}} \color[HTML]{F1F1F1} -0.018 & 0.007 & {\cellcolor[HTML]{7DC765}} \color[HTML]{000000} 0.710 & 0.045 & {\cellcolor[HTML]{30A356}} \color[HTML]{F1F1F1} 0.138 \\
 &  & Zero-Shot LLM & {\cellcolor[HTML]{A50026}} \color[HTML]{F1F1F1} -0.239 & 0.003 & {\cellcolor[HTML]{A50026}} \color[HTML]{F1F1F1} 0.000 & 0.000 & {\cellcolor[HTML]{FDC372}} \color[HTML]{000000} 0.197 \\
 &  & ICL (k=10) & {\cellcolor[HTML]{FEE491}} \color[HTML]{000000} -0.143 & 0.005 & {\cellcolor[HTML]{F88C51}} \color[HTML]{F1F1F1} 0.230 & 0.042 & {\cellcolor[HTML]{FCAA5F}} \color[HTML]{000000} 0.202 \\
 &  & ICL (k=100) & {\cellcolor[HTML]{FEE695}} \color[HTML]{000000} -0.141 & 0.005 & {\cellcolor[HTML]{F47044}} \color[HTML]{F1F1F1} 0.190 & 0.039 & {\cellcolor[HTML]{FDBD6D}} \color[HTML]{000000} 0.198 \\
\midrule
\multirow[c]{22}{*}{MAR} & \multirow[c]{11}{*}{LEFT} & Full Data & -0.004 & 0.003 & 0.960 & 0.020 & 0.126 \\
 &  & Complete Case & -0.034 & 0.005 & 0.910 & 0.029 & 0.190 \\
 &  & Mode Imputation & {\cellcolor[HTML]{F99153}} \color[HTML]{000000} 0.185 & 0.004 & {\cellcolor[HTML]{A50026}} \color[HTML]{F1F1F1} 0.000 & 0.000 & {\cellcolor[HTML]{42AC5A}} \color[HTML]{F1F1F1} 0.144 \\
 &  & Random Sample & {\cellcolor[HTML]{FDB768}} \color[HTML]{000000} -0.171 & 0.004 & {\cellcolor[HTML]{A90426}} \color[HTML]{F1F1F1} 0.010 & 0.010 & {\cellcolor[HTML]{006837}} \color[HTML]{F1F1F1} 0.121 \\
 &  & MICE PMM & {\cellcolor[HTML]{006837}} \color[HTML]{F1F1F1} -0.019 & 0.006 & {\cellcolor[HTML]{006837}} \color[HTML]{F1F1F1} 0.930 & 0.026 & {\cellcolor[HTML]{A50026}} \color[HTML]{F1F1F1} 0.266 \\
 &  & MICE Forest & {\cellcolor[HTML]{39A758}} \color[HTML]{F1F1F1} 0.051 & 0.007 & {\cellcolor[HTML]{91D068}} \color[HTML]{000000} 0.680 & 0.047 & {\cellcolor[HTML]{DCF08F}} \color[HTML]{000000} 0.180 \\
 &  & TabPFN & {\cellcolor[HTML]{73C264}} \color[HTML]{000000} -0.068 & 0.005 & {\cellcolor[HTML]{B7E075}} \color[HTML]{000000} 0.620 & 0.049 & {\cellcolor[HTML]{FFFBB8}} \color[HTML]{000000} 0.196 \\
 &  & Most Similar Embed. & {\cellcolor[HTML]{06733D}} \color[HTML]{F1F1F1} 0.024 & 0.006 & {\cellcolor[HTML]{A0D669}} \color[HTML]{000000} 0.660 & 0.047 & {\cellcolor[HTML]{108647}} \color[HTML]{F1F1F1} 0.131 \\
 &  & Zero-Shot LLM & {\cellcolor[HTML]{A50026}} \color[HTML]{F1F1F1} -0.242 & 0.003 & {\cellcolor[HTML]{A50026}} \color[HTML]{F1F1F1} 0.000 & 0.000 & {\cellcolor[HTML]{ECF7A6}} \color[HTML]{000000} 0.187 \\
 &  & ICL (k=10) & {\cellcolor[HTML]{FDB567}} \color[HTML]{000000} -0.172 & 0.004 & {\cellcolor[HTML]{BB1526}} \color[HTML]{F1F1F1} 0.040 & 0.020 & {\cellcolor[HTML]{E8F59F}} \color[HTML]{000000} 0.185 \\
 &  & ICL (k=100) & {\cellcolor[HTML]{FDB567}} \color[HTML]{000000} -0.172 & 0.005 & {\cellcolor[HTML]{DC3B2C}} \color[HTML]{F1F1F1} 0.110 & 0.031 & {\cellcolor[HTML]{F7FCB4}} \color[HTML]{000000} 0.191 \\
\cmidrule(lr){2-8}
 & \multirow[c]{11}{*}{RIGHT} & Full Data & -0.004 & 0.003 & 0.960 & 0.020 & 0.126 \\
 &  & Complete Case & -0.028 & 0.005 & 0.900 & 0.030 & 0.202 \\
 &  & Mode Imputation & {\cellcolor[HTML]{A50026}} \color[HTML]{F1F1F1} -0.318 & 0.005 & {\cellcolor[HTML]{A50026}} \color[HTML]{F1F1F1} 0.000 & 0.000 & {\cellcolor[HTML]{91D068}} \color[HTML]{000000} 0.166 \\
 &  & Random Sample & {\cellcolor[HTML]{006837}} \color[HTML]{F1F1F1} 0.002 & 0.004 & {\cellcolor[HTML]{128A49}} \color[HTML]{F1F1F1} 0.890 & 0.031 & {\cellcolor[HTML]{006837}} \color[HTML]{F1F1F1} 0.123 \\
 &  & MICE PMM & {\cellcolor[HTML]{05713C}} \color[HTML]{F1F1F1} -0.009 & 0.007 & {\cellcolor[HTML]{006837}} \color[HTML]{F1F1F1} 0.960 & 0.020 & {\cellcolor[HTML]{A50026}} \color[HTML]{F1F1F1} 0.284 \\
 &  & MICE Forest & {\cellcolor[HTML]{39A758}} \color[HTML]{F1F1F1} 0.047 & 0.008 & {\cellcolor[HTML]{B7E075}} \color[HTML]{000000} 0.640 & 0.048 & {\cellcolor[HTML]{DCF08F}} \color[HTML]{000000} 0.189 \\
 &  & TabPFN & {\cellcolor[HTML]{75C465}} \color[HTML]{000000} -0.074 & 0.005 & {\cellcolor[HTML]{7FC866}} \color[HTML]{000000} 0.730 & 0.044 & {\cellcolor[HTML]{FEE593}} \color[HTML]{000000} 0.217 \\
 &  & Most Similar Embed. & {\cellcolor[HTML]{A2D76A}} \color[HTML]{000000} -0.096 & 0.008 & {\cellcolor[HTML]{FEDE89}} \color[HTML]{000000} 0.380 & 0.049 & {\cellcolor[HTML]{39A758}} \color[HTML]{F1F1F1} 0.146 \\
 &  & Zero-Shot LLM & {\cellcolor[HTML]{FBA05B}} \color[HTML]{000000} -0.230 & 0.003 & {\cellcolor[HTML]{A50026}} \color[HTML]{F1F1F1} 0.000 & 0.000 & {\cellcolor[HTML]{F5FBB2}} \color[HTML]{000000} 0.199 \\
 &  & ICL (k=10) & {\cellcolor[HTML]{FFFCBA}} \color[HTML]{000000} -0.163 & 0.004 & {\cellcolor[HTML]{D83128}} \color[HTML]{F1F1F1} 0.100 & 0.030 & {\cellcolor[HTML]{FFF3AC}} \color[HTML]{000000} 0.210 \\
 &  & ICL (k=100) & {\cellcolor[HTML]{FFF0A6}} \color[HTML]{000000} -0.176 & 0.004 & {\cellcolor[HTML]{BE1827}} \color[HTML]{F1F1F1} 0.050 & 0.022 & {\cellcolor[HTML]{FFFEBE}} \color[HTML]{000000} 0.204 \\
\midrule
\multirow[c]{22}{*}{MNAR} & \multirow[c]{11}{*}{LEFT} & Full Data & -0.004 & 0.003 & 0.960 & 0.020 & 0.126 \\
 &  & Complete Case & 0.026 & 0.005 & 0.930 & 0.026 & 0.205 \\
 &  & Mode Imputation & {\cellcolor[HTML]{1E9A51}} \color[HTML]{F1F1F1} -0.041 & 0.006 & {\cellcolor[HTML]{A9DA6C}} \color[HTML]{000000} 0.650 & 0.048 & {\cellcolor[HTML]{1E9A51}} \color[HTML]{F1F1F1} 0.133 \\
 &  & Random Sample & {\cellcolor[HTML]{7DC765}} \color[HTML]{000000} -0.064 & 0.004 & {\cellcolor[HTML]{FEFFBE}} \color[HTML]{000000} 0.470 & 0.050 & {\cellcolor[HTML]{006837}} \color[HTML]{F1F1F1} 0.122 \\
 &  & MICE PMM & {\cellcolor[HTML]{036E3A}} \color[HTML]{F1F1F1} 0.024 & 0.005 & {\cellcolor[HTML]{006837}} \color[HTML]{F1F1F1} 0.940 & 0.024 & {\cellcolor[HTML]{A50026}} \color[HTML]{F1F1F1} 0.223 \\
 &  & MICE Forest & {\cellcolor[HTML]{C5E67E}} \color[HTML]{000000} 0.086 & 0.005 & {\cellcolor[HTML]{FFF5AE}} \color[HTML]{000000} 0.440 & 0.050 & {\cellcolor[HTML]{F7FCB4}} \color[HTML]{000000} 0.170 \\
 &  & TabPFN & {\cellcolor[HTML]{08773F}} \color[HTML]{F1F1F1} -0.027 & 0.005 & {\cellcolor[HTML]{2DA155}} \color[HTML]{F1F1F1} 0.820 & 0.038 & {\cellcolor[HTML]{FEEB9D}} \color[HTML]{000000} 0.179 \\
 &  & Most Similar Embed. & {\cellcolor[HTML]{006837}} \color[HTML]{F1F1F1} 0.022 & 0.006 & {\cellcolor[HTML]{A5D86A}} \color[HTML]{000000} 0.660 & 0.047 & {\cellcolor[HTML]{06733D}} \color[HTML]{F1F1F1} 0.125 \\
 &  & Zero-Shot LLM & {\cellcolor[HTML]{A50026}} \color[HTML]{F1F1F1} -0.201 & 0.003 & {\cellcolor[HTML]{A50026}} \color[HTML]{F1F1F1} 0.000 & 0.000 & {\cellcolor[HTML]{FEEDA1}} \color[HTML]{000000} 0.178 \\
 &  & ICL (k=10) & {\cellcolor[HTML]{FFF3AC}} \color[HTML]{000000} -0.117 & 0.004 & {\cellcolor[HTML]{F7814C}} \color[HTML]{F1F1F1} 0.220 & 0.041 & {\cellcolor[HTML]{FEE28F}} \color[HTML]{000000} 0.182 \\
 &  & ICL (k=100) & {\cellcolor[HTML]{FFF8B4}} \color[HTML]{000000} -0.115 & 0.005 & {\cellcolor[HTML]{FDC776}} \color[HTML]{000000} 0.330 & 0.047 & {\cellcolor[HTML]{FEDE89}} \color[HTML]{000000} 0.183 \\
\cmidrule(lr){2-8}
 & \multirow[c]{11}{*}{RIGHT} & Full Data & -0.004 & 0.003 & 0.960 & 0.020 & 0.126 \\
 &  & Complete Case & -0.065 & 0.007 & 0.890 & 0.031 & 0.301 \\
 &  & Mode Imputation & {\cellcolor[HTML]{B7E075}} \color[HTML]{000000} -0.114 & 0.007 & {\cellcolor[HTML]{EEF8A8}} \color[HTML]{000000} 0.480 & 0.050 & {\cellcolor[HTML]{FEEFA3}} \color[HTML]{000000} 0.238 \\
 &  & Random Sample & {\cellcolor[HTML]{A7D96B}} \color[HTML]{000000} -0.105 & 0.004 & {\cellcolor[HTML]{F67F4B}} \color[HTML]{F1F1F1} 0.200 & 0.040 & {\cellcolor[HTML]{006837}} \color[HTML]{F1F1F1} 0.132 \\
 &  & MICE PMM & {\cellcolor[HTML]{48AE5C}} \color[HTML]{F1F1F1} -0.065 & 0.007 & {\cellcolor[HTML]{006837}} \color[HTML]{F1F1F1} 0.880 & 0.032 & {\cellcolor[HTML]{A50026}} \color[HTML]{F1F1F1} 0.324 \\
 &  & MICE Forest & {\cellcolor[HTML]{006837}} \color[HTML]{F1F1F1} -0.019 & 0.010 & {\cellcolor[HTML]{1B9950}} \color[HTML]{F1F1F1} 0.790 & 0.041 & {\cellcolor[HTML]{FDB365}} \color[HTML]{000000} 0.264 \\
 &  & TabPFN & {\cellcolor[HTML]{98D368}} \color[HTML]{000000} -0.099 & 0.006 & {\cellcolor[HTML]{84CA66}} \color[HTML]{000000} 0.660 & 0.047 & {\cellcolor[HTML]{F8864F}} \color[HTML]{F1F1F1} 0.278 \\
 &  & Most Similar Embed. & {\cellcolor[HTML]{A7D96B}} \color[HTML]{000000} -0.105 & 0.009 & {\cellcolor[HTML]{DCF08F}} \color[HTML]{000000} 0.520 & 0.050 & {\cellcolor[HTML]{F2FAAE}} \color[HTML]{000000} 0.222 \\
 &  & Zero-Shot LLM & {\cellcolor[HTML]{A50026}} \color[HTML]{F1F1F1} -0.305 & 0.003 & {\cellcolor[HTML]{A50026}} \color[HTML]{F1F1F1} 0.000 & 0.000 & {\cellcolor[HTML]{FFF8B4}} \color[HTML]{000000} 0.232 \\
 &  & ICL (k=10) & {\cellcolor[HTML]{FEE491}} \color[HTML]{000000} -0.187 & 0.005 & {\cellcolor[HTML]{F57748}} \color[HTML]{F1F1F1} 0.190 & 0.039 & {\cellcolor[HTML]{FDB567}} \color[HTML]{000000} 0.264 \\
 &  & ICL (k=100) & {\cellcolor[HTML]{FEEC9F}} \color[HTML]{000000} -0.179 & 0.005 & {\cellcolor[HTML]{FA9B58}} \color[HTML]{000000} 0.240 & 0.043 & {\cellcolor[HTML]{FBA35C}} \color[HTML]{000000} 0.270 \\
\bottomrule
\end{tabular}

  \caption{\textbf{Variable-level imputation results for $v_{\text{TRAITBIZWF2C}}$ (W36), a variable selected as a poorly performing case for the LLM-based imputer in the main comparison.} The variable was chosen from the region of the question-embedding PCA space associated with high ICL absolute error and low coverage (Figures~\ref{fig:pca_abs_error} and~\ref{fig:pca_coverage}). Bias: estimated minus true regression coefficient (values closer to 0 indicate lower bias). CR: coverage rate of the 95\% confidence interval ($\uparrow$, nominal = 0.95). CI Width: average width of the 95\% confidence interval ($\downarrow$). MCSE: Monte Carlo standard error over 100 repetitions. Each repetition draws $N=500$ observations with 50\% missingness. LEFT/RIGHT: left- and right-tailed MAR and MNAR. Generator: Qwen3-30B-A3B. Retriever: EmbeddingGemma-300M.}
  \label{tab:study_var_level_bad_llm}
\end{table*}

\begin{table*}[!tbp]
  \centering
  \footnotesize\renewcommand{\arraystretch}{0.92}
  \begin{tabular}{lllrrrrr}
\toprule
 &  &  & Bias & Bias MCSE & CR & CR MCSE & CI Width \\
Mechanism & Tail & Model &  &  &  &  &  \\
\midrule
\multirow[c]{11}{*}{MCAR} & \multirow[c]{11}{*}{RIGHT} & Full Data & 0.004 & 0.004 & 0.920 & 0.027 & 0.168 \\
 &  & Complete Case & 0.004 & 0.006 & 0.960 & 0.020 & 0.245 \\
 &  & Mode Imputation & {\cellcolor[HTML]{07753E}} \color[HTML]{F1F1F1} -0.004 & 0.006 & {\cellcolor[HTML]{05713C}} \color[HTML]{F1F1F1} 0.960 & 0.020 & {\cellcolor[HTML]{ED5F3C}} \color[HTML]{F1F1F1} 0.232 \\
 &  & Random Sample & {\cellcolor[HTML]{CA2427}} \color[HTML]{F1F1F1} -0.070 & 0.003 & {\cellcolor[HTML]{A50026}} \color[HTML]{F1F1F1} 0.460 & 0.050 & {\cellcolor[HTML]{006837}} \color[HTML]{F1F1F1} 0.131 \\
 &  & MICE PMM & {\cellcolor[HTML]{006837}} \color[HTML]{F1F1F1} 0.002 & 0.006 & {\cellcolor[HTML]{148E4B}} \color[HTML]{F1F1F1} 0.930 & 0.026 & {\cellcolor[HTML]{A50026}} \color[HTML]{F1F1F1} 0.254 \\
 &  & MICE Forest & {\cellcolor[HTML]{E2F397}} \color[HTML]{000000} 0.033 & 0.008 & {\cellcolor[HTML]{C1E57B}} \color[HTML]{000000} 0.790 & 0.041 & {\cellcolor[HTML]{FDBD6D}} \color[HTML]{000000} 0.214 \\
 &  & TabPFN & {\cellcolor[HTML]{ECF7A6}} \color[HTML]{000000} -0.035 & 0.004 & {\cellcolor[HTML]{006837}} \color[HTML]{F1F1F1} 0.970 & 0.017 & {\cellcolor[HTML]{ED5F3C}} \color[HTML]{F1F1F1} 0.232 \\
 &  & Most Similar Embed. & {\cellcolor[HTML]{0E8245}} \color[HTML]{F1F1F1} 0.006 & 0.006 & {\cellcolor[HTML]{199750}} \color[HTML]{F1F1F1} 0.920 & 0.027 & {\cellcolor[HTML]{FDC776}} \color[HTML]{000000} 0.211 \\
 &  & Zero-Shot LLM & {\cellcolor[HTML]{A50026}} \color[HTML]{F1F1F1} -0.075 & 0.003 & {\cellcolor[HTML]{96D268}} \color[HTML]{000000} 0.830 & 0.038 & {\cellcolor[HTML]{FED07E}} \color[HTML]{000000} 0.209 \\
 &  & ICL (k=10) & {\cellcolor[HTML]{FED07E}} \color[HTML]{000000} -0.048 & 0.003 & {\cellcolor[HTML]{0F8446}} \color[HTML]{F1F1F1} 0.940 & 0.024 & {\cellcolor[HTML]{FECC7B}} \color[HTML]{000000} 0.210 \\
 &  & ICL (k=100) & {\cellcolor[HTML]{FED884}} \color[HTML]{000000} -0.047 & 0.003 & {\cellcolor[HTML]{05713C}} \color[HTML]{F1F1F1} 0.960 & 0.020 & {\cellcolor[HTML]{FEDC88}} \color[HTML]{000000} 0.206 \\
\midrule
\multirow[c]{22}{*}{MAR} & \multirow[c]{11}{*}{LEFT} & Full Data & 0.004 & 0.004 & 0.920 & 0.027 & 0.168 \\
 &  & Complete Case & -0.013 & 0.007 & 0.900 & 0.030 & 0.227 \\
 &  & Mode Imputation & {\cellcolor[HTML]{A50026}} \color[HTML]{F1F1F1} 0.134 & 0.006 & {\cellcolor[HTML]{FCA85E}} \color[HTML]{000000} 0.330 & 0.047 & {\cellcolor[HTML]{FEE593}} \color[HTML]{000000} 0.231 \\
 &  & Random Sample & {\cellcolor[HTML]{E24731}} \color[HTML]{F1F1F1} -0.116 & 0.004 & {\cellcolor[HTML]{A50026}} \color[HTML]{F1F1F1} 0.090 & 0.029 & {\cellcolor[HTML]{006837}} \color[HTML]{F1F1F1} 0.131 \\
 &  & MICE PMM & {\cellcolor[HTML]{006837}} \color[HTML]{F1F1F1} -0.003 & 0.008 & {\cellcolor[HTML]{006837}} \color[HTML]{F1F1F1} 0.920 & 0.027 & {\cellcolor[HTML]{A50026}} \color[HTML]{F1F1F1} 0.301 \\
 &  & MICE Forest & {\cellcolor[HTML]{42AC5A}} \color[HTML]{F1F1F1} 0.023 & 0.010 & {\cellcolor[HTML]{87CB67}} \color[HTML]{000000} 0.710 & 0.045 & {\cellcolor[HTML]{FFF6B0}} \color[HTML]{000000} 0.222 \\
 &  & TabPFN & {\cellcolor[HTML]{B3DF72}} \color[HTML]{000000} -0.046 & 0.005 & {\cellcolor[HTML]{006837}} \color[HTML]{F1F1F1} 0.920 & 0.027 & {\cellcolor[HTML]{FDB96A}} \color[HTML]{000000} 0.246 \\
 &  & Most Similar Embed. & {\cellcolor[HTML]{FAFDB8}} \color[HTML]{000000} 0.067 & 0.007 & {\cellcolor[HTML]{BFE47A}} \color[HTML]{000000} 0.630 & 0.048 & {\cellcolor[HTML]{F1F9AC}} \color[HTML]{000000} 0.210 \\
 &  & Zero-Shot LLM & {\cellcolor[HTML]{F57748}} \color[HTML]{F1F1F1} -0.106 & 0.003 & {\cellcolor[HTML]{F8FCB6}} \color[HTML]{000000} 0.520 & 0.050 & {\cellcolor[HTML]{FFFCBA}} \color[HTML]{000000} 0.218 \\
 &  & ICL (k=10) & {\cellcolor[HTML]{FDBD6D}} \color[HTML]{000000} -0.091 & 0.003 & {\cellcolor[HTML]{C5E67E}} \color[HTML]{000000} 0.620 & 0.049 & {\cellcolor[HTML]{F1F9AC}} \color[HTML]{000000} 0.210 \\
 &  & ICL (k=100) & {\cellcolor[HTML]{F88950}} \color[HTML]{F1F1F1} -0.102 & 0.004 & {\cellcolor[HTML]{D1EC86}} \color[HTML]{000000} 0.600 & 0.049 & {\cellcolor[HTML]{F8FCB6}} \color[HTML]{000000} 0.213 \\
\cmidrule(lr){2-8}
 & \multirow[c]{11}{*}{RIGHT} & Full Data & 0.004 & 0.004 & 0.920 & 0.027 & 0.168 \\
 &  & Complete Case & -0.017 & 0.006 & 0.940 & 0.024 & 0.216 \\
 &  & Mode Imputation & {\cellcolor[HTML]{A50026}} \color[HTML]{F1F1F1} -0.133 & 0.006 & {\cellcolor[HTML]{A50026}} \color[HTML]{F1F1F1} 0.360 & 0.048 & {\cellcolor[HTML]{FDC171}} \color[HTML]{000000} 0.231 \\
 &  & Random Sample & {\cellcolor[HTML]{48AE5C}} \color[HTML]{F1F1F1} -0.023 & 0.003 & {\cellcolor[HTML]{39A758}} \color[HTML]{F1F1F1} 0.900 & 0.030 & {\cellcolor[HTML]{006837}} \color[HTML]{F1F1F1} 0.129 \\
 &  & MICE PMM & {\cellcolor[HTML]{108647}} \color[HTML]{F1F1F1} -0.009 & 0.006 & {\cellcolor[HTML]{08773F}} \color[HTML]{F1F1F1} 0.970 & 0.017 & {\cellcolor[HTML]{A50026}} \color[HTML]{F1F1F1} 0.283 \\
 &  & MICE Forest & {\cellcolor[HTML]{45AD5B}} \color[HTML]{F1F1F1} 0.022 & 0.008 & {\cellcolor[HTML]{A7D96B}} \color[HTML]{000000} 0.800 & 0.040 & {\cellcolor[HTML]{FEE593}} \color[HTML]{000000} 0.219 \\
 &  & TabPFN & {\cellcolor[HTML]{B5DF74}} \color[HTML]{000000} -0.045 & 0.004 & {\cellcolor[HTML]{108647}} \color[HTML]{F1F1F1} 0.950 & 0.022 & {\cellcolor[HTML]{FED27F}} \color[HTML]{000000} 0.226 \\
 &  & Most Similar Embed. & {\cellcolor[HTML]{FA9857}} \color[HTML]{000000} -0.097 & 0.006 & {\cellcolor[HTML]{FECE7C}} \color[HTML]{000000} 0.590 & 0.049 & {\cellcolor[HTML]{FFFAB6}} \color[HTML]{000000} 0.209 \\
 &  & Zero-Shot LLM & {\cellcolor[HTML]{D1EC86}} \color[HTML]{000000} -0.052 & 0.002 & {\cellcolor[HTML]{08773F}} \color[HTML]{F1F1F1} 0.970 & 0.017 & {\cellcolor[HTML]{F2FAAE}} \color[HTML]{000000} 0.201 \\
 &  & ICL (k=10) & {\cellcolor[HTML]{219C52}} \color[HTML]{F1F1F1} -0.016 & 0.003 & {\cellcolor[HTML]{006837}} \color[HTML]{F1F1F1} 0.990 & 0.010 & {\cellcolor[HTML]{FFFDBC}} \color[HTML]{000000} 0.207 \\
 &  & ICL (k=100) & {\cellcolor[HTML]{006837}} \color[HTML]{F1F1F1} -0.001 & 0.003 & {\cellcolor[HTML]{006837}} \color[HTML]{F1F1F1} 0.990 & 0.010 & {\cellcolor[HTML]{FFFCBA}} \color[HTML]{000000} 0.208 \\
\midrule
\multirow[c]{22}{*}{MNAR} & \multirow[c]{11}{*}{LEFT} & Full Data & 0.004 & 0.004 & 0.920 & 0.027 & 0.168 \\
 &  & Complete Case & -0.036 & 0.007 & 0.930 & 0.026 & 0.256 \\
 &  & Mode Imputation & {\cellcolor[HTML]{DFF293}} \color[HTML]{000000} -0.043 & 0.006 & {\cellcolor[HTML]{279F53}} \color[HTML]{F1F1F1} 0.860 & 0.035 & {\cellcolor[HTML]{FECA79}} \color[HTML]{000000} 0.224 \\
 &  & Random Sample & {\cellcolor[HTML]{DB382B}} \color[HTML]{F1F1F1} -0.081 & 0.004 & {\cellcolor[HTML]{A50026}} \color[HTML]{F1F1F1} 0.350 & 0.048 & {\cellcolor[HTML]{006837}} \color[HTML]{F1F1F1} 0.134 \\
 &  & MICE PMM & {\cellcolor[HTML]{BDE379}} \color[HTML]{000000} -0.037 & 0.006 & {\cellcolor[HTML]{006837}} \color[HTML]{F1F1F1} 0.930 & 0.026 & {\cellcolor[HTML]{A50026}} \color[HTML]{F1F1F1} 0.275 \\
 &  & MICE Forest & {\cellcolor[HTML]{006837}} \color[HTML]{F1F1F1} -0.009 & 0.009 & {\cellcolor[HTML]{51B35E}} \color[HTML]{F1F1F1} 0.830 & 0.038 & {\cellcolor[HTML]{FDC171}} \color[HTML]{000000} 0.227 \\
 &  & TabPFN & {\cellcolor[HTML]{FED07E}} \color[HTML]{000000} -0.061 & 0.004 & {\cellcolor[HTML]{0D8044}} \color[HTML]{F1F1F1} 0.900 & 0.030 & {\cellcolor[HTML]{F46D43}} \color[HTML]{F1F1F1} 0.246 \\
 &  & Most Similar Embed. & {\cellcolor[HTML]{0E8245}} \color[HTML]{F1F1F1} -0.014 & 0.007 & {\cellcolor[HTML]{51B35E}} \color[HTML]{F1F1F1} 0.830 & 0.038 & {\cellcolor[HTML]{F1F9AC}} \color[HTML]{000000} 0.199 \\
 &  & Zero-Shot LLM & {\cellcolor[HTML]{A50026}} \color[HTML]{F1F1F1} -0.091 & 0.003 & {\cellcolor[HTML]{E5F49B}} \color[HTML]{000000} 0.680 & 0.047 & {\cellcolor[HTML]{FEE999}} \color[HTML]{000000} 0.215 \\
 &  & ICL (k=10) & {\cellcolor[HTML]{FBA35C}} \color[HTML]{000000} -0.068 & 0.003 & {\cellcolor[HTML]{42AC5A}} \color[HTML]{F1F1F1} 0.840 & 0.037 & {\cellcolor[HTML]{FEEC9F}} \color[HTML]{000000} 0.213 \\
 &  & ICL (k=100) & {\cellcolor[HTML]{FDAF62}} \color[HTML]{000000} -0.066 & 0.003 & {\cellcolor[HTML]{118848}} \color[HTML]{F1F1F1} 0.890 & 0.031 & {\cellcolor[HTML]{FED07E}} \color[HTML]{000000} 0.223 \\
\cmidrule(lr){2-8}
 & \multirow[c]{11}{*}{RIGHT} & Full Data & 0.004 & 0.004 & 0.920 & 0.027 & 0.168 \\
 &  & Complete Case & 0.049 & 0.007 & 0.930 & 0.026 & 0.281 \\
 &  & Mode Imputation & {\cellcolor[HTML]{FEDA86}} \color[HTML]{000000} 0.053 & 0.007 & {\cellcolor[HTML]{78C565}} \color[HTML]{000000} 0.920 & 0.027 & {\cellcolor[HTML]{E34933}} \color[HTML]{F1F1F1} 0.274 \\
 &  & Random Sample & {\cellcolor[HTML]{FEDA86}} \color[HTML]{000000} -0.053 & 0.003 & {\cellcolor[HTML]{A50026}} \color[HTML]{F1F1F1} 0.650 & 0.048 & {\cellcolor[HTML]{006837}} \color[HTML]{F1F1F1} 0.136 \\
 &  & MICE PMM & {\cellcolor[HTML]{ECF7A6}} \color[HTML]{000000} 0.040 & 0.007 & {\cellcolor[HTML]{39A758}} \color[HTML]{F1F1F1} 0.950 & 0.022 & {\cellcolor[HTML]{A50026}} \color[HTML]{F1F1F1} 0.297 \\
 &  & MICE Forest & {\cellcolor[HTML]{A50026}} \color[HTML]{F1F1F1} 0.086 & 0.008 & {\cellcolor[HTML]{A50026}} \color[HTML]{F1F1F1} 0.650 & 0.048 & {\cellcolor[HTML]{FDB163}} \color[HTML]{000000} 0.248 \\
 &  & TabPFN & {\cellcolor[HTML]{006837}} \color[HTML]{F1F1F1} -0.002 & 0.005 & {\cellcolor[HTML]{0E8245}} \color[HTML]{F1F1F1} 0.980 & 0.014 & {\cellcolor[HTML]{EC5C3B}} \color[HTML]{F1F1F1} 0.269 \\
 &  & Most Similar Embed. & {\cellcolor[HTML]{F8FCB6}} \color[HTML]{000000} 0.042 & 0.008 & {\cellcolor[HTML]{E3F399}} \color[HTML]{000000} 0.850 & 0.036 & {\cellcolor[HTML]{FDB163}} \color[HTML]{000000} 0.248 \\
 &  & Zero-Shot LLM & {\cellcolor[HTML]{FA9656}} \color[HTML]{000000} -0.064 & 0.002 & {\cellcolor[HTML]{9DD569}} \color[HTML]{000000} 0.899 & 0.030 & {\cellcolor[HTML]{ECF7A6}} \color[HTML]{000000} 0.209 \\
 &  & ICL (k=10) & {\cellcolor[HTML]{CBE982}} \color[HTML]{000000} -0.033 & 0.003 & {\cellcolor[HTML]{0E8245}} \color[HTML]{F1F1F1} 0.980 & 0.014 & {\cellcolor[HTML]{FFF7B2}} \color[HTML]{000000} 0.221 \\
 &  & ICL (k=100) & {\cellcolor[HTML]{ADDC6F}} \color[HTML]{000000} -0.028 & 0.003 & {\cellcolor[HTML]{006837}} \color[HTML]{F1F1F1} 1.000 & 0.000 & {\cellcolor[HTML]{FFF5AE}} \color[HTML]{000000} 0.222 \\
\bottomrule
\end{tabular}

  \caption{\textbf{Variable-level imputation results for $v_{\text{VIDOFT}}$ (W45), a variable selected as a strongly performing case for the LLM-based imputer in the main comparison.} The variable was chosen from the region of the question-embedding PCA space associated with low ICL absolute error and high coverage (Figures~\ref{fig:pca_abs_error} and~\ref{fig:pca_coverage}). Bias: estimated minus true regression coefficient (values closer to 0 indicate lower bias). CR: coverage rate of the 95\% confidence interval ($\uparrow$, nominal = 0.95). CI Width: average width of the 95\% confidence interval ($\downarrow$). MCSE: Monte Carlo standard error over 100 repetitions. Each repetition draws $N=500$ observations with 50\% missingness. LEFT/RIGHT: left- and right-tailed MAR and MNAR. Generator: Qwen3-30B-A3B. Retriever: EmbeddingGemma-300M.}
  \label{tab:study_var_level_good_llm}
\end{table*}

\clearpage
\twocolumn[{%
\section{Full Results and Significance Tests}\label{app:full_results}
\noindent
\begin{minipage}[t]{0.54\textwidth}\vspace{0pt}
  \centering
  \tiny\renewcommand{\arraystretch}{0.74}
  \begin{tabular}{llrr@{\hspace{1.2em}}rr@{\hspace{1.2em}}r}
    \toprule
    & & \multicolumn{2}{c}{\textbf{Absolute Error}} & \multicolumn{2}{c}{\textbf{CI Width}} & \\
    \cmidrule(lr){3-4}\cmidrule(lr){5-6}
    \textbf{Miss.} & \textbf{Method} & {Median} & {IQR} & {Median} & {IQR} & \textbf{Coverage} \\
\midrule
\midrule
\multirow{28}{*}{\textbf{MCAR}} & \textit{Full Data} & 0.026 & 0.030 & 0.156 & 0.016 & 0.964 \\
 & \textit{Complete Case} & 0.040 & 0.043 & 0.231 & 0.024 & 0.957 \\
& \multicolumn{6}{l}{\textit{Zero-Shot LLM}} \\
 & Qwen3-30B-A3B & {\cellcolor[HTML]{8ECF67}} \color[HTML]{000000} 0.064 & 0.098 & {\cellcolor[HTML]{199750}} \color[HTML]{F1F1F1} 0.183 & 0.043 & {\cellcolor[HTML]{FFFCBA}} \color[HTML]{000000} 0.636 \\
 & Olmo-3-7B-Instruct & {\cellcolor[HTML]{B9E176}} \color[HTML]{000000} 0.072 & 0.120 & {\cellcolor[HTML]{199750}} \color[HTML]{F1F1F1} 0.183 & 0.058 & {\cellcolor[HTML]{FFF8B4}} \color[HTML]{000000} 0.629 \\
 & GLM-4.7-Flash & {\cellcolor[HTML]{AFDD70}} \color[HTML]{000000} 0.070 & 0.101 & {\cellcolor[HTML]{48AE5C}} \color[HTML]{F1F1F1} 0.189 & 0.054 & {\cellcolor[HTML]{FFFCBA}} \color[HTML]{000000} 0.636 \\
 & Qwen3-30B-A3B-Thinking & {\cellcolor[HTML]{CFEB85}} \color[HTML]{000000} 0.077 & 0.093 & {\cellcolor[HTML]{006837}} \color[HTML]{F1F1F1} 0.173 & 0.048 & {\cellcolor[HTML]{FDC372}} \color[HTML]{000000} 0.543 \\
 & gpt-oss-120b & {\cellcolor[HTML]{17934E}} \color[HTML]{F1F1F1} 0.045 & 0.065 & {\cellcolor[HTML]{148E4B}} \color[HTML]{F1F1F1} 0.181 & 0.050 & {\cellcolor[HTML]{ABDB6D}} \color[HTML]{000000} 0.763 \\
 & Qwen3-8B & {\cellcolor[HTML]{A0D669}} \color[HTML]{000000} 0.067 & 0.093 & {\cellcolor[HTML]{17934E}} \color[HTML]{F1F1F1} 0.182 & 0.057 & {\cellcolor[HTML]{FFF8B4}} \color[HTML]{000000} 0.629 \\
& \multicolumn{6}{l}{\textit{ICL (k=10)}} \\
 & Qwen3-30B-A3B & {\cellcolor[HTML]{108647}} \color[HTML]{F1F1F1} 0.042 & 0.059 & {\cellcolor[HTML]{82C966}} \color[HTML]{000000} 0.197 & 0.046 & {\cellcolor[HTML]{4EB15D}} \color[HTML]{F1F1F1} 0.850 \\
 & Olmo-3-7B-Instruct & {\cellcolor[HTML]{70C164}} \color[HTML]{000000} 0.059 & 0.100 & {\cellcolor[HTML]{60BA62}} \color[HTML]{F1F1F1} 0.192 & 0.053 & {\cellcolor[HTML]{F4FAB0}} \color[HTML]{000000} 0.662 \\
 & GLM-4.7-Flash & {\cellcolor[HTML]{128A49}} \color[HTML]{F1F1F1} 0.043 & 0.067 & {\cellcolor[HTML]{7AC665}} \color[HTML]{000000} 0.196 & 0.039 & {\cellcolor[HTML]{84CA66}} \color[HTML]{000000} 0.800 \\
 & Qwen3-30B-A3B-Thinking & {\cellcolor[HTML]{17934E}} \color[HTML]{F1F1F1} 0.045 & 0.065 & {\cellcolor[HTML]{7AC665}} \color[HTML]{000000} 0.196 & 0.047 & {\cellcolor[HTML]{7FC866}} \color[HTML]{000000} 0.807 \\
 & gpt-oss-120b & {\cellcolor[HTML]{026C39}} \color[HTML]{F1F1F1} 0.036 & 0.048 & {\cellcolor[HTML]{82C966}} \color[HTML]{000000} 0.197 & 0.048 & {\cellcolor[HTML]{148E4B}} \color[HTML]{F1F1F1} 0.907 \\
 & Qwen3-8B & {\cellcolor[HTML]{42AC5A}} \color[HTML]{F1F1F1} 0.052 & 0.058 & {\cellcolor[HTML]{39A758}} \color[HTML]{F1F1F1} 0.187 & 0.048 & {\cellcolor[HTML]{78C565}} \color[HTML]{000000} 0.814 \\
& \multicolumn{6}{l}{\textit{ICL (k=100)}} \\
 & Qwen3-30B-A3B & {\cellcolor[HTML]{026C39}} \color[HTML]{F1F1F1} 0.036 & 0.055 & {\cellcolor[HTML]{B1DE71}} \color[HTML]{000000} 0.205 & 0.054 & {\cellcolor[HTML]{57B65F}} \color[HTML]{F1F1F1} 0.843 \\
 & Olmo-3-7B-Instruct & {\cellcolor[HTML]{89CC67}} \color[HTML]{000000} 0.063 & 0.096 & {\cellcolor[HTML]{60BA62}} \color[HTML]{F1F1F1} 0.192 & 0.045 & {\cellcolor[HTML]{F4FAB0}} \color[HTML]{000000} 0.662 \\
 & GLM-4.7-Flash & {\cellcolor[HTML]{6BBF64}} \color[HTML]{000000} 0.058 & 0.080 & {\cellcolor[HTML]{ADDC6F}} \color[HTML]{000000} 0.204 & 0.067 & {\cellcolor[HTML]{8CCD67}} \color[HTML]{000000} 0.793 \\
 & Qwen3-30B-A3B-Thinking & {\cellcolor[HTML]{108647}} \color[HTML]{F1F1F1} 0.042 & 0.061 & {\cellcolor[HTML]{82C966}} \color[HTML]{000000} 0.197 & 0.048 & {\cellcolor[HTML]{69BE63}} \color[HTML]{F1F1F1} 0.829 \\
 & gpt-oss-120b & {\cellcolor[HTML]{006837}} \color[HTML]{F1F1F1} 0.035 & 0.048 & {\cellcolor[HTML]{6EC064}} \color[HTML]{000000} 0.194 & 0.054 & {\cellcolor[HTML]{148E4B}} \color[HTML]{F1F1F1} 0.906 \\
 & Qwen3-8B & {\cellcolor[HTML]{108647}} \color[HTML]{F1F1F1} 0.042 & 0.053 & {\cellcolor[HTML]{B1DE71}} \color[HTML]{000000} 0.205 & 0.051 & {\cellcolor[HTML]{2AA054}} \color[HTML]{F1F1F1} 0.879 \\
& \multicolumn{6}{l}{\textit{Baselines}} \\
 & Most Similar Embed. & {\cellcolor[HTML]{128A49}} \color[HTML]{F1F1F1} 0.043 & 0.047 & {\cellcolor[HTML]{148E4B}} \color[HTML]{F1F1F1} 0.181 & 0.031 & {\cellcolor[HTML]{45AD5B}} \color[HTML]{F1F1F1} 0.857 \\
 & MICE PMM & {\cellcolor[HTML]{06733D}} \color[HTML]{F1F1F1} 0.038 & 0.048 & {\cellcolor[HTML]{FEE18D}} \color[HTML]{000000} 0.232 & 0.076 & {\cellcolor[HTML]{006837}} \color[HTML]{F1F1F1} 0.957 \\
 & MICE Forest & {\cellcolor[HTML]{57B65F}} \color[HTML]{F1F1F1} 0.055 & 0.064 & {\cellcolor[HTML]{60BA62}} \color[HTML]{F1F1F1} 0.192 & 0.047 & {\cellcolor[HTML]{84CA66}} \color[HTML]{000000} 0.800 \\
 & Mode Imputation & 0.039 & 0.045 & 0.206 & 0.033 & 0.871 \\
 & Random Sample & 0.082 & 0.142 & 0.124 & 0.029 & 0.379 \\
& \multicolumn{6}{l}{\textit{TabPFN}} \\
 & TabPFN (greedy) & {\cellcolor[HTML]{8ECF67}} \color[HTML]{000000} 0.064 & 0.090 & {\cellcolor[HTML]{7AC665}} \color[HTML]{000000} 0.196 & 0.044 & {\cellcolor[HTML]{F7FCB4}} \color[HTML]{000000} 0.657 \\
 & TabPFN (T=0.5) & {\cellcolor[HTML]{17934E}} \color[HTML]{F1F1F1} 0.045 & 0.060 & {\cellcolor[HTML]{BBE278}} \color[HTML]{000000} 0.207 & 0.042 & {\cellcolor[HTML]{78C565}} \color[HTML]{000000} 0.814 \\
 & TabPFN (T=1.0) & {\cellcolor[HTML]{097940}} \color[HTML]{F1F1F1} 0.039 & 0.038 & {\cellcolor[HTML]{CBE982}} \color[HTML]{000000} 0.210 & 0.055 & {\cellcolor[HTML]{05713C}} \color[HTML]{F1F1F1} 0.943 \\
\midrule
\multirow{28}{*}{\textbf{MAR}} & \textit{Full Data} & 0.026 & 0.030 & 0.156 & 0.016 & 0.964 \\
 & \textit{Complete Case} & 0.046 & 0.048 & 0.213 & 0.027 & 0.914 \\
& \multicolumn{6}{l}{\textit{Zero-Shot LLM}} \\
 & Qwen3-30B-A3B & {\cellcolor[HTML]{E0F295}} \color[HTML]{000000} 0.081 & 0.112 & {\cellcolor[HTML]{48AE5C}} \color[HTML]{F1F1F1} 0.189 & 0.053 & {\cellcolor[HTML]{FEE797}} \color[HTML]{000000} 0.596 \\
 & Olmo-3-7B-Instruct & {\cellcolor[HTML]{F67F4B}} \color[HTML]{F1F1F1} 0.120 & 0.121 & {\cellcolor[HTML]{199750}} \color[HTML]{F1F1F1} 0.183 & 0.053 & {\cellcolor[HTML]{D22B27}} \color[HTML]{F1F1F1} 0.387 \\
 & GLM-4.7-Flash & {\cellcolor[HTML]{F8FCB6}} \color[HTML]{000000} 0.088 & 0.102 & {\cellcolor[HTML]{66BD63}} \color[HTML]{F1F1F1} 0.193 & 0.063 & {\cellcolor[HTML]{FDC372}} \color[HTML]{000000} 0.543 \\
 & Qwen3-30B-A3B-Thinking & {\cellcolor[HTML]{FDC171}} \color[HTML]{000000} 0.108 & 0.119 & {\cellcolor[HTML]{05713C}} \color[HTML]{F1F1F1} 0.175 & 0.059 & {\cellcolor[HTML]{E0422F}} \color[HTML]{F1F1F1} 0.411 \\
 & gpt-oss-120b & {\cellcolor[HTML]{8ECF67}} \color[HTML]{000000} 0.064 & 0.077 & {\cellcolor[HTML]{199750}} \color[HTML]{F1F1F1} 0.183 & 0.051 & {\cellcolor[HTML]{FAFDB8}} \color[HTML]{000000} 0.651 \\
 & Qwen3-8B & {\cellcolor[HTML]{FEE28F}} \color[HTML]{000000} 0.100 & 0.127 & {\cellcolor[HTML]{39A758}} \color[HTML]{F1F1F1} 0.187 & 0.061 & {\cellcolor[HTML]{F98E52}} \color[HTML]{F1F1F1} 0.486 \\
& \multicolumn{6}{l}{\textit{ICL (k=10)}} \\
 & Qwen3-30B-A3B & {\cellcolor[HTML]{6BBF64}} \color[HTML]{000000} 0.058 & 0.085 & {\cellcolor[HTML]{57B65F}} \color[HTML]{F1F1F1} 0.191 & 0.050 & {\cellcolor[HTML]{D5ED88}} \color[HTML]{000000} 0.711 \\
 & Olmo-3-7B-Instruct & {\cellcolor[HTML]{FEDC88}} \color[HTML]{000000} 0.102 & 0.109 & {\cellcolor[HTML]{17934E}} \color[HTML]{F1F1F1} 0.182 & 0.049 & {\cellcolor[HTML]{F98E52}} \color[HTML]{F1F1F1} 0.486 \\
 & GLM-4.7-Flash & {\cellcolor[HTML]{BDE379}} \color[HTML]{000000} 0.073 & 0.084 & {\cellcolor[HTML]{8ECF67}} \color[HTML]{000000} 0.199 & 0.048 & {\cellcolor[HTML]{E8F59F}} \color[HTML]{000000} 0.682 \\
 & Qwen3-30B-A3B-Thinking & {\cellcolor[HTML]{ABDB6D}} \color[HTML]{000000} 0.069 & 0.080 & {\cellcolor[HTML]{3FAA59}} \color[HTML]{F1F1F1} 0.188 & 0.047 & {\cellcolor[HTML]{FFFDBC}} \color[HTML]{000000} 0.639 \\
 & gpt-oss-120b & {\cellcolor[HTML]{279F53}} \color[HTML]{F1F1F1} 0.048 & 0.059 & {\cellcolor[HTML]{9BD469}} \color[HTML]{000000} 0.201 & 0.048 & {\cellcolor[HTML]{5DB961}} \color[HTML]{F1F1F1} 0.839 \\
 & Qwen3-8B & {\cellcolor[HTML]{C7E77F}} \color[HTML]{000000} 0.075 & 0.091 & {\cellcolor[HTML]{82C966}} \color[HTML]{000000} 0.197 & 0.054 & {\cellcolor[HTML]{FFFAB6}} \color[HTML]{000000} 0.632 \\
& \multicolumn{6}{l}{\textit{ICL (k=100)}} \\
 & Qwen3-30B-A3B & {\cellcolor[HTML]{7DC765}} \color[HTML]{000000} 0.061 & 0.064 & {\cellcolor[HTML]{A0D669}} \color[HTML]{000000} 0.202 & 0.056 & {\cellcolor[HTML]{98D368}} \color[HTML]{000000} 0.782 \\
 & Olmo-3-7B-Instruct & {\cellcolor[HTML]{FEE28F}} \color[HTML]{000000} 0.100 & 0.097 & {\cellcolor[HTML]{4EB15D}} \color[HTML]{F1F1F1} 0.190 & 0.052 & {\cellcolor[HTML]{FA9656}} \color[HTML]{000000} 0.494 \\
 & GLM-4.7-Flash & {\cellcolor[HTML]{B9E176}} \color[HTML]{000000} 0.072 & 0.093 & {\cellcolor[HTML]{A0D669}} \color[HTML]{000000} 0.202 & 0.048 & {\cellcolor[HTML]{FFF3AC}} \color[HTML]{000000} 0.620 \\
 & Qwen3-30B-A3B-Thinking & {\cellcolor[HTML]{7DC765}} \color[HTML]{000000} 0.061 & 0.073 & {\cellcolor[HTML]{87CB67}} \color[HTML]{000000} 0.198 & 0.064 & {\cellcolor[HTML]{CDEA83}} \color[HTML]{000000} 0.721 \\
 & gpt-oss-120b & {\cellcolor[HTML]{148E4B}} \color[HTML]{F1F1F1} 0.044 & 0.052 & {\cellcolor[HTML]{8ECF67}} \color[HTML]{000000} 0.199 & 0.050 & {\cellcolor[HTML]{279F53}} \color[HTML]{F1F1F1} 0.882 \\
 & Qwen3-8B & {\cellcolor[HTML]{ABDB6D}} \color[HTML]{000000} 0.069 & 0.084 & {\cellcolor[HTML]{D5ED88}} \color[HTML]{000000} 0.212 & 0.066 & {\cellcolor[HTML]{C5E67E}} \color[HTML]{000000} 0.729 \\
& \multicolumn{6}{l}{\textit{Baselines}} \\
 & Most Similar Embed. & {\cellcolor[HTML]{FBFDBA}} \color[HTML]{000000} 0.089 & 0.101 & {\cellcolor[HTML]{17934E}} \color[HTML]{F1F1F1} 0.182 & 0.035 & {\cellcolor[HTML]{FCAA5F}} \color[HTML]{000000} 0.514 \\
 & MICE PMM & {\cellcolor[HTML]{1E9A51}} \color[HTML]{F1F1F1} 0.047 & 0.052 & {\cellcolor[HTML]{A50026}} \color[HTML]{F1F1F1} 0.272 & 0.113 & {\cellcolor[HTML]{016A38}} \color[HTML]{F1F1F1} 0.954 \\
 & MICE Forest & {\cellcolor[HTML]{2DA155}} \color[HTML]{F1F1F1} 0.049 & 0.052 & {\cellcolor[HTML]{9BD469}} \color[HTML]{000000} 0.201 & 0.057 & {\cellcolor[HTML]{4EB15D}} \color[HTML]{F1F1F1} 0.850 \\
 & Mode Imputation & 0.175 & 0.159 & 0.207 & 0.039 & 0.275 \\
 & Random Sample & 0.121 & 0.126 & 0.124 & 0.032 & 0.261 \\
& \multicolumn{6}{l}{\textit{TabPFN}} \\
 & TabPFN (greedy) & {\cellcolor[HTML]{A50026}} \color[HTML]{F1F1F1} 0.145 & 0.156 & {\cellcolor[HTML]{73C264}} \color[HTML]{000000} 0.195 & 0.044 & {\cellcolor[HTML]{A50026}} \color[HTML]{F1F1F1} 0.329 \\
 & TabPFN (T=0.5) & {\cellcolor[HTML]{CBE982}} \color[HTML]{000000} 0.076 & 0.098 & {\cellcolor[HTML]{EEF8A8}} \color[HTML]{000000} 0.218 & 0.051 & {\cellcolor[HTML]{FDFEBC}} \color[HTML]{000000} 0.646 \\
 & TabPFN (T=1.0) & {\cellcolor[HTML]{0D8044}} \color[HTML]{F1F1F1} 0.041 & 0.050 & {\cellcolor[HTML]{EEF8A8}} \color[HTML]{000000} 0.218 & 0.061 & {\cellcolor[HTML]{219C52}} \color[HTML]{F1F1F1} 0.886 \\
\midrule
\multirow{28}{*}{\textbf{MNAR}} & \textit{Full Data} & 0.026 & 0.030 & 0.156 & 0.016 & 0.964 \\
 & \textit{Complete Case} & 0.062 & 0.071 & 0.254 & 0.073 & 0.886 \\
& \multicolumn{6}{l}{\textit{Zero-Shot LLM}} \\
 & Qwen3-30B-A3B & {\cellcolor[HTML]{93D168}} \color[HTML]{000000} 0.065 & 0.093 & {\cellcolor[HTML]{7AC665}} \color[HTML]{000000} 0.196 & 0.067 & {\cellcolor[HTML]{E8F59F}} \color[HTML]{000000} 0.682 \\
 & Olmo-3-7B-Instruct & {\cellcolor[HTML]{E3F399}} \color[HTML]{000000} 0.082 & 0.118 & {\cellcolor[HTML]{30A356}} \color[HTML]{F1F1F1} 0.186 & 0.059 & {\cellcolor[HTML]{FED884}} \color[HTML]{000000} 0.571 \\
 & GLM-4.7-Flash & {\cellcolor[HTML]{C1E57B}} \color[HTML]{000000} 0.074 & 0.104 & {\cellcolor[HTML]{7AC665}} \color[HTML]{000000} 0.196 & 0.059 & {\cellcolor[HTML]{FFF2AA}} \color[HTML]{000000} 0.618 \\
 & Qwen3-30B-A3B-Thinking & {\cellcolor[HTML]{C7E77F}} \color[HTML]{000000} 0.075 & 0.100 & {\cellcolor[HTML]{128A49}} \color[HTML]{F1F1F1} 0.180 & 0.053 & {\cellcolor[HTML]{FED884}} \color[HTML]{000000} 0.571 \\
 & gpt-oss-120b & {\cellcolor[HTML]{2DA155}} \color[HTML]{F1F1F1} 0.049 & 0.072 & {\cellcolor[HTML]{30A356}} \color[HTML]{F1F1F1} 0.186 & 0.060 & {\cellcolor[HTML]{9BD469}} \color[HTML]{000000} 0.778 \\
 & Qwen3-8B & {\cellcolor[HTML]{ABDB6D}} \color[HTML]{000000} 0.069 & 0.095 & {\cellcolor[HTML]{73C264}} \color[HTML]{000000} 0.195 & 0.066 & {\cellcolor[HTML]{FFF2AA}} \color[HTML]{000000} 0.618 \\
& \multicolumn{6}{l}{\textit{ICL (k=10)}} \\
 & Qwen3-30B-A3B & {\cellcolor[HTML]{42AC5A}} \color[HTML]{F1F1F1} 0.052 & 0.069 & {\cellcolor[HTML]{CBE982}} \color[HTML]{000000} 0.210 & 0.074 & {\cellcolor[HTML]{73C264}} \color[HTML]{000000} 0.818 \\
 & Olmo-3-7B-Instruct & {\cellcolor[HTML]{8ECF67}} \color[HTML]{000000} 0.064 & 0.103 & {\cellcolor[HTML]{82C966}} \color[HTML]{000000} 0.197 & 0.062 & {\cellcolor[HTML]{FFFEBE}} \color[HTML]{000000} 0.642 \\
 & GLM-4.7-Flash & {\cellcolor[HTML]{57B65F}} \color[HTML]{F1F1F1} 0.055 & 0.069 & {\cellcolor[HTML]{FFF3AC}} \color[HTML]{000000} 0.226 & 0.082 & {\cellcolor[HTML]{75C465}} \color[HTML]{000000} 0.817 \\
 & Qwen3-30B-A3B-Thinking & {\cellcolor[HTML]{3CA959}} \color[HTML]{F1F1F1} 0.051 & 0.078 & {\cellcolor[HTML]{D1EC86}} \color[HTML]{000000} 0.211 & 0.073 & {\cellcolor[HTML]{93D168}} \color[HTML]{000000} 0.786 \\
 & gpt-oss-120b & {\cellcolor[HTML]{2DA155}} \color[HTML]{F1F1F1} 0.049 & 0.058 & {\cellcolor[HTML]{FFFBB8}} \color[HTML]{000000} 0.224 & 0.084 & {\cellcolor[HTML]{1E9A51}} \color[HTML]{F1F1F1} 0.889 \\
 & Qwen3-8B & {\cellcolor[HTML]{5DB961}} \color[HTML]{F1F1F1} 0.056 & 0.078 & {\cellcolor[HTML]{E2F397}} \color[HTML]{000000} 0.215 & 0.080 & {\cellcolor[HTML]{89CC67}} \color[HTML]{000000} 0.796 \\
& \multicolumn{6}{l}{\textit{ICL (k=100)}} \\
 & Qwen3-30B-A3B & {\cellcolor[HTML]{33A456}} \color[HTML]{F1F1F1} 0.050 & 0.065 & {\cellcolor[HTML]{EEF8A8}} \color[HTML]{000000} 0.218 & 0.089 & {\cellcolor[HTML]{54B45F}} \color[HTML]{F1F1F1} 0.846 \\
 & Olmo-3-7B-Instruct & {\cellcolor[HTML]{9BD469}} \color[HTML]{000000} 0.066 & 0.103 & {\cellcolor[HTML]{82C966}} \color[HTML]{000000} 0.197 & 0.065 & {\cellcolor[HTML]{FFFCBA}} \color[HTML]{000000} 0.636 \\
 & GLM-4.7-Flash & {\cellcolor[HTML]{78C565}} \color[HTML]{000000} 0.060 & 0.083 & {\cellcolor[HTML]{FAFDB8}} \color[HTML]{000000} 0.221 & 0.085 & {\cellcolor[HTML]{98D368}} \color[HTML]{000000} 0.781 \\
 & Qwen3-30B-A3B-Thinking & {\cellcolor[HTML]{5DB961}} \color[HTML]{F1F1F1} 0.056 & 0.071 & {\cellcolor[HTML]{DFF293}} \color[HTML]{000000} 0.214 & 0.079 & {\cellcolor[HTML]{84CA66}} \color[HTML]{000000} 0.800 \\
 & gpt-oss-120b & {\cellcolor[HTML]{279F53}} \color[HTML]{F1F1F1} 0.048 & 0.061 & {\cellcolor[HTML]{FFFDBC}} \color[HTML]{000000} 0.223 & 0.097 & {\cellcolor[HTML]{30A356}} \color[HTML]{F1F1F1} 0.875 \\
 & Qwen3-8B & {\cellcolor[HTML]{48AE5C}} \color[HTML]{F1F1F1} 0.053 & 0.069 & {\cellcolor[HTML]{FEEDA1}} \color[HTML]{000000} 0.228 & 0.097 & {\cellcolor[HTML]{39A758}} \color[HTML]{F1F1F1} 0.868 \\
& \multicolumn{6}{l}{\textit{Baselines}} \\
 & Most Similar Embed. & {\cellcolor[HTML]{57B65F}} \color[HTML]{F1F1F1} 0.055 & 0.074 & {\cellcolor[HTML]{82C966}} \color[HTML]{000000} 0.197 & 0.078 & {\cellcolor[HTML]{89CC67}} \color[HTML]{000000} 0.796 \\
 & MICE PMM & {\cellcolor[HTML]{42AC5A}} \color[HTML]{F1F1F1} 0.052 & 0.068 & {\cellcolor[HTML]{C21C27}} \color[HTML]{F1F1F1} 0.266 & 0.124 & {\cellcolor[HTML]{249D53}} \color[HTML]{F1F1F1} 0.884 \\
 & MICE Forest & {\cellcolor[HTML]{B3DF72}} \color[HTML]{000000} 0.071 & 0.083 & {\cellcolor[HTML]{E9F6A1}} \color[HTML]{000000} 0.217 & 0.087 & {\cellcolor[HTML]{C3E67D}} \color[HTML]{000000} 0.732 \\
 & Mode Imputation & 0.058 & 0.080 & 0.231 & 0.069 & 0.800 \\
 & Random Sample & 0.075 & 0.122 & 0.135 & 0.029 & 0.425 \\
& \multicolumn{6}{l}{\textit{TabPFN}} \\
 & TabPFN (greedy) & {\cellcolor[HTML]{DCF08F}} \color[HTML]{000000} 0.080 & 0.117 & {\cellcolor[HTML]{F1F9AC}} \color[HTML]{000000} 0.219 & 0.093 & {\cellcolor[HTML]{FFF1A8}} \color[HTML]{000000} 0.614 \\
 & TabPFN (T=0.5) & {\cellcolor[HTML]{AFDD70}} \color[HTML]{000000} 0.070 & 0.082 & {\cellcolor[HTML]{FDB567}} \color[HTML]{000000} 0.241 & 0.104 & {\cellcolor[HTML]{C3E67D}} \color[HTML]{000000} 0.732 \\
 & TabPFN (T=1.0) & {\cellcolor[HTML]{1E9A51}} \color[HTML]{F1F1F1} 0.047 & 0.067 & {\cellcolor[HTML]{FEE18D}} \color[HTML]{000000} 0.232 & 0.094 & {\cellcolor[HTML]{42AC5A}} \color[HTML]{F1F1F1} 0.861 \\
\bottomrule
\end{tabular}

\end{minipage}\hfill
\begin{minipage}[t]{0.42\textwidth}\vspace{0pt}
  Table~\ref{tab:study_2_table_appendix} extends Table~\ref{tab:study_2_table} to all six LLMs and further baselines. Tables~\ref{tab:sig-abs-error-mcar}--\ref{tab:sig-coverage-mnar} report the pairwise tests of Section~\ref{sec:comparison} for the methods of Table~\ref{tab:study_2_table}: Wilcoxon signed-rank tests for absolute error and McNemar tests for coverage, each after an omnibus test and with Holm correction.

  \medskip
  \captionof{table}{\textbf{Full results of the main comparison}, extending Table~\ref{tab:study_2_table} with four more generators, TabPFN at two further temperatures (greedy and $T=0.5$), and two simple baselines: Mode Imputation imputes the most frequent observed category, and Random Sample draws uniformly from the observed answer categories. Cell colors as in Table~\ref{tab:study_2_table}. Significance tests (Tables~\ref{tab:sig-abs-error-mcar}--\ref{tab:sig-coverage-mnar}) cover the methods of Table~\ref{tab:study_2_table}.}\label{tab:study_2_table_appendix}
\end{minipage}
\vspace{\baselineskip}
}]

\begin{table*}[ht]
  \centering
  \footnotesize
  \begin{tabular}{llrrrc}
    \toprule
    \textbf{Method 1} & \textbf{Method 2} & \textbf{Med. Diff.} & \textbf{$p_{\text{raw}}$} & \textbf{$p_{\text{Holm}}$} & \textbf{Sig.} \\
    \midrule
    Zero-Shot, Qwen3-30B-A3B & ICL (k=100), Qwen3-30B-A3B & 0.0228 & ${<}\,0.001$ & ${<}\,0.001$ & $\checkmark$ \\
    Zero-Shot, Qwen3-30B-A3B & ICL (k=10), gpt-oss-120b & 0.0279 & ${<}\,0.001$ & ${<}\,0.001$ & $\checkmark$ \\
    Zero-Shot, Qwen3-30B-A3B & ICL (k=10), Qwen3-30B-A3B & 0.0263 & ${<}\,0.001$ & ${<}\,0.001$ & $\checkmark$ \\
    Zero-Shot, Qwen3-30B-A3B & ICL (k=100), gpt-oss-120b & 0.0288 & ${<}\,0.001$ & ${<}\,0.001$ & $\checkmark$ \\
    Zero-Shot, Qwen3-30B-A3B & MICE PMM & 0.0290 & ${<}\,0.001$ & ${<}\,0.001$ & $\checkmark$ \\
    Zero-Shot, Qwen3-30B-A3B & TabPFN & 0.0191 & ${<}\,0.001$ & ${<}\,0.001$ & $\checkmark$ \\
    Complete Case & Zero-Shot, Qwen3-30B-A3B & -0.0218 & ${<}\,0.001$ & ${<}\,0.001$ & $\checkmark$ \\
    MICE PMM & MICE Forest & -0.0176 & ${<}\,0.001$ & ${<}\,0.001$ & $\checkmark$ \\
    Zero-Shot, Qwen3-30B-A3B & Most Similar Embed. & 0.0165 & ${<}\,0.001$ & ${<}\,0.001$ & $\checkmark$ \\
    Zero-Shot, Qwen3-30B-A3B & Zero-Shot, gpt-oss-120b & 0.0169 & ${<}\,0.001$ & $0.003$ & $\checkmark$ \\
    Complete Case & MICE Forest & -0.0151 & ${<}\,0.001$ & $0.003$ & $\checkmark$ \\
    MICE Forest & TabPFN & 0.0186 & ${<}\,0.001$ & $0.004$ & $\checkmark$ \\
    Zero-Shot, gpt-oss-120b & MICE PMM & 0.0085 & $0.002$ & $0.065$ & \\
    ICL (k=100), gpt-oss-120b & MICE Forest & -0.0111 & $0.002$ & $0.100$ & \\
    ICL (k=10), gpt-oss-120b & MICE Forest & -0.0111 & $0.004$ & $0.173$ & \\
    Zero-Shot, Qwen3-30B-A3B & MICE Forest & 0.0168 & $0.005$ & $0.183$ & \\
    Most Similar Embed. & MICE Forest & -0.0087 & $0.007$ & $0.283$ & \\
    Complete Case & MICE PMM & 0.0064 & $0.010$ & $0.388$ & \\
    Zero-Shot, gpt-oss-120b & ICL (k=100), gpt-oss-120b & 0.0067 & $0.013$ & $0.493$ & \\
    Zero-Shot, gpt-oss-120b & TabPFN & 0.0057 & $0.016$ & $0.589$ & \\
    ICL (k=10), Qwen3-30B-A3B & MICE PMM & 0.0050 & $0.019$ & $0.677$ & \\
    Complete Case & Zero-Shot, gpt-oss-120b & -0.0096 & $0.024$ & $0.823$ & \\
    ICL (k=10), gpt-oss-120b & TabPFN & 0.0054 & $0.409$ & $1.000$ & \\
    ICL (k=100), Qwen3-30B-A3B & ICL (k=100), gpt-oss-120b & 0.0016 & $0.119$ & $1.000$ & \\
    ICL (k=100), Qwen3-30B-A3B & Most Similar Embed. & -0.0072 & $0.950$ & $1.000$ & \\
    ICL (k=100), Qwen3-30B-A3B & MICE PMM & 0.0022 & $0.102$ & $1.000$ & \\
    ICL (k=100), Qwen3-30B-A3B & TabPFN & 0.0036 & $0.140$ & $1.000$ & \\
    ICL (k=10), gpt-oss-120b & MICE PMM & 0.0027 & $0.206$ & $1.000$ & \\
    ICL (k=100), gpt-oss-120b & Most Similar Embed. & -0.0055 & $0.206$ & $1.000$ & \\
    ICL (k=100), gpt-oss-120b & MICE PMM & 0.0041 & $0.186$ & $1.000$ & \\
    ICL (k=100), gpt-oss-120b & TabPFN & 0.0046 & $0.461$ & $1.000$ & \\
    Most Similar Embed. & MICE PMM & 0.0080 & $0.033$ & $1.000$ & \\
    Most Similar Embed. & TabPFN & 0.0026 & $0.237$ & $1.000$ & \\
    ICL (k=100), Qwen3-30B-A3B & MICE Forest & -0.0101 & $0.113$ & $1.000$ & \\
    ICL (k=10), gpt-oss-120b & Most Similar Embed. & -0.0014 & $0.381$ & $1.000$ & \\
    ICL (k=10), Qwen3-30B-A3B & TabPFN & 0.0059 & $0.055$ & $1.000$ & \\
    ICL (k=10), gpt-oss-120b & ICL (k=100), Qwen3-30B-A3B & 0.0009 & $0.449$ & $1.000$ & \\
    Complete Case & Most Similar Embed. & -0.0018 & $0.321$ & $1.000$ & \\
    Complete Case & ICL (k=100), gpt-oss-120b & 0.0005 & $0.614$ & $1.000$ & \\
    Complete Case & ICL (k=100), Qwen3-30B-A3B & -0.0003 & $0.413$ & $1.000$ & \\
    Complete Case & ICL (k=10), gpt-oss-120b & 0.0005 & $0.979$ & $1.000$ & \\
    Complete Case & ICL (k=10), Qwen3-30B-A3B & -0.0044 & $0.214$ & $1.000$ & \\
    Zero-Shot, gpt-oss-120b & ICL (k=10), Qwen3-30B-A3B & 0.0024 & $0.633$ & $1.000$ & \\
    Zero-Shot, gpt-oss-120b & ICL (k=10), gpt-oss-120b & 0.0015 & $0.142$ & $1.000$ & \\
    ICL (k=10), gpt-oss-120b & ICL (k=100), gpt-oss-120b & 0.0046 & $0.175$ & $1.000$ & \\
    Zero-Shot, gpt-oss-120b & ICL (k=100), Qwen3-30B-A3B & 0.0047 & $0.166$ & $1.000$ & \\
    Zero-Shot, gpt-oss-120b & MICE Forest & -0.0043 & $0.715$ & $1.000$ & \\
    MICE PMM & TabPFN & -0.0034 & $0.360$ & $1.000$ & \\
    ICL (k=10), Qwen3-30B-A3B & ICL (k=100), Qwen3-30B-A3B & 0.0038 & $0.121$ & $1.000$ & \\
    ICL (k=10), Qwen3-30B-A3B & ICL (k=100), gpt-oss-120b & 0.0028 & $0.040$ & $1.000$ & \\
    ICL (k=10), Qwen3-30B-A3B & Most Similar Embed. & 0.0009 & $0.731$ & $1.000$ & \\
    ICL (k=10), Qwen3-30B-A3B & MICE Forest & -0.0076 & $0.212$ & $1.000$ & \\
    Complete Case & TabPFN & 0.0025 & $0.416$ & $1.000$ & \\
    Zero-Shot, gpt-oss-120b & Most Similar Embed. & 0.0013 & $0.161$ & $1.000$ & \\
    ICL (k=10), Qwen3-30B-A3B & ICL (k=10), gpt-oss-120b & 0.0006 & $0.220$ & $1.000$ & \\
    \bottomrule
\end{tabular}

  \caption{\textbf{Pairwise Wilcoxon signed-rank tests for median absolute error under MCAR missingness} (Holm-corrected, $\alpha=0.05$). Omnibus Friedman test: $\chi^2=59.26$, $p={<}\,0.001$, $N=138$.}
  \label{tab:sig-abs-error-mcar}
\end{table*}

\begin{table*}[ht]
  \centering
  \footnotesize
  \begin{tabular}{llrrrc}
    \toprule
    \textbf{Method 1} & \textbf{Method 2} & \textbf{Med. Diff.} & \textbf{$p_{\text{raw}}$} & \textbf{$p_{\text{Holm}}$} & \textbf{Sig.} \\
    \midrule
    Most Similar Embed. & TabPFN & 0.0473 & ${<}\,0.001$ & ${<}\,0.001$ & $\checkmark$ \\
    Complete Case & Most Similar Embed. & -0.0446 & ${<}\,0.001$ & ${<}\,0.001$ & $\checkmark$ \\
    ICL (k=100), gpt-oss-120b & Most Similar Embed. & -0.0436 & ${<}\,0.001$ & ${<}\,0.001$ & $\checkmark$ \\
    Most Similar Embed. & MICE PMM & 0.0433 & ${<}\,0.001$ & ${<}\,0.001$ & $\checkmark$ \\
    Zero-Shot, Qwen3-30B-A3B & ICL (k=100), gpt-oss-120b & 0.0369 & ${<}\,0.001$ & ${<}\,0.001$ & $\checkmark$ \\
    ICL (k=10), gpt-oss-120b & Most Similar Embed. & -0.0356 & ${<}\,0.001$ & ${<}\,0.001$ & $\checkmark$ \\
    Most Similar Embed. & MICE Forest & 0.0348 & ${<}\,0.001$ & ${<}\,0.001$ & $\checkmark$ \\
    Zero-Shot, Qwen3-30B-A3B & TabPFN & 0.0339 & ${<}\,0.001$ & ${<}\,0.001$ & $\checkmark$ \\
    Complete Case & Zero-Shot, Qwen3-30B-A3B & -0.0305 & ${<}\,0.001$ & ${<}\,0.001$ & $\checkmark$ \\
    Zero-Shot, Qwen3-30B-A3B & MICE PMM & 0.0314 & ${<}\,0.001$ & ${<}\,0.001$ & $\checkmark$ \\
    Zero-Shot, Qwen3-30B-A3B & ICL (k=10), gpt-oss-120b & 0.0275 & ${<}\,0.001$ & ${<}\,0.001$ & $\checkmark$ \\
    Zero-Shot, gpt-oss-120b & ICL (k=100), gpt-oss-120b & 0.0224 & ${<}\,0.001$ & ${<}\,0.001$ & $\checkmark$ \\
    Zero-Shot, Qwen3-30B-A3B & ICL (k=10), Qwen3-30B-A3B & 0.0222 & ${<}\,0.001$ & ${<}\,0.001$ & $\checkmark$ \\
    Zero-Shot, Qwen3-30B-A3B & ICL (k=100), Qwen3-30B-A3B & 0.0194 & ${<}\,0.001$ & ${<}\,0.001$ & $\checkmark$ \\
    Zero-Shot, Qwen3-30B-A3B & MICE Forest & 0.0239 & ${<}\,0.001$ & ${<}\,0.001$ & $\checkmark$ \\
    ICL (k=10), Qwen3-30B-A3B & ICL (k=100), gpt-oss-120b & 0.0145 & ${<}\,0.001$ & ${<}\,0.001$ & $\checkmark$ \\
    Complete Case & Zero-Shot, gpt-oss-120b & -0.0168 & ${<}\,0.001$ & ${<}\,0.001$ & $\checkmark$ \\
    Zero-Shot, gpt-oss-120b & TabPFN & 0.0251 & ${<}\,0.001$ & ${<}\,0.001$ & $\checkmark$ \\
    ICL (k=100), Qwen3-30B-A3B & Most Similar Embed. & -0.0260 & ${<}\,0.001$ & ${<}\,0.001$ & $\checkmark$ \\
    Zero-Shot, gpt-oss-120b & MICE PMM & 0.0165 & ${<}\,0.001$ & ${<}\,0.001$ & $\checkmark$ \\
    ICL (k=100), Qwen3-30B-A3B & ICL (k=100), gpt-oss-120b & 0.0089 & ${<}\,0.001$ & ${<}\,0.001$ & $\checkmark$ \\
    Zero-Shot, gpt-oss-120b & ICL (k=10), gpt-oss-120b & 0.0153 & ${<}\,0.001$ & ${<}\,0.001$ & $\checkmark$ \\
    ICL (k=10), Qwen3-30B-A3B & Most Similar Embed. & -0.0247 & ${<}\,0.001$ & ${<}\,0.001$ & $\checkmark$ \\
    Zero-Shot, gpt-oss-120b & MICE Forest & 0.0159 & ${<}\,0.001$ & ${<}\,0.001$ & $\checkmark$ \\
    Complete Case & ICL (k=10), Qwen3-30B-A3B & -0.0103 & ${<}\,0.001$ & ${<}\,0.001$ & $\checkmark$ \\
    ICL (k=100), Qwen3-30B-A3B & TabPFN & 0.0097 & ${<}\,0.001$ & $0.001$ & $\checkmark$ \\
    Complete Case & ICL (k=100), Qwen3-30B-A3B & -0.0074 & ${<}\,0.001$ & $0.002$ & $\checkmark$ \\
    ICL (k=10), Qwen3-30B-A3B & TabPFN & 0.0139 & ${<}\,0.001$ & $0.002$ & $\checkmark$ \\
    Zero-Shot, gpt-oss-120b & Most Similar Embed. & -0.0219 & ${<}\,0.001$ & $0.002$ & $\checkmark$ \\
    ICL (k=10), gpt-oss-120b & ICL (k=100), gpt-oss-120b & 0.0087 & ${<}\,0.001$ & $0.003$ & $\checkmark$ \\
    ICL (k=10), Qwen3-30B-A3B & MICE PMM & 0.0105 & ${<}\,0.001$ & $0.005$ & $\checkmark$ \\
    ICL (k=100), Qwen3-30B-A3B & MICE PMM & 0.0089 & ${<}\,0.001$ & $0.011$ & $\checkmark$ \\
    ICL (k=10), Qwen3-30B-A3B & ICL (k=10), gpt-oss-120b & 0.0080 & ${<}\,0.001$ & $0.020$ & $\checkmark$ \\
    Zero-Shot, Qwen3-30B-A3B & Zero-Shot, gpt-oss-120b & 0.0124 & ${<}\,0.001$ & $0.020$ & $\checkmark$ \\
    ICL (k=10), Qwen3-30B-A3B & MICE Forest & 0.0100 & $0.005$ & $0.111$ & \\
    ICL (k=10), gpt-oss-120b & ICL (k=100), Qwen3-30B-A3B & -0.0051 & $0.008$ & $0.155$ & \\
    Zero-Shot, gpt-oss-120b & ICL (k=100), Qwen3-30B-A3B & 0.0084 & $0.014$ & $0.263$ & \\
    MICE Forest & TabPFN & 0.0099 & $0.019$ & $0.331$ & \\
    Complete Case & ICL (k=10), gpt-oss-120b & -0.0032 & $0.018$ & $0.331$ & \\
    Zero-Shot, gpt-oss-120b & ICL (k=10), Qwen3-30B-A3B & 0.0081 & $0.020$ & $0.331$ & \\
    ICL (k=10), gpt-oss-120b & TabPFN & 0.0039 & $0.026$ & $0.383$ & \\
    ICL (k=100), Qwen3-30B-A3B & MICE Forest & 0.0071 & $0.029$ & $0.401$ & \\
    ICL (k=100), gpt-oss-120b & MICE Forest & -0.0054 & $0.044$ & $0.570$ & \\
    ICL (k=10), gpt-oss-120b & MICE Forest & -0.0008 & $0.886$ & $1.000$ & \\
    ICL (k=10), gpt-oss-120b & MICE PMM & 0.0048 & $0.124$ & $1.000$ & \\
    Complete Case & TabPFN & 0.0009 & $0.809$ & $1.000$ & \\
    Complete Case & MICE Forest & -0.0033 & $0.093$ & $1.000$ & \\
    ICL (k=100), gpt-oss-120b & MICE PMM & -0.0030 & $0.315$ & $1.000$ & \\
    ICL (k=100), gpt-oss-120b & TabPFN & -0.0019 & $0.697$ & $1.000$ & \\
    Complete Case & MICE PMM & -0.0014 & $0.498$ & $1.000$ & \\
    Complete Case & ICL (k=100), gpt-oss-120b & 0.0002 & $0.725$ & $1.000$ & \\
    MICE PMM & TabPFN & 0.0027 & $0.592$ & $1.000$ & \\
    MICE PMM & MICE Forest & -0.0023 & $0.101$ & $1.000$ & \\
    Zero-Shot, Qwen3-30B-A3B & Most Similar Embed. & -0.0087 & $0.304$ & $1.000$ & \\
    ICL (k=10), Qwen3-30B-A3B & ICL (k=100), Qwen3-30B-A3B & 0.0020 & $0.188$ & $1.000$ & \\
    \bottomrule
\end{tabular}

  \caption{\textbf{Pairwise Wilcoxon signed-rank tests for median absolute error under MAR missingness} (Holm-corrected, $\alpha=0.05$). Omnibus Friedman test: $\chi^2=218.56$, $p={<}\,0.001$, $N=277$.}
  \label{tab:sig-abs-error-mar}
\end{table*}

\begin{table*}[ht]
  \centering
  \footnotesize
  \begin{tabular}{llrrrc}
    \toprule
    \textbf{Method 1} & \textbf{Method 2} & \textbf{Med. Diff.} & \textbf{$p_{\text{raw}}$} & \textbf{$p_{\text{Holm}}$} & \textbf{Sig.} \\
    \midrule
    ICL (k=100), gpt-oss-120b & MICE Forest & -0.0284 & ${<}\,0.001$ & ${<}\,0.001$ & $\checkmark$ \\
    MICE Forest & TabPFN & 0.0255 & ${<}\,0.001$ & ${<}\,0.001$ & $\checkmark$ \\
    ICL (k=10), gpt-oss-120b & MICE Forest & -0.0215 & ${<}\,0.001$ & ${<}\,0.001$ & $\checkmark$ \\
    Zero-Shot, Qwen3-30B-A3B & ICL (k=10), Qwen3-30B-A3B & 0.0154 & ${<}\,0.001$ & ${<}\,0.001$ & $\checkmark$ \\
    MICE PMM & MICE Forest & -0.0208 & ${<}\,0.001$ & ${<}\,0.001$ & $\checkmark$ \\
    Zero-Shot, Qwen3-30B-A3B & ICL (k=10), gpt-oss-120b & 0.0142 & ${<}\,0.001$ & ${<}\,0.001$ & $\checkmark$ \\
    Zero-Shot, Qwen3-30B-A3B & ICL (k=100), gpt-oss-120b & 0.0181 & ${<}\,0.001$ & ${<}\,0.001$ & $\checkmark$ \\
    Zero-Shot, Qwen3-30B-A3B & ICL (k=100), Qwen3-30B-A3B & 0.0144 & ${<}\,0.001$ & ${<}\,0.001$ & $\checkmark$ \\
    Complete Case & ICL (k=100), gpt-oss-120b & 0.0161 & ${<}\,0.001$ & ${<}\,0.001$ & $\checkmark$ \\
    Zero-Shot, Qwen3-30B-A3B & Zero-Shot, gpt-oss-120b & 0.0144 & ${<}\,0.001$ & ${<}\,0.001$ & $\checkmark$ \\
    Complete Case & MICE Forest & -0.0133 & ${<}\,0.001$ & ${<}\,0.001$ & $\checkmark$ \\
    ICL (k=100), Qwen3-30B-A3B & MICE Forest & -0.0209 & ${<}\,0.001$ & ${<}\,0.001$ & $\checkmark$ \\
    ICL (k=10), Qwen3-30B-A3B & MICE Forest & -0.0233 & ${<}\,0.001$ & ${<}\,0.001$ & $\checkmark$ \\
    Zero-Shot, gpt-oss-120b & MICE Forest & -0.0225 & ${<}\,0.001$ & ${<}\,0.001$ & $\checkmark$ \\
    Complete Case & ICL (k=10), gpt-oss-120b & 0.0145 & ${<}\,0.001$ & ${<}\,0.001$ & $\checkmark$ \\
    Most Similar Embed. & MICE Forest & -0.0185 & ${<}\,0.001$ & ${<}\,0.001$ & $\checkmark$ \\
    Complete Case & TabPFN & 0.0115 & ${<}\,0.001$ & ${<}\,0.001$ & $\checkmark$ \\
    ICL (k=100), gpt-oss-120b & Most Similar Embed. & -0.0110 & ${<}\,0.001$ & ${<}\,0.001$ & $\checkmark$ \\
    Zero-Shot, Qwen3-30B-A3B & TabPFN & 0.0133 & ${<}\,0.001$ & ${<}\,0.001$ & $\checkmark$ \\
    ICL (k=10), gpt-oss-120b & Most Similar Embed. & -0.0111 & ${<}\,0.001$ & $0.002$ & $\checkmark$ \\
    Complete Case & MICE PMM & 0.0065 & ${<}\,0.001$ & $0.002$ & $\checkmark$ \\
    Zero-Shot, Qwen3-30B-A3B & MICE PMM & 0.0096 & $0.001$ & $0.034$ & $\checkmark$ \\
    Complete Case & ICL (k=100), Qwen3-30B-A3B & 0.0117 & $0.001$ & $0.034$ & $\checkmark$ \\
    Complete Case & ICL (k=10), Qwen3-30B-A3B & 0.0101 & $0.002$ & $0.057$ & \\
    ICL (k=100), gpt-oss-120b & MICE PMM & -0.0073 & $0.002$ & $0.072$ & \\
    ICL (k=10), Qwen3-30B-A3B & ICL (k=10), gpt-oss-120b & 0.0049 & $0.006$ & $0.166$ & \\
    ICL (k=10), Qwen3-30B-A3B & ICL (k=100), gpt-oss-120b & 0.0038 & $0.007$ & $0.198$ & \\
    Complete Case & Zero-Shot, gpt-oss-120b & 0.0112 & $0.007$ & $0.200$ & \\
    ICL (k=10), gpt-oss-120b & MICE PMM & -0.0075 & $0.008$ & $0.211$ & \\
    ICL (k=100), Qwen3-30B-A3B & ICL (k=100), gpt-oss-120b & 0.0054 & $0.009$ & $0.236$ & \\
    Zero-Shot, Qwen3-30B-A3B & Most Similar Embed. & 0.0081 & $0.014$ & $0.353$ & \\
    Most Similar Embed. & TabPFN & 0.0054 & $0.020$ & $0.487$ & \\
    ICL (k=100), gpt-oss-120b & TabPFN & -0.0054 & $0.025$ & $0.566$ & \\
    ICL (k=10), gpt-oss-120b & ICL (k=100), Qwen3-30B-A3B & -0.0045 & $0.027$ & $0.590$ & \\
    Zero-Shot, gpt-oss-120b & Most Similar Embed. & -0.0053 & $0.044$ & $0.930$ & \\
    Complete Case & Most Similar Embed. & 0.0038 & $0.077$ & $1.000$ & \\
    Most Similar Embed. & MICE PMM & 0.0023 & $0.166$ & $1.000$ & \\
    Zero-Shot, Qwen3-30B-A3B & MICE Forest & -0.0078 & $0.197$ & $1.000$ & \\
    Zero-Shot, gpt-oss-120b & ICL (k=10), Qwen3-30B-A3B & -0.0025 & $0.362$ & $1.000$ & \\
    ICL (k=100), Qwen3-30B-A3B & TabPFN & -0.0014 & $0.866$ & $1.000$ & \\
    Zero-Shot, gpt-oss-120b & ICL (k=10), gpt-oss-120b & 0.0005 & $0.253$ & $1.000$ & \\
    ICL (k=100), Qwen3-30B-A3B & MICE PMM & -0.0033 & $0.414$ & $1.000$ & \\
    Zero-Shot, gpt-oss-120b & TabPFN & -0.0021 & $0.869$ & $1.000$ & \\
    Zero-Shot, gpt-oss-120b & MICE PMM & -0.0018 & $0.414$ & $1.000$ & \\
    ICL (k=10), gpt-oss-120b & TabPFN & -0.0036 & $0.164$ & $1.000$ & \\
    Zero-Shot, gpt-oss-120b & ICL (k=100), Qwen3-30B-A3B & -0.0039 & $0.539$ & $1.000$ & \\
    Zero-Shot, gpt-oss-120b & ICL (k=100), gpt-oss-120b & 0.0017 & $0.106$ & $1.000$ & \\
    ICL (k=10), gpt-oss-120b & ICL (k=100), gpt-oss-120b & 0.0012 & $0.284$ & $1.000$ & \\
    ICL (k=10), Qwen3-30B-A3B & TabPFN & 0.0009 & $0.669$ & $1.000$ & \\
    ICL (k=10), Qwen3-30B-A3B & MICE PMM & -0.0047 & $0.408$ & $1.000$ & \\
    ICL (k=10), Qwen3-30B-A3B & Most Similar Embed. & -0.0088 & $0.059$ & $1.000$ & \\
    ICL (k=10), Qwen3-30B-A3B & ICL (k=100), Qwen3-30B-A3B & 0.0008 & $0.460$ & $1.000$ & \\
    MICE PMM & TabPFN & 0.0004 & $0.406$ & $1.000$ & \\
    ICL (k=100), Qwen3-30B-A3B & Most Similar Embed. & -0.0073 & $0.051$ & $1.000$ & \\
    Complete Case & Zero-Shot, Qwen3-30B-A3B & -0.0026 & $0.118$ & $1.000$ & \\
    \bottomrule
\end{tabular}

  \caption{\textbf{Pairwise Wilcoxon signed-rank tests for median absolute error under MNAR missingness} (Holm-corrected, $\alpha=0.05$). Omnibus Friedman test: $\chi^2=125.37$, $p={<}\,0.001$, $N=275$.}
  \label{tab:sig-abs-error-mnar}
\end{table*}

\begin{table*}[ht]
  \centering
  \footnotesize
  \begin{tabular}{llrrrrc}
    \toprule
    \textbf{Method 1} & \textbf{Method 2} & \textbf{Cov.\textsubscript{1}} & \textbf{Cov.\textsubscript{2}} & \textbf{$p_{\text{raw}}$} & \textbf{$p_{\text{Holm}}$} & \textbf{Sig.} \\
    \midrule
    Zero-Shot, Qwen3-30B-A3B & MICE PMM & 0.638 & 0.957 & ${<}\,0.001$ & ${<}\,0.001$ & $\checkmark$ \\
    Zero-Shot, Qwen3-30B-A3B & TabPFN & 0.638 & 0.949 & ${<}\,0.001$ & ${<}\,0.001$ & $\checkmark$ \\
    Complete Case & Zero-Shot, Qwen3-30B-A3B & 0.957 & 0.638 & ${<}\,0.001$ & ${<}\,0.001$ & $\checkmark$ \\
    Zero-Shot, Qwen3-30B-A3B & ICL (k=100), gpt-oss-120b & 0.638 & 0.906 & ${<}\,0.001$ & ${<}\,0.001$ & $\checkmark$ \\
    Zero-Shot, Qwen3-30B-A3B & ICL (k=10), gpt-oss-120b & 0.638 & 0.906 & ${<}\,0.001$ & ${<}\,0.001$ & $\checkmark$ \\
    Zero-Shot, Qwen3-30B-A3B & ICL (k=10), Qwen3-30B-A3B & 0.638 & 0.848 & ${<}\,0.001$ & ${<}\,0.001$ & $\checkmark$ \\
    Zero-Shot, Qwen3-30B-A3B & ICL (k=100), Qwen3-30B-A3B & 0.638 & 0.841 & ${<}\,0.001$ & ${<}\,0.001$ & $\checkmark$ \\
    Complete Case & MICE Forest & 0.957 & 0.804 & ${<}\,0.001$ & ${<}\,0.001$ & $\checkmark$ \\
    Complete Case & Zero-Shot, gpt-oss-120b & 0.957 & 0.768 & ${<}\,0.001$ & ${<}\,0.001$ & $\checkmark$ \\
    Zero-Shot, gpt-oss-120b & MICE PMM & 0.768 & 0.957 & ${<}\,0.001$ & ${<}\,0.001$ & $\checkmark$ \\
    Zero-Shot, Qwen3-30B-A3B & Most Similar Embed. & 0.638 & 0.870 & ${<}\,0.001$ & ${<}\,0.001$ & $\checkmark$ \\
    Zero-Shot, gpt-oss-120b & TabPFN & 0.768 & 0.949 & ${<}\,0.001$ & $0.002$ & $\checkmark$ \\
    MICE PMM & MICE Forest & 0.957 & 0.804 & ${<}\,0.001$ & $0.002$ & $\checkmark$ \\
    MICE Forest & TabPFN & 0.804 & 0.949 & ${<}\,0.001$ & $0.004$ & $\checkmark$ \\
    ICL (k=100), Qwen3-30B-A3B & MICE PMM & 0.841 & 0.957 & $0.002$ & $0.063$ & \\
    Zero-Shot, Qwen3-30B-A3B & MICE Forest & 0.638 & 0.804 & $0.002$ & $0.071$ & \\
    Complete Case & Most Similar Embed. & 0.957 & 0.870 & $0.002$ & $0.071$ & \\
    Zero-Shot, gpt-oss-120b & ICL (k=100), gpt-oss-120b & 0.768 & 0.906 & $0.002$ & $0.071$ & \\
    Complete Case & ICL (k=100), Qwen3-30B-A3B & 0.957 & 0.841 & $0.002$ & $0.092$ & \\
    Zero-Shot, gpt-oss-120b & ICL (k=10), gpt-oss-120b & 0.768 & 0.906 & $0.003$ & $0.092$ & \\
    Zero-Shot, Qwen3-30B-A3B & Zero-Shot, gpt-oss-120b & 0.638 & 0.768 & $0.004$ & $0.138$ & \\
    Complete Case & ICL (k=10), Qwen3-30B-A3B & 0.957 & 0.848 & $0.004$ & $0.139$ & \\
    ICL (k=10), Qwen3-30B-A3B & MICE PMM & 0.848 & 0.957 & $0.004$ & $0.139$ & \\
    ICL (k=100), Qwen3-30B-A3B & TabPFN & 0.841 & 0.949 & $0.004$ & $0.139$ & \\
    Most Similar Embed. & MICE PMM & 0.870 & 0.957 & $0.008$ & $0.234$ & \\
    Most Similar Embed. & TabPFN & 0.870 & 0.949 & $0.013$ & $0.376$ & \\
    ICL (k=10), Qwen3-30B-A3B & TabPFN & 0.848 & 0.949 & $0.013$ & $0.376$ & \\
    ICL (k=10), gpt-oss-120b & MICE Forest & 0.906 & 0.804 & $0.020$ & $0.562$ & \\
    ICL (k=100), gpt-oss-120b & MICE Forest & 0.906 & 0.804 & $0.024$ & $0.656$ & \\
    ICL (k=100), Qwen3-30B-A3B & MICE Forest & 0.841 & 0.804 & $0.511$ & $1.000$ & \\
    ICL (k=100), gpt-oss-120b & MICE PMM & 0.906 & 0.957 & $0.143$ & $1.000$ & \\
    ICL (k=10), gpt-oss-120b & TabPFN & 0.906 & 0.949 & $0.180$ & $1.000$ & \\
    ICL (k=100), Qwen3-30B-A3B & Most Similar Embed. & 0.841 & 0.870 & $0.597$ & $1.000$ & \\
    ICL (k=100), gpt-oss-120b & TabPFN & 0.906 & 0.949 & $0.180$ & $1.000$ & \\
    Most Similar Embed. & MICE Forest & 0.870 & 0.804 & $0.122$ & $1.000$ & \\
    ICL (k=100), Qwen3-30B-A3B & ICL (k=100), gpt-oss-120b & 0.841 & 0.906 & $0.064$ & $1.000$ & \\
    ICL (k=100), gpt-oss-120b & Most Similar Embed. & 0.906 & 0.870 & $0.424$ & $1.000$ & \\
    Complete Case & TabPFN & 0.957 & 0.949 & $1.000$ & $1.000$ & \\
    ICL (k=10), gpt-oss-120b & ICL (k=100), Qwen3-30B-A3B & 0.906 & 0.841 & $0.078$ & $1.000$ & \\
    ICL (k=10), gpt-oss-120b & Most Similar Embed. & 0.906 & 0.870 & $0.424$ & $1.000$ & \\
    ICL (k=10), gpt-oss-120b & ICL (k=100), gpt-oss-120b & 0.906 & 0.906 & $1.000$ & $1.000$ & \\
    ICL (k=10), Qwen3-30B-A3B & MICE Forest & 0.848 & 0.804 & $0.418$ & $1.000$ & \\
    ICL (k=10), Qwen3-30B-A3B & Most Similar Embed. & 0.848 & 0.870 & $0.720$ & $1.000$ & \\
    ICL (k=10), Qwen3-30B-A3B & ICL (k=100), gpt-oss-120b & 0.848 & 0.906 & $0.134$ & $1.000$ & \\
    ICL (k=10), Qwen3-30B-A3B & ICL (k=100), Qwen3-30B-A3B & 0.848 & 0.841 & $1.000$ & $1.000$ & \\
    MICE PMM & TabPFN & 0.957 & 0.949 & $1.000$ & $1.000$ & \\
    Complete Case & ICL (k=10), gpt-oss-120b & 0.957 & 0.906 & $0.118$ & $1.000$ & \\
    Zero-Shot, gpt-oss-120b & MICE Forest & 0.768 & 0.804 & $0.560$ & $1.000$ & \\
    Zero-Shot, gpt-oss-120b & Most Similar Embed. & 0.768 & 0.870 & $0.038$ & $1.000$ & \\
    Zero-Shot, gpt-oss-120b & ICL (k=100), Qwen3-30B-A3B & 0.768 & 0.841 & $0.087$ & $1.000$ & \\
    Zero-Shot, gpt-oss-120b & ICL (k=10), Qwen3-30B-A3B & 0.768 & 0.848 & $0.043$ & $1.000$ & \\
    Complete Case & ICL (k=100), gpt-oss-120b & 0.957 & 0.906 & $0.118$ & $1.000$ & \\
    Complete Case & MICE PMM & 0.957 & 0.957 & $1.000$ & $1.000$ & \\
    ICL (k=10), gpt-oss-120b & MICE PMM & 0.906 & 0.957 & $0.118$ & $1.000$ & \\
    ICL (k=10), Qwen3-30B-A3B & ICL (k=10), gpt-oss-120b & 0.848 & 0.906 & $0.152$ & $1.000$ & \\
    \bottomrule
\end{tabular}

  \caption{\textbf{Pairwise McNemar tests for coverage rate under MCAR missingness} (Holm-corrected, $\alpha=0.05$). Omnibus Cochran's Q test: $Q=124.97$, $p={<}\,0.001$, $N=138$.}
  \label{tab:sig-coverage-mcar}
\end{table*}

\begin{table*}[ht]
  \centering
  \footnotesize
  \begin{tabular}{llrrrrc}
    \toprule
    \textbf{Method 1} & \textbf{Method 2} & \textbf{Cov.\textsubscript{1}} & \textbf{Cov.\textsubscript{2}} & \textbf{$p_{\text{raw}}$} & \textbf{$p_{\text{Holm}}$} & \textbf{Sig.} \\
    \midrule
    Most Similar Embed. & MICE PMM & 0.513 & 0.953 & ${<}\,0.001$ & ${<}\,0.001$ & $\checkmark$ \\
    Complete Case & Most Similar Embed. & 0.917 & 0.513 & ${<}\,0.001$ & ${<}\,0.001$ & $\checkmark$ \\
    Most Similar Embed. & TabPFN & 0.513 & 0.884 & ${<}\,0.001$ & ${<}\,0.001$ & $\checkmark$ \\
    Zero-Shot, Qwen3-30B-A3B & MICE PMM & 0.596 & 0.953 & ${<}\,0.001$ & ${<}\,0.001$ & $\checkmark$ \\
    ICL (k=100), gpt-oss-120b & Most Similar Embed. & 0.881 & 0.513 & ${<}\,0.001$ & ${<}\,0.001$ & $\checkmark$ \\
    Complete Case & Zero-Shot, Qwen3-30B-A3B & 0.917 & 0.596 & ${<}\,0.001$ & ${<}\,0.001$ & $\checkmark$ \\
    Zero-Shot, gpt-oss-120b & MICE PMM & 0.650 & 0.953 & ${<}\,0.001$ & ${<}\,0.001$ & $\checkmark$ \\
    ICL (k=10), gpt-oss-120b & Most Similar Embed. & 0.841 & 0.513 & ${<}\,0.001$ & ${<}\,0.001$ & $\checkmark$ \\
    Most Similar Embed. & MICE Forest & 0.513 & 0.848 & ${<}\,0.001$ & ${<}\,0.001$ & $\checkmark$ \\
    Zero-Shot, Qwen3-30B-A3B & ICL (k=100), gpt-oss-120b & 0.596 & 0.881 & ${<}\,0.001$ & ${<}\,0.001$ & $\checkmark$ \\
    Zero-Shot, Qwen3-30B-A3B & TabPFN & 0.596 & 0.884 & ${<}\,0.001$ & ${<}\,0.001$ & $\checkmark$ \\
    Complete Case & Zero-Shot, gpt-oss-120b & 0.917 & 0.650 & ${<}\,0.001$ & ${<}\,0.001$ & $\checkmark$ \\
    ICL (k=10), Qwen3-30B-A3B & MICE PMM & 0.715 & 0.953 & ${<}\,0.001$ & ${<}\,0.001$ & $\checkmark$ \\
    Zero-Shot, Qwen3-30B-A3B & ICL (k=10), gpt-oss-120b & 0.596 & 0.841 & ${<}\,0.001$ & ${<}\,0.001$ & $\checkmark$ \\
    Zero-Shot, Qwen3-30B-A3B & MICE Forest & 0.596 & 0.848 & ${<}\,0.001$ & ${<}\,0.001$ & $\checkmark$ \\
    Zero-Shot, gpt-oss-120b & TabPFN & 0.650 & 0.884 & ${<}\,0.001$ & ${<}\,0.001$ & $\checkmark$ \\
    ICL (k=100), Qwen3-30B-A3B & Most Similar Embed. & 0.780 & 0.513 & ${<}\,0.001$ & ${<}\,0.001$ & $\checkmark$ \\
    Zero-Shot, gpt-oss-120b & ICL (k=100), gpt-oss-120b & 0.650 & 0.881 & ${<}\,0.001$ & ${<}\,0.001$ & $\checkmark$ \\
    Complete Case & ICL (k=10), Qwen3-30B-A3B & 0.917 & 0.715 & ${<}\,0.001$ & ${<}\,0.001$ & $\checkmark$ \\
    Zero-Shot, Qwen3-30B-A3B & ICL (k=100), Qwen3-30B-A3B & 0.596 & 0.780 & ${<}\,0.001$ & ${<}\,0.001$ & $\checkmark$ \\
    ICL (k=100), Qwen3-30B-A3B & MICE PMM & 0.780 & 0.953 & ${<}\,0.001$ & ${<}\,0.001$ & $\checkmark$ \\
    ICL (k=10), Qwen3-30B-A3B & ICL (k=100), gpt-oss-120b & 0.715 & 0.881 & ${<}\,0.001$ & ${<}\,0.001$ & $\checkmark$ \\
    Zero-Shot, gpt-oss-120b & MICE Forest & 0.650 & 0.848 & ${<}\,0.001$ & ${<}\,0.001$ & $\checkmark$ \\
    Zero-Shot, gpt-oss-120b & ICL (k=10), gpt-oss-120b & 0.650 & 0.841 & ${<}\,0.001$ & ${<}\,0.001$ & $\checkmark$ \\
    ICL (k=10), Qwen3-30B-A3B & TabPFN & 0.715 & 0.884 & ${<}\,0.001$ & ${<}\,0.001$ & $\checkmark$ \\
    Complete Case & ICL (k=100), Qwen3-30B-A3B & 0.917 & 0.780 & ${<}\,0.001$ & ${<}\,0.001$ & $\checkmark$ \\
    ICL (k=10), Qwen3-30B-A3B & Most Similar Embed. & 0.715 & 0.513 & ${<}\,0.001$ & ${<}\,0.001$ & $\checkmark$ \\
    ICL (k=10), gpt-oss-120b & MICE PMM & 0.841 & 0.953 & ${<}\,0.001$ & ${<}\,0.001$ & $\checkmark$ \\
    MICE PMM & MICE Forest & 0.953 & 0.848 & ${<}\,0.001$ & ${<}\,0.001$ & $\checkmark$ \\
    ICL (k=10), Qwen3-30B-A3B & ICL (k=10), gpt-oss-120b & 0.715 & 0.841 & ${<}\,0.001$ & $0.002$ & $\checkmark$ \\
    ICL (k=10), Qwen3-30B-A3B & MICE Forest & 0.715 & 0.848 & ${<}\,0.001$ & $0.004$ & $\checkmark$ \\
    Zero-Shot, Qwen3-30B-A3B & ICL (k=10), Qwen3-30B-A3B & 0.596 & 0.715 & ${<}\,0.001$ & $0.004$ & $\checkmark$ \\
    Zero-Shot, gpt-oss-120b & ICL (k=100), Qwen3-30B-A3B & 0.650 & 0.780 & ${<}\,0.001$ & $0.007$ & $\checkmark$ \\
    ICL (k=100), Qwen3-30B-A3B & ICL (k=100), gpt-oss-120b & 0.780 & 0.881 & ${<}\,0.001$ & $0.014$ & $\checkmark$ \\
    ICL (k=100), Qwen3-30B-A3B & TabPFN & 0.780 & 0.884 & ${<}\,0.001$ & $0.019$ & $\checkmark$ \\
    Complete Case & ICL (k=10), gpt-oss-120b & 0.917 & 0.841 & $0.001$ & $0.029$ & $\checkmark$ \\
    ICL (k=100), gpt-oss-120b & MICE PMM & 0.881 & 0.953 & $0.002$ & $0.032$ & $\checkmark$ \\
    Zero-Shot, gpt-oss-120b & Most Similar Embed. & 0.650 & 0.513 & $0.002$ & $0.038$ & $\checkmark$ \\
    MICE PMM & TabPFN & 0.953 & 0.884 & $0.003$ & $0.057$ & \\
    ICL (k=10), Qwen3-30B-A3B & ICL (k=100), Qwen3-30B-A3B & 0.715 & 0.780 & $0.008$ & $0.127$ & \\
    Complete Case & MICE Forest & 0.917 & 0.848 & $0.013$ & $0.190$ & \\
    ICL (k=10), gpt-oss-120b & ICL (k=100), Qwen3-30B-A3B & 0.841 & 0.780 & $0.043$ & $0.601$ & \\
    ICL (k=100), Qwen3-30B-A3B & MICE Forest & 0.780 & 0.848 & $0.059$ & $0.753$ & \\
    Zero-Shot, Qwen3-30B-A3B & Most Similar Embed. & 0.596 & 0.513 & $0.058$ & $0.753$ & \\
    Zero-Shot, gpt-oss-120b & ICL (k=10), Qwen3-30B-A3B & 0.650 & 0.715 & $0.069$ & $0.763$ & \\
    ICL (k=10), gpt-oss-120b & ICL (k=100), gpt-oss-120b & 0.841 & 0.881 & $0.080$ & $0.801$ & \\
    Complete Case & ICL (k=100), gpt-oss-120b & 0.917 & 0.881 & $0.132$ & $0.934$ & \\
    Complete Case & MICE PMM & 0.917 & 0.953 & $0.110$ & $0.934$ & \\
    Zero-Shot, Qwen3-30B-A3B & Zero-Shot, gpt-oss-120b & 0.596 & 0.650 & $0.163$ & $0.934$ & \\
    ICL (k=100), gpt-oss-120b & MICE Forest & 0.881 & 0.848 & $0.298$ & $0.934$ & \\
    Complete Case & TabPFN & 0.917 & 0.884 & $0.150$ & $0.934$ & \\
    ICL (k=10), gpt-oss-120b & TabPFN & 0.841 & 0.884 & $0.104$ & $0.934$ & \\
    MICE Forest & TabPFN & 0.848 & 0.884 & $0.229$ & $0.934$ & \\
    ICL (k=10), gpt-oss-120b & MICE Forest & 0.841 & 0.848 & $0.904$ & $1.000$ & \\
    ICL (k=100), gpt-oss-120b & TabPFN & 0.881 & 0.884 & $1.000$ & $1.000$ & \\
    \bottomrule
\end{tabular}

  \caption{\textbf{Pairwise McNemar tests for coverage rate under MAR missingness} (Holm-corrected, $\alpha=0.05$). Omnibus Cochran's Q test: $Q=369.80$, $p={<}\,0.001$, $N=277$.}
  \label{tab:sig-coverage-mar}
\end{table*}

\begin{table*}[ht]
  \centering
  \footnotesize
  \begin{tabular}{llrrrrc}
    \toprule
    \textbf{Method 1} & \textbf{Method 2} & \textbf{Cov.\textsubscript{1}} & \textbf{Cov.\textsubscript{2}} & \textbf{$p_{\text{raw}}$} & \textbf{$p_{\text{Holm}}$} & \textbf{Sig.} \\
    \midrule
    Zero-Shot, Qwen3-30B-A3B & ICL (k=10), gpt-oss-120b & 0.680 & 0.891 & ${<}\,0.001$ & ${<}\,0.001$ & $\checkmark$ \\
    Complete Case & MICE Forest & 0.884 & 0.727 & ${<}\,0.001$ & ${<}\,0.001$ & $\checkmark$ \\
    Zero-Shot, Qwen3-30B-A3B & MICE PMM & 0.680 & 0.884 & ${<}\,0.001$ & ${<}\,0.001$ & $\checkmark$ \\
    Zero-Shot, Qwen3-30B-A3B & ICL (k=100), gpt-oss-120b & 0.680 & 0.876 & ${<}\,0.001$ & ${<}\,0.001$ & $\checkmark$ \\
    Complete Case & Zero-Shot, Qwen3-30B-A3B & 0.884 & 0.680 & ${<}\,0.001$ & ${<}\,0.001$ & $\checkmark$ \\
    Zero-Shot, Qwen3-30B-A3B & ICL (k=100), Qwen3-30B-A3B & 0.680 & 0.844 & ${<}\,0.001$ & ${<}\,0.001$ & $\checkmark$ \\
    MICE PMM & MICE Forest & 0.884 & 0.727 & ${<}\,0.001$ & ${<}\,0.001$ & $\checkmark$ \\
    Zero-Shot, Qwen3-30B-A3B & TabPFN & 0.680 & 0.862 & ${<}\,0.001$ & ${<}\,0.001$ & $\checkmark$ \\
    ICL (k=10), gpt-oss-120b & MICE Forest & 0.891 & 0.727 & ${<}\,0.001$ & ${<}\,0.001$ & $\checkmark$ \\
    Zero-Shot, Qwen3-30B-A3B & ICL (k=10), Qwen3-30B-A3B & 0.680 & 0.818 & ${<}\,0.001$ & ${<}\,0.001$ & $\checkmark$ \\
    ICL (k=100), gpt-oss-120b & MICE Forest & 0.876 & 0.727 & ${<}\,0.001$ & ${<}\,0.001$ & $\checkmark$ \\
    MICE Forest & TabPFN & 0.727 & 0.862 & ${<}\,0.001$ & $0.002$ & $\checkmark$ \\
    ICL (k=10), gpt-oss-120b & Most Similar Embed. & 0.891 & 0.793 & ${<}\,0.001$ & $0.004$ & $\checkmark$ \\
    Complete Case & Most Similar Embed. & 0.884 & 0.793 & ${<}\,0.001$ & $0.010$ & $\checkmark$ \\
    Zero-Shot, gpt-oss-120b & ICL (k=10), gpt-oss-120b & 0.778 & 0.891 & ${<}\,0.001$ & $0.010$ & $\checkmark$ \\
    Most Similar Embed. & MICE PMM & 0.793 & 0.884 & ${<}\,0.001$ & $0.032$ & $\checkmark$ \\
    Zero-Shot, gpt-oss-120b & MICE PMM & 0.778 & 0.884 & $0.001$ & $0.049$ & $\checkmark$ \\
    ICL (k=100), Qwen3-30B-A3B & MICE Forest & 0.844 & 0.727 & $0.001$ & $0.054$ & \\
    Complete Case & Zero-Shot, gpt-oss-120b & 0.884 & 0.778 & $0.001$ & $0.054$ & \\
    ICL (k=100), gpt-oss-120b & Most Similar Embed. & 0.876 & 0.793 & $0.002$ & $0.064$ & \\
    Zero-Shot, Qwen3-30B-A3B & Zero-Shot, gpt-oss-120b & 0.680 & 0.778 & $0.002$ & $0.086$ & \\
    Zero-Shot, Qwen3-30B-A3B & Most Similar Embed. & 0.680 & 0.793 & $0.003$ & $0.090$ & \\
    Zero-Shot, gpt-oss-120b & ICL (k=100), gpt-oss-120b & 0.778 & 0.876 & $0.003$ & $0.092$ & \\
    Most Similar Embed. & TabPFN & 0.793 & 0.862 & $0.005$ & $0.173$ & \\
    ICL (k=10), Qwen3-30B-A3B & ICL (k=10), gpt-oss-120b & 0.818 & 0.891 & $0.008$ & $0.241$ & \\
    ICL (k=10), Qwen3-30B-A3B & MICE Forest & 0.818 & 0.727 & $0.011$ & $0.344$ & \\
    Zero-Shot, gpt-oss-120b & TabPFN & 0.778 & 0.862 & $0.017$ & $0.479$ & \\
    Zero-Shot, gpt-oss-120b & ICL (k=100), Qwen3-30B-A3B & 0.778 & 0.844 & $0.025$ & $0.693$ & \\
    ICL (k=10), Qwen3-30B-A3B & MICE PMM & 0.818 & 0.884 & $0.030$ & $0.810$ & \\
    ICL (k=10), Qwen3-30B-A3B & ICL (k=100), gpt-oss-120b & 0.818 & 0.876 & $0.033$ & $0.852$ & \\
    Complete Case & ICL (k=10), Qwen3-30B-A3B & 0.884 & 0.818 & $0.033$ & $0.852$ & \\
    ICL (k=100), Qwen3-30B-A3B & MICE PMM & 0.844 & 0.884 & $0.177$ & $1.000$ & \\
    ICL (k=100), Qwen3-30B-A3B & Most Similar Embed. & 0.844 & 0.793 & $0.109$ & $1.000$ & \\
    ICL (k=100), gpt-oss-120b & MICE PMM & 0.876 & 0.884 & $0.878$ & $1.000$ & \\
    ICL (k=100), Qwen3-30B-A3B & ICL (k=100), gpt-oss-120b & 0.844 & 0.876 & $0.222$ & $1.000$ & \\
    ICL (k=100), gpt-oss-120b & TabPFN & 0.876 & 0.862 & $0.627$ & $1.000$ & \\
    Most Similar Embed. & MICE Forest & 0.793 & 0.727 & $0.060$ & $1.000$ & \\
    ICL (k=100), Qwen3-30B-A3B & TabPFN & 0.844 & 0.862 & $0.597$ & $1.000$ & \\
    ICL (k=10), gpt-oss-120b & TabPFN & 0.891 & 0.862 & $0.256$ & $1.000$ & \\
    Complete Case & TabPFN & 0.884 & 0.862 & $0.377$ & $1.000$ & \\
    ICL (k=10), gpt-oss-120b & MICE PMM & 0.891 & 0.884 & $0.868$ & $1.000$ & \\
    ICL (k=10), gpt-oss-120b & ICL (k=100), gpt-oss-120b & 0.891 & 0.876 & $0.585$ & $1.000$ & \\
    ICL (k=10), gpt-oss-120b & ICL (k=100), Qwen3-30B-A3B & 0.891 & 0.844 & $0.066$ & $1.000$ & \\
    ICL (k=10), Qwen3-30B-A3B & Most Similar Embed. & 0.818 & 0.793 & $0.494$ & $1.000$ & \\
    ICL (k=10), Qwen3-30B-A3B & ICL (k=100), Qwen3-30B-A3B & 0.818 & 0.844 & $0.248$ & $1.000$ & \\
    MICE PMM & TabPFN & 0.884 & 0.862 & $0.377$ & $1.000$ & \\
    Zero-Shot, gpt-oss-120b & MICE Forest & 0.778 & 0.727 & $0.184$ & $1.000$ & \\
    Zero-Shot, gpt-oss-120b & Most Similar Embed. & 0.778 & 0.793 & $0.749$ & $1.000$ & \\
    Zero-Shot, gpt-oss-120b & ICL (k=10), Qwen3-30B-A3B & 0.778 & 0.818 & $0.200$ & $1.000$ & \\
    Complete Case & ICL (k=100), Qwen3-30B-A3B & 0.884 & 0.844 & $0.185$ & $1.000$ & \\
    Zero-Shot, Qwen3-30B-A3B & MICE Forest & 0.680 & 0.727 & $0.255$ & $1.000$ & \\
    Complete Case & ICL (k=100), gpt-oss-120b & 0.884 & 0.876 & $0.878$ & $1.000$ & \\
    Complete Case & MICE PMM & 0.884 & 0.884 & $1.000$ & $1.000$ & \\
    Complete Case & ICL (k=10), gpt-oss-120b & 0.884 & 0.891 & $0.871$ & $1.000$ & \\
    ICL (k=10), Qwen3-30B-A3B & TabPFN & 0.818 & 0.862 & $0.169$ & $1.000$ & \\
    \bottomrule
\end{tabular}

  \caption{\textbf{Pairwise McNemar tests for coverage rate under MNAR missingness} (Holm-corrected, $\alpha=0.05$). Omnibus Cochran's Q test: $Q=117.20$, $p={<}\,0.001$, $N=275$.}
  \label{tab:sig-coverage-mnar}
\end{table*}

\clearpage
\section{Prompt Formats}\label{app:prompts}

Figures~\ref{fig:prompt_zero_shot}--\ref{fig:prompt_single_turn_verbalized} reproduce the four prompt formats: the zero-shot prompt of the baseline (Section~\ref{sec:baseline_methods}) and the \textit{Multi-turn}, \textit{Single-turn}, and \textit{Single-turn verbalized} prompts of the configuration selection (Appendix~\ref{app:icl}). Each is rendered for one target query with $k=2$ in-context examples for brevity ($k\in\{10, 100\}$ in the experiments) and with four of the twelve demographic variables, the others indicated by \texttt{[\ldots]}. Text in {\color{promptvarcolor}\textbf{orange}} is data-derived and varies across queries, including the header of the opinion variable. Text in {\color{promptcfgcolor}\textbf{teal}} depends on the configuration (the number of examples~$k$ and the retrieval strategy). The remaining text is a fixed template. The snippets are generated by the prompt-construction functions of the experiments, via an extraction script in the released code.

\begin{figure}[!ht]
  \input{figures/prompts/prompt_zero_shot.tex}
  \caption{\textbf{Zero-shot prompt} ($k=0$ retrieved examples). The
    single-turn template (Figure~\ref{fig:prompt_single_turn}) is reused without any in-context examples, so the
    model is asked to produce a verbalized distribution from the target row
    and the system instructions alone. This is the \textit{Zero-Shot LLM}
    baseline of Section~\ref{sec:baseline_methods}.}
  \label{fig:prompt_zero_shot}
\end{figure}

\begin{figure}[!tbp]
  \input{figures/prompts/prompt_multi_turn.tex}
  \caption{\textbf{Multi-turn prompt.} Each retrieved example is presented as
    a separate user/assistant exchange so the model sees the target column
    header repeated in every turn. The trailing user message contains the row
    to be imputed.}
  \label{fig:prompt_multi_turn}
\end{figure}

\begin{figure}[!tbp]
  \input{figures/prompts/prompt_single_turn.tex}
  \caption{\textbf{Single-turn prompt.} All retrieved examples are
    concatenated into a single user message together with a natural-language
    description of the retrieval strategy. The wording of the description
    changes with the retrieval strategy (``most similar'', ``least similar'',
    ``diverse'', ``random sample'') and with stratification; the
    \textit{Most Similar}, non-stratified variant is shown.}
  \label{fig:prompt_single_turn}
\end{figure}

\begin{figure}[!tbp]
  \input{figures/prompts/prompt_single_turn_verbalized.tex}
  \caption{\textbf{Single-turn verbalized prompt.} Identical to the
    single-turn prompt except that the format instructions ask the model to
    emit a JSON-encoded probability distribution over the answer options,
    from which we sample the
    imputation. This is the format
    we selected.}
  \label{fig:prompt_single_turn_verbalized}
\end{figure}

\end{document}